\documentclass[11pt,onecolumn,letterpaper,final]{IEEEtran}
\IEEEoverridecommandlockouts

\usepackage[colorlinks=true,citecolor=blue,linkcolor=blue]{hyperref}
\newcommand{\circled}[1]{\small{\raisebox{.6pt}{\textcircled{\raisebox{-.8pt}{#1}}}}}

\usepackage{amssymb}
\usepackage{amsmath,mathrsfs,dsfont}
\usepackage{nicefrac}
\usepackage{algorithm}
\usepackage{algorithmicx}
\usepackage{algpseudocode}

\usepackage{booktabs}

\usepackage{color}
\usepackage{enumitem}

\usepackage{array,cite}
\usepackage{graphicx,tikz}
\usepackage[mathscr]{euscript}
\usepackage{amsthm}
\usepackage{cite}

\usepackage{bm}
\usepackage{bbm}
\usepackage{color}

\usepackage{color}
\usepackage{epstopdf}
\usepackage{subcaption}
\usepackage{cleveref}
\usepackage{thmtools}
\usepackage{thm-restate}

\usepackage{bbding}
\usepackage{mathtools}
\usepackage{wrapfig}
\usepackage{colortbl}
\usepackage{slashbox}
\usepackage[htt]{hyphenat}
\allowdisplaybreaks
\newcommand{\cfrakR}{\mathfrak{R}} 
 
\newcommand{\relu}[1]{\sigma\pth{#1}}

\newcommand{\tK}{\tilde K}

\newcommand{\bx}{\tilde \bx}
\newcommand{\by}{{\tilde \by}}

\newcommand{\bbx}{\overset{\rightharpoonup}{\bx}}

\newcommand{\bbw}{\overset{\rightharpoonup}{\bw}}
\newcommand{\bbe}{\overset{\rightharpoonup}{\be}}
\newcommand{\bbq}{\overset{\rightharpoonup}{\bq}}
\newcommand{\tmu}{\tilde \mu}

\newcommand{\cFnn}{\cF_{\mathop\mathrm{NN}}}

\newcommand{\Kattn}{{K}^{\mathop\mathrm{(att)}}}
\newcommand{\hKattn}{{\hat K}^{\mathop\mathrm{(att)}}}
\newcommand{\barKattn}{{\bar K}^{\mathop\mathrm{(att)}}}
\newcommand{\bKattn}{{\bK}^{\mathop\mathrm{(att)}}}
\newcommand{\hbKattn}{{\hat \bK}^{\mathop\mathrm{(att)}}}

\newcommand{\bSigmaattn}{{\bSigma}^{\mathop\mathrm{(att)}}}
\newcommand{\bUattn}{{\bU}^{\mathop\mathrm{(att)}}}
\newcommand{\lambdaattn}{{\lambda}^{\mathop\mathrm{(att)}}}
\newcommand{\hlambdaattn}{{\hat \lambda}^{\mathop\mathrm{(att)}}}

\newcommand{\Kint}{{K}^{\mathop\mathrm{(int)}}}
\newcommand{\hKint}{{\hat K}^{\mathop\mathrm{(int)}}}
\newcommand{\bKint}{{\bK}^{\mathop\mathrm{(int)}}}
\newcommand{\hbKint}{{\hat \bK}^{\mathop\mathrm{(int)}}}

\newcommand{\cHKattn}{\cH_{\Kattn}}

\title{Sharp Generalization for Nonparametric Regression in Interpolation Space by Shallow Over-Parameterized Neural Networks with Channel Attention}

\author{Yingzhen Yang$^{1}$ and Ping Li$^{2}$,
\\
$^{1}$Arizona State University,
699 S Mill RD, Tempe, AZ 85281, USA\\
$^{2}$VecML Inc., Bellevue, WA 98004, USA \\
}
\graphicspath{{illustrations/}}
\newcounter{optproblem}

\newtheoremstyle{mytheoremstyle} 
    {\topsep}                    
    {\topsep}                    
    {\normalfont}                
    {}                           
    {\bfseries}                   
    {.}                          
    {.5em}                       
    {}  

\theoremstyle{mytheoremstyle}
\newtheorem{theorem}{Theorem}[section]

\newtheorem{proposition}[theorem]{Proposition}
\newtheorem{corollary}[theorem]{Corollary}

\newtheorem*{theorem*}{Theorem}
\newtheorem*{lemma*}{Lemma}
\newtheorem*{remark*}{Remark}

\newtheorem{lemma}[theorem]{Lemma}

\theoremstyle{remark}
\newtheorem{definition}{Definition}[section]

\DeclareMathAlphabet{\pazocal}{OMS}{zplm}{m}{n}
\DeclareMathAlphabet{\mathpzc}{OMS}{pzc}{m}{it}

\setlist[itemize]{leftmargin=*}

\renewcommand{\hat}{\widehat}

\newcommand{\bfm}[1]{\ensuremath{\mathbf{#1}}}
\newcommand{\bfsym}[1]{\ensuremath{\boldsymbol{#1}}}

\def\ba{\boldsymbol a}   \def\bA{\bfm A}  
     
\def\bc{\bfm c}     
   \def\bD{\bfm D}  
\def\be{\bfm e}   \def\bE{\bfm E}  
  \def\bF{\bfm F}  
\def\bg{\bfm g}   \def\bG{\bfm G}  
\def\bh{\bfm h}   \def\bH{\bfm H}  
   \def\bI{\bfm I}  
   \def\bJ{\bfm J}  
\def\bk{\bfm k}   \def\bK{\bfm K}  
   \def\bL{\bfm L}  
   \def\bM{\bfm M}

\def\bq{\bfm q}   \def\bQ{\bfm Q}  
   \def\bR{\bfm R}  \def\RR{\mathbb{R}}
   \def\bS{\bfm S}  
     
\def\bu{\bfm u}   \def\bU{\bfm U}  
\def\bv{\bfm v}   \def\bV{\bfm V}  
\def\bw{\bfm w}   \def\bW{\bfm W}  
\def\bx{\bfm x}   \def\bX{\bfm X}  
\def\by{\bfm y}   \def\bY{\bfm Y}  
\def\bz{\bfm z}   \def\bZ{\bfm Z}  
\def\bzero{\bfm 0}

 \def\cD{{\cal  D}}
 \def\cE{{\cal  E}}
 \def\cF{{\cal  F}}
 \def\cG{{\cal  G}}
 \def\cH{{\cal  H}}
 
 \def\cJ{{\cal  J}}
 
 \def\cL{{\cal  L}}
 
 \def\cN{{\cal  N}}
 \def\cO{{\cal  O}}

 \def\cT{{\cal  T}}
 
 \def\cV{{\cal  V}}
 \def\cW{{\cal  W}}
 \def\cX{{\cal  X}}

\def\btheta{\bfsym {\theta}}

\def\bsigma{\bfsym \sigma}
\def\bSigma{\bfsym \Sigma}

\def\bxi{\bfsym {\xi}}

\def\hlambda{\hat{\lambda}}

\def\+#1{\mathcal{#1}}
\def\-#1{\textup{#1}}

\def\set#1{\left\{ #1 \right\}}
\def\pth#1{\left( #1 \right)}
\def\bth#1{\left[ #1 \right]}
\def\abth#1{\left | #1 \right |}

\def\defeq {\coloneqq}

\newcommand{\vect}[1]{{\textup{vec}\pth{#1}}}

\DeclareMathSymbol{\relcolon}{\mathrel}{operators}{"3A}

\newcommand{\diag}{\textup{diag}}

\newcommand{\La}{\left\langle\kern-0.64ex\left\langle}
\newcommand{\Ra}{\right\rangle\kern-0.64ex\right\rangle}

\def\Norm#1#2{{\left\vert\kern-0.4ex\left\vert\kern-0.4ex\left\vert #1
    \right\vert\kern-0.4ex\right\vert\kern-0.4ex\right\vert}_{#2}}
\def\norm#1#2{{\left\|#1\right\|}_{#2}}

\def\ltwonorm#1{\norm{#1}{2}}
\def\fnorm#1{\norm{#1}{\textup{F}}}
\def\supnorm#1{\norm{#1}{\infty}}

\def\tr#1{\textup{tr}\left(#1\right)}

\newcommand{\1}{{\rm 1}\kern-0.25em{\rm I}}
\def\indict#1{{\rm 1}\kern-0.25em{\rm I}_{\set{#1}}}

\def \eps  {\epsilon}
\def \eps {\varepsilon}

\def \diff {{\rm d}}
\def \iprod#1#2{\left\langle #1, #2 \right\rangle}

\def\set#1{\left\{#1\right\}}

\def\unitsphere#1{\mathbb{S}^{#1}}

\def \E {\mathbb{E}}
\def\Expect#1#2{\E_{#1}\left[#2\right]}

\def \Pr {\textup{Pr}}
\newcommand{\Prob}[1]{\Pr\left[#1\right]}
\def \Var#1{\textup{Var}\left[#1\right]}

\def \lsim {\lesssim}
\def \gsim {\gtrsim}

\newcommand{\Unif}[1]{{\mathop\mathrm{Unif}}\left( #1 \right)}

\newcommand{\beq}{\begin{equation}}
\newcommand{\eeq}{\end{equation}}
\newcommand{\beqa}{\begin{eqnarray}}
\newcommand{\eeqa}{\end{eqnarray}}
\newcommand{\beqas}{\begin{eqnarray*}}
\newcommand{\eeqas}{\end{eqnarray*}}
\def\bal#1\eal{\begin{align}#1\end{align}}
\def\bals#1\eals{\begin{align*}#1\end{align*}}
\def\bsal#1\esal{\begin{small}\begin{align}#1\end{align}\end{small}}
\def\bsals#1\esals{\begin{small}\begin{align*}#1\end{align*}\end{small}}
\def\bsfal#1\esfal{\begin{small}\begin{flalign}#1\end{flalign}\end{small}}

\begin{document}

\def\theHtheorem{\theHsection.\arabic{theorem}}
\def\theHlemma{\theHtheorem}
\def\theHproposition{\theHtheorem}
\def\theHcorollary{\theHtheorem}
\def\theHdefinition{\theHsection.\arabic{definition}}
\captionsetup[figure]{hypcap=false}

\maketitle

\renewcommand{\baselinestretch}{1.10}
\selectfont

\begin{abstract}
We study nonparametric regression using an over-parameterized two-layer ReLU neural network with fixed covariance-based channel attention, trained by gradient descent with both layers updated simultaneously. The spherical covariates follow an arbitrary fixed non-atomic distribution, and the target function lies in an interpolation space with spectral bias widely studied in the statistical learning literature. We prove that early-stopped training achieves regression risk $\cO(\eps_n^2)$, where $\eps_n$ is the critical population rate of the attention kernel $\Kattn$. Its positive eigenvalues are $\lambda_j^{s'}$, where $\lambda_j$ are those of the ReLU feature covariance kernel $K$ and $s'=2s+3$ for $s\ge1$. Under polynomial eigenvalue decay $\lambda_j\asymp j^{-2\alpha}$ with $\alpha>1/2$, the regression risk bound $\cO(\eps_n^2)$ specializes to the minimax optimal rate $\cO(n^{-2\alpha s'/(2\alpha s'+1)})$. Our analysis is based on two new technical results on the optimization and generalization of the attention network. First, the neural network function after every gradient descent iteration admits a decomposition into a function with bounded Reproducing Kernel Hilbert Space (RKHS) norm and a uniformly small residual. Second, local Rademacher complexity converts this decomposition into a regression risk bound. Kernel gradient descent on the same ball attains the same rate, while our theorem gives explicit sufficient width and auxiliary-sample conditions for simultaneous network training. Our analysis is further extended to covariates supported on $\RR^d$ through a novel normalization method. Under a sub-Gaussian covariate distribution, the attention network achieves a nearly optimal regression risk $\cO((\log n)n^{-s'(d+3)/(s'(d+3)+d)})$ without a density assumption.
\end{abstract}
\begin{IEEEkeywords}

\smallskip
\noindent  Nonparametric Regression, Interpolation Space, Over-Parameterized Neural Network, Channel Attention, Minimax Optimal Rate
\end{IEEEkeywords}

\section{Introduction}
\vspace{-.05in}
The remarkable success of deep learning~\cite{YannLecunNature05-DeepLearning} has motivated studies on the optimization and generalization of deep neural networks (DNNs). Many works show that gradient descent (GD) and stochastic gradient descent (SGD) fit training data in over-parameterized networks~\cite{du2018gradient-gd-dnns,AllenZhuLS19-convergence-dnns,DuLL0Z19-GD-dnns}. The Neural Tangent Kernel (NTK)~\cite{JacotHG18-NTK} connects this training with kernel methods: when weights remain near initialization, a first-order approximation permits analysis of the training dynamics and regression risk~\cite{CaoG19a-sgd-wide-dnns,AroraDHLW19-fine-grained-two-layer,Ghorbani2021-linearized-two-layer-nn}.

For nonparametric regression with noisy data, a further question is whether training attains the statistical rate suggested by the target class. Approximation results give minimax rates for smooth~\cite{Yarotsky17-error-approximation-dnns,Bauer2019-regression-dnns-curse-dim,SchmidtHieber2020-regression-relu-dnns} and non-smooth~\cite{ImaizumiF19-nonsmooth-regression} targets, but do not by themselves show that GD realizes the required functions. More recent works~\cite{HuWLC21-regularization-minimax-uniform-spherical,SuhKH22-overparameterized-gd-minimax,Li2024-edr-general-domain,Yang2025-generalization-two-layer-regression} analyze the regression risk of trained networks. Our goal of this paper  is to quantify how simultaneous training of both layers of a two-layer neural network (NN) with channel attention achieve sharp regression risk associated with an interpolation-space target.

\smallskip
\noindent\textbf{Existing Empirical and Theoretical Works about Channel Attention and General Attention Mechanism.}
Popular channel attention methods~\cite{DANet,ECA,XCiT} reweight channels to improve neural-network representations. In particular, XCiT~\cite{XCiT} views channel attention as cross-covariance across features. Covariance pooling~\cite{ChenS0LS25,SongS021,WangZGXZLZH23}, kernelizable attention~\cite{ChoromanskiLDSG21-attention-random-feature-approx,Peng0Y0SK21-rf-attention}, infinite-head attention~\cite{HronBSN20-infinite-attention} and in-context learning~\cite{kim2024transformers-optimal-in-context} provide related constructions and analyses. These studies motivate analyzing the regression risk of trained neural networks with channel attention.

In this paper, we study a two-layer NN with fixed covariance-based channel attention inspired by cross-covariance attention. The matrix in Section~\ref{sec:setup} reweights the channels to induce stronger kernel spectral decay and differs from XCiT's normalized softmax matrix. Section~\ref{sec:main-results} presents the regression risk bounds for spherical covariates and their polynomial-spectrum specialization. Section~\ref{sec:results-unbounded-input-space} extends these results to covariate distributions supported on $\RR^d$ by normalizing augmented inputs and responses and controlling the covariate tail when restoring the response scale. A finite second moment gives the eigenvalue upper bound, and a sub-Gaussian tail yields an explicit nearly optimal regression risk bound without a density assumption. Section~\ref{sec:proof-roadmap} presents the key technical results and their role in the proof. Appendix~\ref{sec:proof-vanilla-bias} establishes the biased vanilla comparison on the same Euclidean target ball.

\smallskip
\noindent\textbf{Related works on regression risk.}
 RKHS denotes a reproducing kernel Hilbert space. The works in~\cite{HuWLC21-regularization-minimax-uniform-spherical} and \cite{SuhKH22-overparameterized-gd-minimax} obtain minimax regression risk bounds for NTK-RKHS targets under uniform spherical covariates using regularized GD. The work in~\cite{Li2024-edr-general-domain} analyzes interpolation-space targets using early-stopped gradient flow. The work in~\cite{Yang2025-generalization-two-layer-regression} obtains critical-radius regression risk bounds for continuous spherical laws with fixed output weights. Table~\ref{table:regression-risk-comparison} records their kernels, target classes and distributional assumptions. The work in~\cite{NitandaS21} establishes source-dependent minimax rates in the NTK regime for regularized online averaged SGD on smooth two-layer networks, including smooth ReLU approximations. They assume bounded inputs and labels and source exponents from one-half to one. Their regression risk averages over sampled SGD observations on a high-probability initialization event, with sufficiently large width. Our theorem instead quantifies full-batch ReLU GD with simultaneous training on the stated higher-power interpolation ball through explicit sufficient width and auxiliary-sample conditions.
\smallskip
\noindent The work in~\cite{pmlr-v313-yang26a} studies low-degree spherical polynomials under uniform covariates. For target degree $\ell_0$, a learnable harmonic-degree selection stage precedes second-layer GD, giving a rank-$\Theta(d^{\ell_0})$ kernel and regression risk $\Theta(d^{\ell_0}/n)$. Our fixed-$d$ interpolation-space problem has infinitely many positive eigenvalues, fixes attention before training and updates both layers throughout the stopping interval.

\smallskip
\noindent\textbf{Contribution and comparison on a common statistical model.}
Our analysis connects the statistical rate determined by the attention kernel to simultaneous training of a finite NN. The joint optimization of both layers in our attention network enables interaction between changing output weights and moving ReLU features, which retains the spectral filtering of accumulated errors. Together with uniform covariance estimates, this yields explicit sufficient width and auxiliary-sample conditions and a decomposition into a fixed RKHS ball and a uniformly small residual. Local Rademacher complexity then gives the regression risk bound. The roadmaps after the main results and in Section~\ref{sec:proof-roadmap} describe these steps in detail.

On the same spherical target ball with the same kernel evaluations, Proposition~\ref{prop:kernel-baseline} attains the same regression risk rate by kernel GD. Our  contribution is the quantitative guarantee that a finite network with both layers trained retains this rate under explicit size conditions, with advantages over existing analysis~\cite{YangLi2026PGD} under the same source condition explained in
Section~\ref{sec:comparison-pgd}. Section~\ref{sec:comparison-pgd} compares these conditions with those of preconditioned gradient descent (PGD). The independent sample $\bQ$ and exact evaluations of the regular NTK and its induced kernels remain information assumptions of the construction and of the stated comparisons.

\smallskip
\noindent\textbf{Notations.}
Let $P$ be the fixed non-atomic covariate law and $\mu$ its Borel probability measure. For spherical covariates, $\cX=\operatorname{supp}(\mu)\subseteq\unitsphere{d-1}$ is compact, where $\unitsphere{d-1}=\set{\bx\in\RR^d:\ltwonorm{\bx}=1}$. For unboudned covariate, $\cX=\RR^d$. Spectral statements are under $\mu$. Bold letters denote matrices and vectors, superscript $\top$ denotes transpose, $\bA^{(i)}$ denotes column $i$ and subscripts denote rows or entries. An arrow indicates a vector such as $\bbx_i$. We use $\fnorm{\cdot}$ for the Frobenius norm, $\norm{\cdot}{p}$ for vector or matrix $p$-norm, $\vect{\cdot}$ for vectorization by stacking rows and $\tr{\cdot}$ for trace. $\ltwonorm{\cdot}$ also denotes the operator norm depending on the context. The notation $[m\relcolon n]$ denotes the inclusive integer range, abbreviated $[n]$ when $m=1$. In iteration statements $[a,b]=[a\relcolon b]$. The symbols $\bI_n$, $\indict{E}$, $A^c$ and $\Var{\cdot}$ denote the identity matrix, indicator, complement and variance. Vertical bars denote absolute value for scalars and cardinality for sets.
Write $\iprod{f}{g}_{\mu}=\iprod{f}{g}_{L^2}=\int_{\cX}fg\,d\mu$ and $\norm{f}{\mu}=\norm{f}{L^2}$ on $L^2(\cX,\mu)$. Inner products are real. For a Hilbert space $\cH$, $\iprod{\cdot}{\cdot}_{\cH}$ and $\norm{\cdot}{\cH}$ denote its inner product and norm, and $\cH(\mu_0)$ is its radius-$\mu_0$ ball centered at zero. $\norm{g}{\infty}=\sup_{\bx\in\cX}\abth{g(\bx)}$ denotes the supremum norm.  $\Unif{\unitsphere{d-1}}$ denotes the spherical uniform distribution and $\Expect{P}{\cdot}$ is an expectation under $P$.
The relations $a=\cO(b)$ or $a\lsim b$ mean $a\le cb$, $a\asymp b$ or $a=\Theta(b)$ mean two-sided constant bounds, and $a=o(b)$ or $a=\omega(b)$ mean $\abth{a/b}\to0$ or $\abth{a/b}\to\infty$. Comparison constants may vary and depend on the fixed law $\mu$ and fixed parameters $d,s,\kappa,\mu_0,\sigma_0,\eta,c_t$ and any two-sided eigenvalue-decay constants, but not on $n,m,N,T,\delta$. Fix $\mu_0,\sigma_0>0$, $s\ge1$ and $0<\kappa<1$, and we set $s'=2s+3$ and $b=\kappa^2/2$.

\vspace{-.05in}
\section{Problem Setup}
\label{sec:setup}
\vspace{-.05in}
Throughout this paper, the covariate law $P$ has a fixed non-atomic measure $\mu$. Non-atomicity excludes repeated and antipodal sample points and yields an infinite positive spectrum. Gaussian initialization permits uniform feature-covariance concentration. The interpolation-space condition specifies target regularity, while compact spherical support bounds the features. For $P$ supported on $\RR^d$, normalization provides bounded network inputs, and the finite second moment and covariate tail control the original regression risk. We first specify the spherical construction, then give the changes required on $\RR^d$.

\subsection{Two-Layer Neural Network}
\label{sec:setup-two-layer-nn}

This subsection and Section~\ref{sec:setup-kernels-target-function} describe the spherical construction. Section~\ref{sec:setup-unbounded} specifies the changes for $\cX=\RR^d$. Let $\bS=\set{\bbx_i}_{i=1}^n$ consist of i.i.d. covariates distributed according to an unknown law $P$, with measure $\mu$ and empirical law $P_n$. Almost surely $\bbx_i\ne\pm\bbx_j$ for $i\ne j$, since $\mu\set{\bx,-\bx}=0$. We observe responses $y_i=f^*(\bbx_i)+w_i$, where the noises are i.i.d., centered and sub-Gaussian with variance proxy $\sigma_0^2$, namely $\Expect{}{\exp\pth{\lambda w_i}}\le \exp\pth{\lambda^2\sigma_0^2/2}$ for every $\lambda\in\RR$. We write $\by=(y_i)_{i=1}^n$, $\bw=(w_i)_{i=1}^n$ and $f^*(\bS)=(f^*(\bbx_i))_{i=1}^n$ as column vectors.
We consider a two-layer NN whose channel attention matrix reweights the ReLU activations, termed the attention network in the sequel. Its mapping is
\bal\label{eq:two-layer-nn}
f(\bW,\ba,\bx) =  \frac{1}{\sqrt{m}}\sum\limits_{r'=1}^{m}
\sum\limits_{r=1}^{m}
a_r \relu{{\bbw_{r'}}^\top \bx} \bA_{r'r},
\eal
where $\sigma(z)=\max\set{z,0}$, $\bW=\set{\bbw_r}_{r=1}^m\in\RR^{m\times d}$, $\ba=(a_r)_{r=1}^m\in\RR^m$ and $\bA\in\RR^{m\times m}$. For any weights $\bW$ and sample $\bV=(\bv_i)_{i=1}^k$, the matrix $\bsigma(\bW,\bV)$ has column $i$ equal to $\bsigma(\bW,\bv_i)$, whose entry $r$ is $\sigma(\bbw_r^\top \bv_i)$. These $m$ activations are the channels. Here $\cN$ denotes a Gaussian distribution with the indicated mean and covariance. The initialization rows satisfy $\bbw_r(0)\overset{\textup{i.i.d.}}\sim\cN(\bzero,\kappa^2\bI_d)$, with $\kappa=\Theta(1)\in(0,1)$, and $\ba(0)=0$. The sample $\bQ=\set{\bbq_i}_{i=1}^N$ has the training covariate law and satisfies the independence assumptions above. An independent unlabeled sample from $P$ provides such a $\bQ$. Section~\ref{sec:Q-generation} discusses the sampling and kernel-evaluation assumptions.
For $t\ge0$, write $f_t=f(\bW(t),\ba(t),\cdot)$ and $\hat\by(t)=(f_t(\bbx_i))_{i=1}^n$, so $f_0=0$. Section~\ref{sec:training} specifies the simultaneous updates of both layers. To define the fixed matrix $\bA$ and the target class, we next describe the regular NTK and its spectral powers.

\vspace{-.1in}
\subsection{Kernel and Target Function for Nonparametric Regression}
\label{sec:setup-kernels-target-function}

We begin with the regular NTK defined as
\bal\label{eq:kernel-two-layer}
K(\bu,\bv) \defeq \kappa^2\cdot \frac{\bu^{\top}\bv\pth{\pi -\arccos (\bu^{\top}\bv)} + \sqrt{1-(\bu^{\top}\bv)^2} }{2\pi}, \quad \forall \bu, \bv \in \cX,
\eal
which is the ReLU feature covariance of (\ref{eq:two-layer-nn}) and the  NTK of the vanilla network
\bal\label{eq:vanilla-nn}
f^{\textup{(vanilla)}}(\bW,\ba,\bx) = \frac{1}{\sqrt{m}} \sum_{r=1}^{m}
a_r \relu{{\bbw_{r}}^\top \bx},
\eal
with $\ba(0)=0$ in (\ref{eq:vanilla-nn}). It can be verified that the regular NTK, $K$ in (\ref{eq:kernel-two-layer}), is a positive-definite (PD) kernel. 
We write $\bK=(K(\bbx_i,\bbx_j))_{i,j\le n}$, $\bK_n=\bK/n=\bU\bSigma\bU^\top$, with eigenvalues $\hlambda_i$ in nonincreasing order. The strict nonsingularity of the sample Gram matrix $\bK$ is proved separately under the non-antipodal condition. Let $\cH_K$ be the  RKHS associated with $K$.
The operator $T_Kf(\bx)=\int_{\cX}K(\bx,\bx')f(\bx')\,d\mu(\bx')$ is positive, self-adjoint and compact. List its positive eigenvalues with multiplicity as $b\ge\lambda_1\ge\lambda_2\ge\cdots>0$. The associated eigenfunctions $(e_j)$ form an orthonormal basis of $\overline{\operatorname{ran}(T_K)}$ in $L^2(\mu)$, excluding any null space, and the eigenfunctions have 
the continuous representatives $e_j=\lambda_j^{-1}T_Ke_j$. Mercer expansion is on the compact support $\cX$, where $\mu$ has full support. The functions $v_j=\sqrt{\lambda_j}e_j$ form an orthonormal basis of $\cH_K$, and
\bals
\cH_K(\mu_0)=\set{f=\sum_{j=1}^{\infty}\beta_je_j:\sum_{j=1}^{\infty}\frac{\beta_j^2}{\lambda_j}\le\mu_0^2}.
\eals
Spectral-bias studies~\cite{rahaman19a-spectral-bias,CaoFWZG21-spectral-bias} motivate stronger control of coefficients in eigenspaces with smaller eigenvalues. We impose this regularity through the attention kernel, whose spectral powers determine both the target ball and its regression risk.

We now construct the attention kernel from $K$ and then identify its target ball.
For every $s\ge1$, we first define the kernel $K^{(s)}$ as
\bal\label{eq:Ks-def}
K^{(s)}(\bx,\bx') \defeq \sum\limits_{j=1}^{\infty} \lambda_j^{s} e_j(\bx)e_j(\bx'),
\quad \forall \bx,\bx' \in \cX.
\eal
$K^{(s)}$ is a PD kernel. Theorem~\ref{theorem:Ks-bounded} proves uniform convergence on $\cX\times\cX$ and $\sup_{\bx,\bx'}\abth{K^{(s)}(\bx,\bx')}\le1/2$ for $s\ge1$. Define $\bK^{(s)}=(K^{(s)}(\bbq_i,\bbq_j))_{i,j\le N}$ and $\bK_N^{(s)}=\bK^{(s)}/N$.
The following integral kernel, $\Kint$, is induced by $K^{(s)}$ as follows:
\bals
\Kint(\bx,\bx') \defeq
 \int_{\cX \times \cX} K^{(s)}(\bx,\bv) K(\bv,\bv')  K^{(s)}(\bv',\bx')  \diff \mu(\bv) \otimes \mu(\bv'). 
\eals
Composing this integral kernel with $K$ on both sides gives the attention kernel $\Kattn$ and its empirical version $\hKattn$:
\bal
\Kattn (\bx,\bx') &\defeq  \int_{\cX \times \cX} K(\bx,\bv) \Kint(\bv,\bv')  K(\bv',\bx')  \diff \mu(\bv) \otimes \mu(\bv'),  \nonumber\\
\hKattn (\bx,\bx') &\defeq \frac{1}{N^2} \sum\limits_{i,j=1}^N K(\bx,\bbq_i)  \Kint(\bbq_i,\bbq_j) K(\bbq_j,\bx'), \label{eq:hatKint-def}
\eal
The attention network initialization also involves the empirical integral kernel
\bals
\hKint(\bx,\bx')=N^{-2}\sum_{i,j=1}^N K^{(s)}(\bx,\bbq_i)K(\bbq_i,\bbq_j)K^{(s)}(\bbq_j,\bx').
\eals
Its Gram matrices on a sample are denoted by $\hbKint$, while those of $\Kint$ are denoted by $\bKint$. We note that the PGD method in~\cite{YangLi2026PGD} also uses an independent auxiliary sample $\bQ$ and evaluations of the spectral kernel $K^{(s)}$ induced by its regular NTK.

To implement these compositions in the network (\ref{eq:two-layer-nn}), we construct the channel attention matrix at initialization and hold it fixed throughout training of the attention network:
\bals
\bA = \bsigma(\bW(0),\bQ)\bK^{(s)}_N  \bsigma(\bW(0),\bQ)^{\top}/(Nm).
\eals
The same sample $\bQ$ occurs in $\bA$ and (\ref{eq:hatKint-def}), so the approximation proof must account for its repeated use. Theorem~\ref{theorem:spectrum-Kattn} identifies the population limit: $T_{\Kattn}=T_K^{s'}$ and $\lambdaattn_j=\lambda_j^{s'}$, with the same positive-eigenvalue eigenfunctions. As a result, the attention kernel specifies the interpolation ball
\bals
\bth{\cH_K}^{s'}(\mu_0)=\cH_{\Kattn}(\mu_0)
&=\set{f=\sum_{j=1}^{\infty}\beta_je_j:\sum_{j=1}^{\infty}\frac{\beta_j^2}{\lambda_j^{s'}}\le\mu_0^2}.
\eals
We assume $f^*$ lies in this interpolation ball, $f^* \in \bth{\cH_K}^{s'}(\mu_0)$,
the same type of source condition used 
in~\cite{Steinwart2012,Fischer2020-sobolev-norm-regularized-least-square,YangLi2026PGD}. Since $\lambda_j\le b<1$, we have $\bth{\cH_K}^{s'}(\mu_0) \subseteq \cH_K(\mu_0)$. Every finite sum of positive-eigenvalue eigenfunctions belongs to $\cH_{\Kattn}$. Under  spherical uniform distribution, the eigenfunctions may be chosen as spherical harmonics, so a low-degree polynomial belongs to this space when all its harmonic components have positive eigenvalues. For a general covariate law, regularity is specified by the eigenfunctions of its own integral operator.
The regression task is to estimate $f^*$ from $(\bbx_i,y_i)_{i=1}^n$ with small clean regression risk $\Expect{P}{(\hat f-f^*)^2}$. Existing kernel learning methods with early stopping balances signal bias and noise at the critical radius of the target ball~\cite{RaskuttiWY14-early-stopping-kernel-regression}. Our analysis extends this balance to simultaneous training of the finite attention network in a nontrivial manner.

\smallskip
\noindent\textbf{Empirical attention spectrum and operator notation.}
Let $\bKattn=(\Kattn(\bbx_i,\bbx_j))_{i,j\le n}$ and $\bKattn_n=\bKattn/n$. Write $\bKattn_n=\bUattn\bSigmaattn{\bUattn}^\top$, with eigenvalues $\hlambdaattn_i$ listed in nonincreasing order. Let $\hbKattn$ be the Gram matrix of $\hKattn$ on the same sample. The population attention eigenvalues are $\lambdaattn_j=\lambda_j^{s'}$. We define the fixed RKHS radius by
\bal\label{eq:B_h}
B_h=\mu_0+1+\sqrt2.
\eal
For $u,v,h$ in a real Hilbert space $\cH$, the rank-one operator $u\otimes v$ is defined by $(u\otimes v)h\defeq u\iprod{v}{h}_{\cH}$. In particular, $K(\cdot,\bx)\otimes K(\cdot,\bx)$ maps $h\in\cH_K$ to $h(\bx)K(\cdot,\bx)$ by the reproducing identity. For each input construction, with its own kernel $K$ and measure $\mu$, define $C\defeq\int K(\cdot,\bx)\otimes K(\cdot,\bx)\,d\mu(\bx)$ and $C_N\defeq N^{-1}\sum_{i=1}^{N}K(\cdot,\bbq_i)\otimes K(\cdot,\bbq_i)$. The power $C^{s-1}$ is defined by spectral calculus as follows. If $v\in\cH_K$ satisfies $Cv=\lambda v$ with $\lambda>0$, then $C^{s-1}v=\lambda^{s-1}v$. For $s>1$, the power is zero on $\ker C$, and these actions extend uniquely by linearity and continuity to a bounded positive self-adjoint operator on $\cH_K$. For $s=1$, we set $C^0=I_{\cH_K}$ on the entire space, including $\ker C$.  For a Hilbert-space operator $A$, the Hilbert--Schmidt norm is defined by $\norm{A}{\textup{HS}}^2=\sum_{j=1}^{\infty}\norm{Av_j}{\cH_K}^2$ for an orthonormal basis $(v_j)$ of $\cH_K$. In the spherical case, put $a_N(\delta)=(\sqrt{b^2-\sum_{j=1}^{\infty}\lambda_j^2}+b\sqrt{\log(1/\delta)})/\sqrt N$. Mercer's trace identity gives $\sum_{j=1}^{\infty}\lambda_j=b$, so the radicand is nonnegative. Section~\ref{sec:calibration} supplies the detailed justification and concentration bounds.

\subsection{Unbounded Input Space and Normalized Attention Network}
\label{sec:setup-unbounded}
We next extend the construction to $\cX=\RR^d$ under the same noise, independence and kernel-evaluation assumptions. To apply the bounded-input analysis, we normalize the augmented inputs and responses and assume $M_2\defeq\Expect{P}{1+\ltwonorm{\bX}^2}<\infty$. This condition makes the feature covariance trace class and the regression risk finite for the target class below. Our results for unbounded input space allow laws without a density.

For $\bx\in\RR^d$, define $\tilde \bx=(1,\bx^\top)^\top$, $q(\bx)=\ltwonorm{\tilde \bx}=\sqrt{1+\ltwonorm{\bx}^2}$ and $\phi(\bx)=\tilde \bx/q(\bx)$. The augmented input represents the bias, and normalization gives $\phi(\bx)\in\unitsphere d$ with a positive first coordinate. Let $\tmu=\phi_*\mu$ be the law of $\phi(\bX)$, defined by $\tmu(E)=\mu(\phi^{-1}(E))$ for Borel sets $E$. The network is trained on these bounded inputs. 
The augmented ReLU features have covariance kernel
\bal\label{eq:kernel-unbounded}
K(\bx,\bx')=q(\bx)q(\bx')K_0(\phi(\bx),\phi(\bx')),\qquad
K_0(\bz,\bz') \defeq \frac{\kappa^2}{2\pi}\set{\bz^\top \bz'(\pi-\arccos(\bz^\top \bz'))+\sqrt{1-(\bz^\top \bz')^2}}.
\eal
Here $\bz,\bz'\in\unitsphere d$ and $K(\bx,\bx)=bq(\bx)^2$, with $b=\kappa^2/2$. Equation~(\ref{eq:kernel-unbounded}) is the uncentered second moment of the augmented ReLU features $\sigma(\bbw^\top\tilde \bx)$ and $\sigma(\bbw^\top\tilde \bx')$ for $\bbw\sim\cN(0,\kappa^2\bI_{d+1})$.

To retain the original interpolation-space condition, we define the operator powers before normalizing the kernel. Write $K_{\bx}=K(\cdot,\bx)$ and use the covariance operator $C$ defined in Section~\ref{sec:setup-kernels-target-function}, now for the Euclidean kernel and law. Its trace is $bM_2$, and its positive eigenvalues are those of $T_K$, denoted by $(\lambda_j)_{j\ge1}$, with $\lambda_1\le bM_2$. We define the kernels by
\bal\label{eq:unbounded-powered-kernels}
K^{(s)}(\bx,\bx')=\iprod{K_{\bx}}{C^{s-1}K_{\bx'}}_{\cH_K},\qquad
\Kattn(\bx,\bx')=\iprod{K_{\bx}}{C^{2s+2}K_{\bx'}}_{\cH_K}.
\eal
We take $C^0=I_{\cH_K}$ when $s=1$. These definitions agree on the support with the spectral series and iterated integrals in Section~\ref{sec:setup-kernels-target-function}. The covariance representation makes these definitions valid on $\RR^d$. Lemma~\ref{lemma:unbounded-normalization} proves that $T_{\Kattn}=T_K^{s'}$, with $s'=2s+3$. We retain exactly the target assumption $f^*\in[\cH_K]^{s'}(\mu_0)=\cH_{\Kattn}(\mu_0)$, using its canonical pointwise representative given by (\ref{eq:unbounded-powered-kernels}). The spherical bound $\lambda_1\le b$ and the associated inclusion of equal-radius RKHS balls are not asserted in this unbounded case.

We then divide both the completed attention kernel and the target by the input norms. On the image of $\phi$, we define
\bal\label{eq:normalized-attention-kernel}
\barKattn(\phi(\bx),\phi(\bx'))=\frac{\Kattn(\bx,\bx')}{q(\bx)q(\bx')},\qquad
g^*(\phi(\bx))=\frac{f^*(\bx)}{q(\bx)}.
\eal
The transformation in (\ref{eq:normalized-attention-kernel}) is an isometry from $\cH_{\Kattn}$ to $\cH_{\barKattn}$, so it preserves the target norm: $\norm{g^*}{\cH_{\barKattn}}\le\mu_0$. Moreover, $\barKattn(\bz,\bz)\le D_0\defeq b\lambda_1^{s'-1}$, and this kernel extends continuously to the compact support of $\tmu$ in $\unitsphere d$. This construction permits bounded-input analysis while preserving the original target class, which generally differs from the class obtained by taking powers of the operator of $K_0$ under $\tmu$.

This normalization must also preserve the auxiliary weights in the attention matrix. Let $\bD$ be diagonal with $\bD_{ii}=q(\bbq_i)$ for $i\in[N]$. Define $\bK_N^{(s)}=(K^{(s)}(\bbq_i,\bbq_j))_{i,j\in[N]}/N$, and let $\bG_{\bQ}$ have columns $\bsigma(\bW(0),\phi(\bbq_i))$. The normalized attention network uses
\bal\label{eq:normalized-attention-network}
\bA=\frac1{Nm}\bG_{\bQ}\bD\bK_N^{(s)}\bD \bG_{\bQ}^\top,\qquad
f^{(\textup{nor})}(\bW,\ba,\bx)=\frac1{\sqrt m}\sum_{r'=1}^m\sum_{r=1}^m a_r\sigma(\bbw_{r'}^\top\phi(\bx))\bA_{r'r}.
\eal
The rows of $\bW(0)\in\RR^{m\times(d+1)}$ are independent $\cN(0,\kappa^2\bI_{d+1})$ vectors, and $\ba(0)=0$. We construct the matrix in (\ref{eq:normalized-attention-network}) once and hold it fixed. Both layers are trained by the simultaneous GD updates of Section~\ref{sec:training}, with inputs $\phi(\bbx_i)$, responses $y_i/q(\bbx_i)$, and this matrix $\bA$. Equivalently, the loss is
\bal\label{eq:normalized-attention-loss}
L^{(\textup{nor})}(\bW,\ba)=\frac1{2n}\sum_{i=1}^n\left(f^{(\textup{nor})}(\bW,\ba,\bbx_i)-\frac{y_i}{q(\bbx_i)}\right)^2,
\qquad f_t(\bx)=q(\bx)f_t^{(\textup{nor})}(\bx).
\eal
We use $f_t^{(\textup{nor})}(\bx)=f^{(\textup{nor})}(\bW(t),\ba(t),\bx)$ in (\ref{eq:normalized-attention-loss}). Conditional on the training design, the normalized noises $w_i/q(\bbx_i)$ are independent, centered and sub-Gaussian with variance proxies at most $\sigma_0^2$. They need not be independent of that design. Every noise estimate below is applied conditionally on the design.

The transfer to the original regression risk depends on the covariate tail and its weighted second moment. We define these quantities and the auxiliary-covariance approximation bound by
\bal\label{eq:unbounded-tail-definitions}
\Phi(R)&=\mu\set{\bx:\ltwonorm{\bx}>R},
\Psi(R)=\Expect{P}{q(\bX)^2\indict{\ltwonorm{\bX}>R}},
a_{N,\Phi}(\delta)&=\inf_{R\ge1}\set{(1+R^2)\sqrt{\frac{\log(8/\delta)}N}+\frac{\Psi(R)}\delta}.
\eal
Here $R\ge0$ in the definitions of $\Phi$ and $\Psi$, and $\delta\in(0,1)$. The identity $\Psi(R)=(1+R^2)\Phi(R)+\int_R^\infty2u\Phi(u)\,du$ expresses the weighted tail in terms of $\Phi$. The quantity $a_{N,\Phi}(\delta)$ combines concentration on a bounded ball with its omitted weighted mass. Since $M_2<\infty$, we have $\Psi(R)\to0$ and $a_{N,\Phi}(\delta)\to0$ as $N\to\infty$ for fixed $\delta$. Lemma~\ref{lemma:unbounded-covariance} justifies its use in (\ref{eq:unbounded-tail-definitions}) in place of the spherical covariance error $a_N(\delta)$.

\subsection{Kernel Complexity}
\label{sec:kernel-complexity}
The stopping rule must balance decreasing signal bias against the noise accumulated during training. Local kernel complexity measures the contribution of the kernel spectrum at a given scale~\cite{bartlett2005,koltchinskii2006,Mendelson02-geometric-kernel-machine}. For a trace-class kernel with nonzero population and empirical spectra, we define the empirical and population complexities by
\bals
\hat R_K(\eps)=\sqrt{\frac1n\sum_{i=1}^n\min\set{\hat\lambda_i,\eps^2}},\qquad
R_K(\eps)=\sqrt{\frac1n\sum_{j=1}^{\infty}\min\set{\lambda_j,\eps^2}}.
\eals
The critical radii $\hat\eps_{K,n}$ and $\eps_{K,n}$ are the smallest positive solutions of $\hat R_K(\eps)\le\eps^2/\sigma_0$ and $R_K(\eps)\le\eps^2/\sigma_0$, respectively. A sub-root function is nonnegative and nondecreasing, and its ratio to the square root of its positive argument is nonincreasing. Definition~\ref{def:sub-root-function} gives the formal definition. The squared radii are the fixed points of the corresponding sub-root functions of $\eps^2$~\cite{bartlett2005,koltchinskii2006,Mendelson02-geometric-kernel-machine}. For the spherical attention kernel, write $\eps_n=\eps_{\Kattn,n}$ for the population radius and $\hat\eps_n=\hat\eps_{\Kattn,n}$ for the empirical radius. Proposition~\ref{prop:infinite-spectrum-radius} gives infinite positive spectra under non-atomicity, hence $\eps_n\to0$ and $n\eps_n^2\to\infty$, also for $\eps_{K,n}$. Section~\ref{sec:results-unbounded-input-space} uses the same definitions for $\barKattn$ under $\tmu$.
The empirical radius determines the maximum number of training steps. For learning rate $\eta$, we define the stopping time by
\bal\label{eq:stopping-time-hatT}
\hat T\defeq\max\bigl(\set{0}\cup\set{t\in\set{1,2,\ldots}\colon \eta t\,\hat\eps_n^2\le1}\bigr).
\eal
For sufficiently large $n$, (\ref{eq:stopping-time-hatT}) gives a positive stopping time, with value zero otherwise when its positive-time set is empty. It is equivalent to the first-crossing rule because $\hat R_{\Kattn}(\eps)/\eps^2$ strictly decreases. Algorithm~\ref{alg:GD} uses $T\le\hat T$ to keep the training trajectory within this statistical time scale.

\vspace{-.05in}
\section{Main Results}
\label{sec:main-results}
\vspace{-.05in}
We work at fixed dimension $d\ge5$. The first two theorems concern the spherical construction under an arbitrary non-atomic law $P$ on $\cX$. The first gives regression risk in terms of the attention kernel's critical radius without prescribing an eigenvalue decay rate. The second makes this bound explicit under polynomial decay. Section~\ref{sec:results-unbounded-input-space} then transfers the bounded-input argument to $\RR^d$.

\subsection{Spherical Input Space}
The first theorem connects the critical-radius stopping rule to simultaneous training of the finite network. Its width and auxiliary-sample conditions keep the approximation errors at the statistical scale.
\begin{theorem}
\label{theorem:minimax-nonparametric-regression-Kattn-general}
Let $f^*\in\cH_{\Kattn}(\mu_0)$ and let $n$ be sufficiently large for the fixed covariate law and fixed model parameters.
Suppose $P$ is an arbitrary non-atomic distribution on $\cX$, $\delta \in (0,1)$,
\bal\label{eq:m-N-cond-LRC-population-NN-general}
m \gsim \max\set{d\log m/\eps_n^6,n^{1/(2d)}},
\quad
N \gsim \log (2/{\delta})/\eps_n^6,
\eal
and the neural network
$f_t = f(\bW(t),\ba(t),\cdot)$ is
trained by GD in Algorithm~\ref{alg:GD}
with the learning rate $\eta = \Theta(1) \in (0,1)$ with
$T=\hat T\ge1$.
Fix $c_t\in(0,1)$. With probability at least
$1-C/n-\delta-C\exp(-cn)-C\exp(-cn\eps_n^2)$
over $\bw,\bS,\bQ,\bW(0)$, the stopping time satisfies
$\hat T\asymp\eps_n^{-2}$ and, simultaneously for every $t\in[\lceil c_tT\rceil,T]$,
\bal\label{eq:minimax-nonparametric-regression-Kattn-general}
\Expect{P}{(f_t-f^*)^2} \lsim \eps_n^{2}.
\eal
\end{theorem}

\smallskip
\noindent\textbf{Technical roadmap for the general result.}
The proof is based on the following steps. First, Lemma~\ref{lemma:hat-eps-eps-relation} compares the empirical and population critical radii, giving $T=\hat T\asymp\eps_n^{-2}$. Theorems~\ref{theorem:good-random-initialization} and~\ref{theorem:hatKattn-close-to-Kattn-supnorm}, together with Lemma~\ref{lemma:empirical-loss-convergence-contraction-bKattn-approx}, then control the initialization kernel using uniform feature and auxiliary-covariance estimates. Next, Lemma~\ref{lemma:early-stopped-residual-recursion} bounds both trained layers and the filtered recursion errors. The resulting pointwise approximation gives the RKHS decomposition in Theorem~\ref{theorem:bounded-NN-class}. Finally, Theorem~\ref{theorem:empirical-loss-bound} bounds the clean empirical error, and Theorem~\ref{theorem:LRC-population-NN-eigenvalue} gives the population comparison with approximation contribution $Cw^2$. Taking $w=\eps_n$ proves (\ref{eq:minimax-nonparametric-regression-Kattn-general}). The covariance estimate is uniform over the kernel arguments, giving the factor $\log(2/\delta)$ in the auxiliary requirement.
The stopping interval ensures that signal bias has reached the critical-radius scale. Proposition~\ref{prop:infinite-spectrum-radius} justifies the sufficiently-large-$n$ qualification for each fixed law.

We next specialize the general theorem to a polynomial eigenvalue decay rate (EDR), $\lambda_j\asymp j^{-2\alpha}$ with $\alpha>1/2$. Here the eigenvalues and the exponent $\alpha$ refer to the ReLU feature covariance kernel (\ref{eq:kernel-two-layer}). Decay estimates for a derivative NTK concern a different kernel. Both proofs are given in Section~\ref{sec:proofs-main-results}.
\begin{theorem}
\label{theorem:minimax-nonparametric-regression-Kattn-poly-edr}
Let $f^*\in\cH_{\Kattn}(\mu_0)$ and let $n$ be sufficiently large for the fixed covariate law and fixed model parameters.
Suppose the distribution $P$ admits a polynomial
EDR of $\lambda_j \asymp j^{-2\alpha}$ for
$\alpha > 1/2$ and all $j \ge 1$, $\delta \in (0,1)$,
\bal\label{eq:m-N-cond-LRC-population-NN-poly-edr}
m \gsim n^{6\alpha s'/(2\alpha s'+1)} d\log m,
\quad
N \gsim n^{\frac{6\alpha s'}{2\alpha s'+1}} \log (2/{\delta}),
\eal
and the neural network
$f_t = f(\bW(t),\ba(t),\cdot)$ is
trained by GD in Algorithm~\ref{alg:GD}
with the learning rate $\eta = \Theta(1) \in (0,1)$ with
$T=\hat T\ge1$.
Fix $c_t\in(0,1)$. With probability at least
$1-C/n-\delta-C\exp(-cn)-C\exp(-cn^{1/(2\alpha s'+1)})$
over $\bw,\bS,\bQ,\bW(0)$, the stopping time satisfies
$\hat T\asymp n^{\frac{2\alpha s'}{2\alpha s'+1}}$ and, simultaneously for every $t\in[\lceil c_tT\rceil,T]$,
\bals
\Expect{P}{(f_t-f^*)^2} \lsim n^{-\frac{2\alpha s'}{2\alpha s'+1}}.
\eals
\end{theorem}

\smallskip
\noindent\textbf{Technical roadmap for the polynomial-spectrum result.}
Theorem~\ref{theorem:spectrum-Kattn} first gives $\lambdaattn_j=\lambda_j^{s'}\asymp j^{-2\alpha s'}$. We split the complexity sum where $\lambdaattn_j$ crosses $\eps^2$. The head and tail each have order $\eps^2\eps^{-1/(\alpha s')}$. Solving $\sigma_0R_{\Kattn}(\eps)=\eps^2$ therefore gives $\eps_n^2\asymp n^{-2\alpha s'/(2\alpha s'+1)}$. Substituting this expression into the general theorem yields the regression risk, stopping and size bounds.

\subsection{Unbounded Input Space}
\label{sec:results-unbounded-input-space}
On $\RR^d$, the normalized estimator has bounded inputs, but the desired regression risk concerns the original responses. We first apply the bounded-input argument to $\barKattn$ under $\tmu$ and then restore the factor $q$. Define $\eps_{0,n}=\eps_{\barKattn,n}$ and $\hat\eps_{0,n}=\hat\eps_{\barKattn,n}$ by Section~\ref{sec:kernel-complexity}, with $\barKattn$ and $\tmu$ in place of $\Kattn$ and $\mu$. The index $0$ distinguishes these radii from the spherical ones. The corresponding stopping time is
\bal\label{eq:normalized-stopping-time}
\hat T_0=\max\bigl(\set{0}\cup\set{t\in\set{1,2,\ldots}:\eta t\hat\eps_{0,n}^2\le1}\bigr).
\eal
The kernel in (\ref{eq:normalized-attention-kernel}) has infinite positive spectrum under non-atomicity, as proved in Lemma~\ref{lemma:unbounded-normalization}. Thus $\eps_{0,n}\to0$ and $n\eps_{0,n}^2\to\infty$. We use a fixed sufficiently small learning rate $\eta=\Theta(1)\in(0,\eta_0)$, where $\eta_0>0$ depends only on the fixed law and model parameters. Unlike the spherical case, the diagonal bound $D_0$ need not be at most $1$. Lemma~\ref{lemma:normalized-training} specifies how to choose $\eta_0$.

\begin{theorem}\label{theorem:regression-unbounded}
Let $\cX=\RR^d$, let $P$ be non-atomic with $M_2<\infty$, and let $f^*\in\cH_{\Kattn}(\mu_0)$ for the kernel in (\ref{eq:unbounded-powered-kernels}). Fix $\delta\in(0,1)$ and $c_t\in(0,1)$. There is a finite integer $n_0$, depending only on the fixed covariate law and model parameters, such that the following holds for every integer $n\ge n_0$. Let $N_0(P,n,\delta)$ be the finite deterministic auxiliary-sample threshold defined in Lemma~\ref{lemma:unbounded-covariance} at tolerance $c\eps_{0,n}^3$, with a sufficiently small fixed $c>0$. Choose deterministic positive integers $m,N$ satisfying
\bal\label{eq:unbounded-network-sizes}
m\gsim\max\set{(d+1)\log m\,\eps_{0,n}^{-6},n^{1/(2(d+1))}},\qquad N\ge N_0(P,n,\delta).
\eal
The width constant is sufficiently large. The threshold $N_0$ may also depend on the fixed model parameters, but all sample sizes are chosen before drawing the training data, initialization and auxiliary sample. The integer $n_0$ is independent of $m,N,\delta$. We take $n_0\ge2$ with $\eta\sigma_0\sqrt{D_0/n_0}\le1$ and, for all $n\ge n_0$, $\eps_{0,n}<1$ and $c\eps_{0,n}^3\le1$. Train (\ref{eq:normalized-attention-network}) by (\ref{eq:normalized-attention-loss}), with the fixed learning rate specified above and $T=\hat T_0\ge1$ from (\ref{eq:normalized-stopping-time}). With probability at least $1-C/n-\delta-C\exp\pth{-cn}-C\exp\pth{-cn\eps_{0,n}^2}$ over $\bW(0),\bw,\bS,\bQ$, we have $\hat T_0\asymp\eps_{0,n}^{-2}$ and, simultaneously for all $t\in[\lceil c_tT\rceil,T]$,
\bal\label{eq:normalized-regression-risk}
\Expect{P}{\left(f_t^{(\textup{nor})}(\bX)-\frac{f^*(\bX)}{q(\bX)}\right)^2}\lsim\eps_{0,n}^2,\qquad
\sup_{\bx\in\operatorname{supp}(\mu)}\abth{f_t^{(\textup{nor})}(\bx)-\frac{f^*(\bx)}{q(\bx)}}\le B.
\eal
Here $B=(B_h+\mu_0)\sqrt{D_0}+1$ is fixed independently of $n,m,N,T,\delta$. On the same event, for every $R\ge0$ and every such $t$, the unnormalized estimator satisfies
\bal\label{eq:unbounded-regression-tail}
\Expect{P}{(f_t-f^*)^2}\le C(1+R^2)\eps_{0,n}^2+B^2\Psi(R).
\eal
\end{theorem}

\smallskip
\noindent\textbf{Technical roadmap for the unbounded-input result.}
First, Lemma~\ref{lemma:unbounded-normalization} identifies the RKHS isometry under division by $q$. Lemma~\ref{lemma:unbounded-covariance} then proves that $N_0(P,n,\delta)$ is finite under $M_2<\infty$, with a sufficient choice determined by truncation and $\Psi$. Lemma~\ref{lemma:normalized-initialization} converts this covariance accuracy into uniform kernel approximation for the attention matrix with weights $\bD$. Next, Lemma~\ref{lemma:normalized-training} verifies the matrix and row bounds for the joint induction and the conditional concentration bounds for the normalized noise. The empirical error and local-complexity estimates give (\ref{eq:normalized-regression-risk}), whose uniform bound yields the tail estimate (\ref{eq:unbounded-regression-tail}). Appendix~\ref{sec:proofs-unbounded} proves these results for the normalized completed kernel. This kernel generally differs from a power of $K_0$ under $\tmu$.
Under a finite second moment, the threshold $N_0(P,n,\delta)$ depends on the covariate tail and need not admit a uniform polynomial bound. The following corollary gives explicit auxiliary-sample and regression risk bounds under a sub-Gaussian tail. Direct covariance concentration gives the auxiliary requirement, while restoring the factor $q^2$ introduces the logarithm in the regression risk.
\begin{corollary}\label{corollary:regression-unbounded-subgaussian}
Retain the distributional, target, initialization and training assumptions of Theorem~\ref{theorem:regression-unbounded}, and suppose $\Phi(R)\le2\exp\pth{-aR^2}$ for every $R\ge0$, with a fixed $a>0$. For every $n\ge n_0$ and $\delta\in(0,1)$, its threshold satisfies $N_0(P,n,\delta)\lsim\eps_{0,n}^{-6}\log(2/\delta)$. Thus the auxiliary-sample requirement in (\ref{eq:unbounded-network-sizes}) is satisfied by deterministic $N$ with
\bal\label{eq:unbounded-subgaussian-N}
N\gsim\eps_{0,n}^{-6}\log(2/\delta).
\eal
Under this sufficient condition and the width condition of that theorem, on its probability event and for all $t\in[\lceil c_tT\rceil,T]$, with $\alpha=(d+3)/(2d)>1/2$,
\bal\label{eq:unbounded-subgaussian-risk}
\Expect{P}{(f_t-f^*)^2}\le C(\log n)\eps_{0,n}^2\le C\,(\log n)n^{-\frac{2\alpha s'}{2\alpha s'+1}}=C\,(\log n)n^{-\frac{s'(d+3)}{s'(d+3)+d}}.
\eal
\end{corollary}

\smallskip
\noindent\textbf{Technical roadmap for the sub-Gaussian result.}
Lemma~\ref{lemma:unbounded-covariance} gives joint covariance and empirical-second-moment error of order $\sqrt{\log(2/\delta)/N}+\log(2/\delta)/N$, which yields (\ref{eq:unbounded-subgaussian-N}). The tail identity gives $\Psi(R)\le2(1+R^2+a^{-1})\exp\pth{-aR^2}$. Taking $R^2=3a^{-1}\log n$ in (\ref{eq:unbounded-regression-tail}) proves the first inequality in (\ref{eq:unbounded-subgaussian-risk}), using $n\eps_{0,n}^2\to\infty$. Finally, Lemma~\ref{lemma:euclidean-base-eigenvalue-upper} gives $\lambda_j\lesssim j^{-(d+3)/d}$ under $M_2<\infty$, and Proposition~\ref{prop:normalized-eigenvalue-upper} gives $\eps_{0,n}^2\lesssim n^{-s'(d+3)/(s'(d+3)+d)}$. These one-sided estimates require no density assumption and establish an upper regression risk bound.

\smallskip
\noindent\textbf{Finite second moment and tail dependence.}
The same eigenvalue upper bound also gives a useful conclusion before imposing a sub-Gaussian tail. Under the assumptions and size conditions of Theorem~\ref{theorem:regression-unbounded}, we obtain, for every $R\ge0$,
\bal\label{eq:unbounded-polynomial-tail-risk}
\Expect{P}{(f_t-f^*)^2}\le C(1+R^2)n^{-\frac{2\alpha s'}{2\alpha s'+1}}+B^2\Psi(R),\qquad \alpha=(d+3)/(2d).
\eal
A finite second moment gives $\Psi(R)\to0$ in (\ref{eq:unbounded-polynomial-tail-risk}), while the sub-Gaussian tail supplies the logarithmic specialization. Since the spectral estimate is one-sided, the width and stopping conditions remain expressed in terms of the actual normalized critical radius.

\smallskip
\noindent\textbf{Comparison with existing regression risk bounds.}
Table~\ref{table:regression-risk-comparison} compares the regression risk bounds under the respective initialization, width, noise and training assumptions. Each row specifies its kernel spectrum, target ball, covariate assumptions and probability formulation. We use $\tK$ for the NTK of the cited architecture and $\eps_{\tK,n}$ for its critical radius. The result in~\cite[Proposition~13]{Li2024-edr-general-domain} is specialized to interpolation exponent $s'=2s+3>1/(d+1)$.

\begin{table}[!htbp]
\centering
\caption{Comparison with existing regression risk bounds. The positive spectrum and target class belong to the kernel identified in each row. All target classes have a fixed norm bound.}
\label{table:regression-risk-comparison}
\begingroup
\footnotesize
\setlength{\tabcolsep}{5pt}
\renewcommand{\arraystretch}{1.0}
\begin{tabular}{@{}>{\raggedright\arraybackslash}p{.22\linewidth}>{\raggedright\arraybackslash}p{.32\linewidth}>{\raggedright\arraybackslash}p{.20\linewidth}>{\raggedright\arraybackslash}p{.20\linewidth}@{}}
\toprule
Existing works and our results & Distribution and target assumptions & Positive spectrum & Regression risk\\
\midrule
\multicolumn{4}{c}{\textbf{Spherical input space $\cX\subseteq\unitsphere{d-1}$}}\\
\midrule
\cite[Theorem~5.2]{HuWLC21-regularization-minimax-uniform-spherical} and \cite[Theorem~3.11]{SuhKH22-overparameterized-gd-minimax}
& Uniform distribution on $\unitsphere{d-1}$, $f^*\in\cH_{\tK}(\mu_0)$
& $\tK$: $j^{-d/(d-1)}$
& Order $n^{-d/(2d-1)}$ in probability\\
\addlinespace[2pt]
\cite[Theorem~5.1]{Yang2025-generalization-two-layer-regression}
& Arbitrary continuous distribution on the sphere, $f^*\in\cH_{\tK}(\mu_0)$
& No EDR assumption
& $\cO(\eps_{\tK,n}^2)$\\
\addlinespace[2pt]
\cite[Corollary~5.2]{Yang2025-generalization-two-layer-regression}
& Spherical covariates, $f^*\in\cH_{\tK}(\mu_0)$, polynomial EDR with $\alpha>1/2$
& $\tK$: $j^{-2\alpha}$
& $\cO(n^{-\frac{2\alpha}{2\alpha+1}})$\\
\addlinespace[2pt]
Theorem~\ref{theorem:minimax-nonparametric-regression-Kattn-general}
& Arbitrary non-atomic spherical $P$ on $\cX$, $f^*\in\cH_{\Kattn}(\mu_0)$
& No EDR assumption
& $\cO(\eps_n^2)$\\
\addlinespace[2pt]
Theorem~\ref{theorem:minimax-nonparametric-regression-Kattn-poly-edr}
& Non-atomic spherical $P$ on $\cX$, $f^*\in\cH_{\Kattn}(\mu_0)$, polynomial EDR with $\alpha>1/2$
& $K$: $j^{-2\alpha}$\par $\Kattn$: $j^{-2\alpha s'}$
& $\cO(n^{-\frac{2\alpha s'}{2\alpha s'+1}})$\\
\midrule
\multicolumn{4}{c}{\textbf{Unbounded input space $\cX=\RR^d$}}\\
\midrule
\cite[Proposition~13]{Li2024-edr-general-domain}
& Sub-Gaussian design on $\RR^d$ with the density condition in the text, bounded $f^*\in[\cH_{\tK}]^{s'}(\mu_0)$
& $\tK$: $j^{-(d+1)/d}$
& $\cO(n^{-\frac{s'(d+1)}{s'(d+1)+d}})$\par $\times\log^2(12/\delta)$\\
\addlinespace[2pt]
Theorem~\ref{theorem:regression-unbounded}
& Non-atomic $P$ on $\RR^d$, $M_2<\infty$, $f^*\in[\cH_K]^{s'}(\mu_0)$
& No EDR assumption\par No density assumption
& $C(1+R^2)\eps_{0,n}^2$\par $+B^2\Psi(R)$, $R\ge0$\\
\addlinespace[2pt]
Corollary~\ref{corollary:regression-unbounded-subgaussian}
& Non-atomic sub-Gaussian $P$ on $\RR^d$, $f^*\in[\cH_K]^{s'}(\mu_0)$
& $K$: $\lambda_j\lesssim j^{-(d+3)/d}$\par No density assumption
& $\cO((\log n)n^{-\frac{s'(d+3)}{s'(d+3)+d}})$\\
\addlinespace[2pt]
Proposition~\ref{prop:vanilla-bias-regression}\par Biased vanilla network
& Non-atomic sub-Gaussian $P$ on $\RR^d$, $f^*\in[\cH_K]^{s'}(\mu_0)$\par Joint gradient flow, $\ba(0)=0$
& $K$: $\lambda_j\lesssim j^{-(d+3)/d}$
& $\cO(n^{-\frac{s'(d+3)}{s'(d+3)+d}})$\par $\times\log^2(12/\delta)$\\
\bottomrule
\end{tabular}
\endgroup
\end{table}

\smallskip
\noindent\textbf{Comparison with Vanilla GD~\cite{Li2024-edr-general-domain}.}
The authors of~\cite[Section~3 and Proposition~13]{Li2024-edr-general-domain} use a sub-Gaussian design with a bounded Riemann-integrable density $p$, early-stopped gradient flow and width polynomial in $n$, $\log(1/\delta)$ and the reciprocal minimum empirical NTK eigenvalue. The proof of~\cite[Proposition~13]{Li2024-edr-general-domain} requires the density bound $p(\bx)\lsim (1+\ltwonorm{\bx}^2)^{-(d+3)/2}$ to invoke the eigenvalue estimate in~\cite[Theorem~10]{Li2024-edr-general-domain}. Corollary~\ref{corollary:regression-unbounded-subgaussian} requires no density condition on the non-atomic sub-Gaussian law $P$, which need not admit a density. Since their NTK differs from $K$, we next compare on our source ball. Appendix~\ref{sec:proof-vanilla-bias} verifies the concentration and finite-width estimates for the biased vanilla network
\bal\label{eq:vanilla-bias-network}
f^{\textup{(vanilla,bias)}}(\bW,\ba,\bx)=\frac1{\sqrt m}\sum_{r=1}^m a_r\relu{\bbw_r^\top\tilde\bx},\qquad \bx\in\RR^d.
\eal
Here $\tilde\bx=(1,\bx^\top)^\top$, $\ba(0)=0$ and $\bbw_r(0)\overset{\textup{i.i.d.}}\sim\cN(0,\kappa^2\bI_{d+1})$. Both layers follow the negative Clarke subgradient inclusion for the loss $(2n)^{-1}\sum_{i=1}^n(f^{\textup{(vanilla,bias)}}(\bW,\ba,\bbx_i)-y_i)^2$. The solution is absolutely continuous, with measurable selection and chain rule justified in that appendix. Its limiting initialization NTK is (\ref{eq:kernel-unbounded}). Write $f_t^{\textup{(vanilla,bias)}}$ for its output at continuous time $t\ge0$ and $\lambda_0=\lambda_{\min}((K(\bbx_i,\bbx_j))_{i,j\le n})>0$ almost surely by Lemma~\ref{lemma:unbounded-normalization}.
Proposition~\ref{prop:vanilla-bias-regression} uses the kernel regression risk and network approximation strategy of~\cite[Proposition~13, Appendix~C]{Li2024-edr-general-domain}. Its uniform approximation after division by $q$ gives an $L^2(P)$ bound under $M_2<\infty$. Appendix~\ref{sec:proof-vanilla-bias} supplies the concentration and coupling estimates required by this architecture and zero output initialization. Under the same sub-Gaussian tail, noise and source ball $[\cH_K]^{s'}(\mu_0)$, for fixed $\delta\in(0,1)$ and deterministic width $m$, it gives an event of probability at least $1-\delta$ on which the polynomial condition (\ref{eq:vanilla-comparison-width}) implies, for sufficiently large $n$ at $t_{\rm op}\asymp n^{(d+3)/(s'(d+3)+d)}$,
\bal\label{eq:vanilla-li-regression-risk}
\Expect{P}{(f_{t_{\rm op}}^{\textup{(vanilla,bias)}}-f^*)^2}\le C\log^2(12/\delta)\,n^{-\frac{s'(d+3)}{s'(d+3)+d}}.
\eal
The proof combines the source-condition bias bound, conditional filtered-noise concentration and uniform control of both layers during gradient flow. The source condition provides the required linear-growth target bound, so the result holds without a density condition or a separate bounded-target assumption.
The polynomial exponent in (\ref{eq:vanilla-li-regression-risk}) equals that of Corollary~\ref{corollary:regression-unbounded-subgaussian}. For fixed $\delta$, its confidence factor is constant in $n$, whereas $\delta=1/n$ gives $\cO(\log^2n)$. The attention bound has a factor $\log n$, and comparison at failure probability $\cO(1/n)$ requires $n\eps_{0,n}^2\gtrsim\log n$ with a sufficiently large constant. Its sufficient width and auxiliary-sample conditions are explicit and independent of the reciprocal minimum empirical eigenvalue, under the stated kernel-evaluation and auxiliary-sampling assumptions.

\smallskip
\noindent\textbf{Comparison under identical assumptions.}
For a spherical law with $\lambda_j\asymp j^{-2\alpha}$, fix the stated sub-Gaussian noise, target ball $\cH_{\Kattn}(\mu_0)$ and exact kernel evaluations. Proposition~\ref{prop:kernel-baseline} gives kernel GD using the labeled data, kernel evaluations, noise proxy and stopping rule (\ref{eq:stopping-time-hatT}). Both estimators satisfy simultaneous high-probability regression risk bounds of order $n^{-2\alpha s'/(2\alpha s'+1)}$ on the stated stopping interval. The network guarantee additionally requires (\ref{eq:m-N-cond-LRC-population-NN-poly-edr}).
Proposition~\ref{prop:sharper-eps-n-eps-Kn} also gives $\eps_n\le\eps_{K,n}$, where the larger ball $\cH_K(\mu_0)$ has critical squared radius $n^{-2\alpha/(2\alpha+1)}$. The improvement over this unspecialized guarantee reflects the stronger source condition. It establishes neither a lower bound for vanilla networks on the smaller ball nor a statistical advantage over kernel estimators adapted to that source condition.

\subsection[Comparison with PGD]{Comparison with PGD~\cite{YangLi2026PGD}}
\label{sec:comparison-pgd}

We compare our training analysis with the PGD results of~\cite{YangLi2026PGD}, which study interpolation-space regression using an over-parameterized NN using Preconditioned Gradient Descent (PGD) instead of channel attention. This subsection concerns the spherical construction and expresses the size conditions in terms of the stopping time. Section~\ref{sec:results-unbounded-input-space} gives the Euclidean results, including Corollary~\ref{corollary:regression-unbounded-subgaussian}, which requires no density condition in contrast to~\cite[Section~3 and Proposition~13]{Li2024-edr-general-domain}.

\smallskip
\noindent\textbf{Relation between the results.}
Both analyses use an independent auxiliary sample $\bQ$, an interpolation-space target and early stopping. We compare the decomposition of the trained functions and the sufficient conditions on $m,N$ needed for the corresponding training guarantees.

\smallskip
PGD trains only the first layer with fixed output signs, whereas Algorithm~\ref{alg:GD} trains both layers from $\ba(0)=0$ with $\bA$ fixed. The base kernel in~\cite[Eq.~(2)]{YangLi2026PGD} is the derivative NTK, and our kernel (\ref{eq:kernel-two-layer}) is the ReLU feature covariance. Their induced operators have powers $s+2$ and $s'=2s+3$, respectively. The resulting source balls depend on these different base kernels, so the powers alone do not compare risks on identical targets.

\smallskip
\noindent\textbf{Joint control of the two trained layers.}
Simultaneous training introduces an interaction between changing output weights and moving features that is absent from the fixed-output-weight PGD theorem. Lemma~\ref{lemma:early-stopped-residual-recursion} controls the actual residuals, both layers and their interaction on one event throughout $0\le t\le T\le\hat T$. Its bounds in (\ref{eq:early-stopped-weight-bounds}) give $\ltwonorm{\ba(t)}\lesssim T^{1/4}$ and first-layer motion of order $(d+\log(2mn))T^{5/4}/m$. To describe this interaction, let $F_k=\bsigma(\bW(k),\bS)$ and $\bu(k)=\hat\by(k)-\by$ for $0\le k<T$. The simultaneous updates satisfy
\bal\label{eq:comparison-pgd-update}
\hat\by(k+1)-\hat\by(k)
&=-\frac\eta{nm}F_k^\top\bA^2F_k\bu(k)
  +\frac1{\sqrt m}(F_{k+1}-F_k)^\top\bA\ba(k+1).
\eal
The second term in (\ref{eq:comparison-pgd-update}) couples the feature increment with the new output weights. The induction therefore first bounds the increment using $\ba(k)$, then bounds $\ba(k+1)$ using the preceding recursion errors, and finally estimates the next residual. This order yields simultaneous control of both trained layers.

\smallskip
The fixed symmetry of $\bA$ also gives a preconditioner interpretation: the output coefficients $\bA\ba$ follow the vanilla output-gradient update multiplied by $\bA^2$, while the first layer follows the vanilla gradient at these coefficients.

\smallskip
\noindent\textbf{Spectral filtering of the actual recursion errors.}
We next bound the RKHS contribution of accumulated recursion errors. Let $0<\tau\le1$ be the tolerance for an accumulated error of Euclidean norm at most $\sqrt n\tau$. The PGD estimate~\cite[Lemma~B.5, Eq.~(86)]{YangLi2026PGD} bounds these vectors separately, giving $C\eta T\tau$ and requiring $\tau\lesssim1/(\eta T)$. We instead retain the filtering generated by the actual updates. Let $\bE(j)$ be the one-step errors in (\ref{eq:early-stopped-one-step-error}), with $\bbe_2(0)=0$ and $\bbe_2(j)=\sum_{i=1}^j(I-\eta\bKattn_n)^{j-i}\bE(i)$ their accumulated contribution. The finite geometric sum in (\ref{eq:early-stopped-filtered-error}) retains this structure. Under the proved bound $\max_{j\in[T]}\ltwonorm{\bE(j)}\le\sqrt n/(4T^{3/2})$, it gives, for every $t\in[T]$,
\bal\label{eq:comparison-pgd-filter}
\norm{\frac\eta n\sum_{j=0}^{t-1}\sum_{i=1}^n\Kattn(\cdot,\bbx_i)[\bbe_2(j)]_i}{\cH_{\Kattn}}
&\le\frac{\sqrt\eta}{4T^{3/2}}\sum_{j=1}^{t-1}\sqrt{t-j}\le\frac14.
\eal
The one-step estimate also gives $\ltwonorm{\bbe_2(j)}\le\sqrt n/(4\sqrt T)$ for $0\le j\le T$. Together with the signal and noise bounds of Lemma~\ref{lemma:early-stopped-residual-recursion}, (\ref{eq:comparison-pgd-filter}) bounds the RKHS component by $B_h$ with $\tau=T^{-1/2}$. The estimate uses the filtered errors of the actual recursion. The underlying spectral identity is also applicable to PGD, whereas its stated bound for arbitrary error vectors remains unchanged.

\smallskip
\noindent\textbf{Sharper sufficient width and auxiliary-sample conditions.}
These training and filtering estimates determine the sufficient network sizes. For a pointwise approximation tolerance $w\in(0,1)$, the PGD decomposition theorem~\cite[Theorem~B.6, Eqs.~(87)--(88)]{YangLi2026PGD} requires
\bal\label{eq:comparison-pgd-sizes}
m&\gsim d^{5/2}\max\set{T^{15/2}/w^5,T^{25/2}},\qquad
N\gsim\log(n/\delta)\max\set{T^2/w^2,T^4}.
\eal
\begin{wrapfigure}{r}{0.60\textwidth}
\centering
\small
\setlength{\tabcolsep}{4pt}
\captionof{table}{Sufficient size conditions with comparison to~\cite{YangLi2026PGD}. The symbol $T$ denotes the stopping time in the respective analysis.}
\label{tab:comparison-pgd-sizes}
\begin{tabular}{lcc}
\toprule
 & PGD~\cite{YangLi2026PGD} & Present paper\\
\midrule
Width $m$ & $\gsim T^{25/2}d^{5/2}$ & $\gsim\max\set{T^3d\log m,n^{1/(2d)}}$\\
Auxiliary size $N$ & $\gsim T^4\log(n/\delta)$ & $\gsim T^3\log(2/\delta)$\\
\bottomrule
\end{tabular}
\end{wrapfigure}
Theorem~\ref{theorem:bounded-NN-class} replaces the conditions in (\ref{eq:comparison-pgd-sizes}) by (\ref{eq:m-cond-bounded-NN-class}) and (\ref{eq:N-cond-bounded-NN-class}). The width terms are $T^2d\log m/w^2$ and $T^3d\log m$, together with $m\gsim n^{1/(2d)}$, and the auxiliary requirement is $\log(2/\delta)\max\set{T^2/w^2,T^3}$. The joint induction, (\ref{eq:comparison-pgd-filter}) and (\ref{eq:refined-NN-pointwise-error}) give these estimates.

\smallskip
\noindent Theorem~\ref{theorem:LRC-population-NN-eigenvalue} has approximation contribution $Cw^2$, so our main theorem takes $w=\eps_n\asymp T^{-1/2}$. The PGD proof~\cite[Theorem~5.1, Section~6.3]{YangLi2026PGD} uses a contribution of order $w$ and takes $w\asymp T^{-1}$. Table~\ref{tab:comparison-pgd-sizes} substitutes these choices into the respective size conditions. Our sample-size specializations are (\ref{eq:m-N-cond-LRC-population-NN-general}) and (\ref{eq:m-N-cond-LRC-population-NN-poly-edr}).

\smallskip
\noindent\textbf{Novelty and scope.}
Relative to the stated PGD results, our contribution is a finite-width guarantee for simultaneous training of both layers with smaller sufficient width and auxiliary-sample requirements, showing the benefits of channel attention and joint training. The joint induction and spectral filtering control the interaction of output-weight changes and feature motion up to the stopping time. Table~\ref{tab:comparison-pgd-sizes} compares sufficient conditions for the respective kernels and source balls under the stated exact-kernel and auxiliary-sampling assumptions. It makes no assertion about necessary widths or identical-target risk superiority.

\begin{samepage}
\section{Training by Gradient Descent}
\label{sec:training}
\begin{wrapfigure}{r}{0.53\textwidth}
\vspace{-\intextsep}
\begin{minipage}{\linewidth}
\small
\begin{algorithm}[H]
\caption{Training the Two-Layer NN with Channel Attention
(\ref{eq:two-layer-nn}) by GD}
\label{alg:GD}
\begin{algorithmic}[1]
        \State \textbf{input: } $T,\bW(0),\eta,\bS,\by,\bQ$
        \State Initialize $\ba(0)=0$ and construct the fixed matrix $\bA$
        \State \textbf{\bf for } $t=1,\ldots,T$ \,\,\textbf{\bf do }
            \State \quad Perform the $t$-th step of GD by
            (\ref{eq:GD-w-two-layer-nn})-(\ref{eq:GD-a-two-layer-nn})
        \State \textbf{\bf end for }
        \State \Return $(\bW(T),\ba(T))$
    \end{algorithmic}
\end{algorithm}
\end{minipage}
\end{wrapfigure}

We next describe the training procedure used in the main results. The algorithm and displayed updates are for the spherical network. For the Euclidean construction, the same updates use the attention matrix, normalized inputs and normalized responses of (\ref{eq:normalized-attention-network})--(\ref{eq:normalized-attention-loss}) in dimension $d+1$.
Algorithm~\ref{alg:GD} updates both layers simultaneously with $\bA$ fixed. We choose the ReLU subgradient at zero to be $1$, and define GD by the following explicit updates using the parameters at time $t$. The empirical quadratic loss is $L(\bW,\ba)\defeq(2n)^{-1}\sum_{i=1}^n\pth{f(\bW,\ba,\bbx_i)-y_i}^2$.

\end{samepage}

At step $t+1$, with $t\ge0$, we update the two layers by
\bal
\vect{\bW_{\bS}(t+1)} &=  \vect{\bW_{\bS}(t)} - \frac{\eta}{n} \bZ_{\bS}(t)
 (\hat \by(t) -  \by), \label{eq:GD-w-two-layer-nn} \\
\ba(t+1) &= \ba(t) - \frac{\eta}{n {\sqrt m}} \bA\bsigma(\bW(t),\bS)  (\hat \by(t) -  \by), \label{eq:GD-a-two-layer-nn}
\eal
The subscripts $\bS$ record dependence on the training covariates.
The matrix $\bZ_{\bS}(t)\in\RR^{md\times n}$ collects the first-layer backpropagation factors under the stated ReLU convention. Its blocks are defined by
$\bth{\bZ_{\bS}(t)}_{[(r-1)d+1:rd]i} = \frac {1}{{\sqrt m}}
\indict{\bbw_r(t)^\top \bbx_i \ge 0} \bbx_i  \bth{\bA \ba(t)}_r$ for all $i \in [n], r \in [m]$,
where $\bth{\bZ_{\bS}(t)}_{[(r-1)d+1:rd]i} \in \RR^d$ is a vector with elements in the $i$-th column of $\bZ_{\bS}(t)$ with indices in
$[(r-1)d+1:rd]$. We run $T\le\hat T$ simultaneous gradient steps.

\vspace{-.05in}
\section{Roadmap of Proofs}
\label{sec:proof-roadmap}
\vspace{-.05in}
We present the intermediate results leading to Theorems~\ref{theorem:minimax-nonparametric-regression-Kattn-general} and~\ref{theorem:minimax-nonparametric-regression-Kattn-poly-edr}. They connect the training updates with the critical radius through initialization-kernel approximation, function decomposition and population regression risk comparison.

\subsection{Detailed Roadmap and Key Technical Results}
\label{sec:detailed-roadmap-key-results}
The proof is based on three technical results: uniform approximation of the initialization kernel, an RKHS decomposition of the trained functions, and an empirical-to-population regression risk comparison. We present these results under the spherical setup. The polynomial-spectrum theorem then follows by evaluating the critical radius.

\smallskip
\noindent\textbf{Roadmap and Key Technical Results.}
First, we establish uniform convergence of the empirical attention kernel to its population counterpart. The covariance estimate in Theorem~\ref{theorem:hatKattn-close-to-Kattn-supnorm} accounts for repeated use of $\bQ$ and gives the following result.
\begin{theorem}\label{theorem:hatKattn-close-to-Kattn-supnorm-informal}
For every $\delta\in(0,1)$, there is one event of probability at least $1-\delta$ over $\bQ$ on which
\bals
\sup_{\bx,\bx'\in\cX}\abth{\hKattn(\bx,\bx')-\Kattn(\bx,\bx')}\lsim\sqrt{\frac{\log(2/\delta)}N}.
\eals
\end{theorem}
The uniform estimate applies to the training points on the same event. Lemma~\ref{lemma:empirical-loss-convergence-contraction-bKattn-approx} further controls the empirical intermediate kernel in the initialization NTK. Lemma~\ref{lemma:early-stopped-residual-recursion} then bounds both trained layers and their filtered recursion errors, giving the following function decomposition together with pointwise approximation.
\begin{theorem}\label{theorem:bounded-NN-class-informal}
Fix $\delta,w\in(0,1)$ and deterministic positive integers $m,N$. For each fixed training design and $1\le T\le\hat T$, under the size conditions and training prescription of Theorem~\ref{theorem:bounded-NN-class}, one event of conditional probability at least $1-2/n-\delta-C\exp\pth{-cn}-C\exp\pth{-cn\hat\eps_n^2}$ over $\bW(0),\bw,\bQ$ gives, simultaneously for every $t\in[T]$, the decomposition on $\cX$
\bals
f_t=h_t+e_t,
\eals
where $h_t\in\cH_{\Kattn}(B_h)$ and $\supnorm{e_t}\le w$. The fixed radius $B_h$ is defined in (\ref{eq:B_h}).
\end{theorem}
Second, the function decomposition permits local Rademacher complexity analysis on a fixed RKHS ball. Theorem~\ref{theorem:LRC-population-NN-eigenvalue} gives the following comparison, in which the uniformly bounded residual contributes $Cw^2$.
\begin{theorem}\label{theorem:LRC-population-NN-eigenvalue-informal}
Fix $\delta,w\in(0,1)$ and deterministic positive integers $m,N$. Let $T=T(\bS)$ be a measurable positive integer depending only on the training design and fixed parameters. Under the hypotheses of Theorem~\ref{theorem:LRC-population-NN-eigenvalue}, there is an event of probability at least $1-C/n-\delta-C\exp\pth{-cn}-C\exp\pth{-cn\eps_n^2}$ over $\bW(0),\bw,\bS,\bQ$ on which, whenever its stopping and size conditions hold, simultaneously for every $t\in[T]$,
\bals
\Expect{P}{(f_t-f^*)^2}-2\Expect{P_n}{(f_t-f^*)^2}\lsim\eps_n^2+w^2.
\eals
\end{theorem}
Finally, take $T=\hat T$ and $w=\eps_n$ for sufficiently large $n$. Lemma~\ref{lemma:hat-eps-eps-relation} gives $T\asymp\eps_n^{-2}$, so the main size conditions imply the intermediate ones. Theorem~\ref{theorem:empirical-loss-bound} gives $\Expect{P_n}{(f_t-f^*)^2}\lesssim(\eta t)^{-1}\lesssim\eps_n^2$ for $\lceil c_tT\rceil\le t\le T$. Applying Theorem~\ref{theorem:LRC-population-NN-eigenvalue-informal} proves Theorem~\ref{theorem:minimax-nonparametric-regression-Kattn-general} on the intersection of the design and conditional training events. Section~\ref{sec:proofs-main-results} gives the probability calculation.

\smallskip
\noindent The joint parameter bounds and filtered kernel-gradient sum connect initialization-kernel approximation to the fixed RKHS ball used in the population comparison. Appendix~\ref{sec:proofs-unbounded} proves the corresponding results after normalization on $\RR^d$, including conditional noise concentration and weighted auxiliary-covariance approximation. The tail estimate then gives the original regression risk.

\smallskip
\noindent\textbf{Relation to kernel linearization.} Zero second-layer initialization makes the first-layer derivative vanish at initialization. The resulting attention NTK is $m^{-1}\bsigma(\bW(0),\bx)^\top\bA^2\bsigma(\bW(0),\bx')$, with population limit $\Kattn$. The proof controls subsequent motion through kernel linearization for this architecture. The change from the vanilla initialization kernel does not establish escape from the NTK regime or feature learning.

\section{Conclusion}
We establish a quantitative regression risk guarantee for simultaneous GD training of a finite NN with fixed covariance-based channel attention. The population initialization operator is $T_K^{2s+3}$, and joint parameter control with filtered recursion errors yields explicit sufficient network sizes and critical-radius regression risk under early stopping. Under a two-sided spherical polynomial spectrum, the regression risk has order $n^{-2\alpha s'/(2\alpha s'+1)}$. On $\RR^d$, normalization preserves the target ball and gives a tail-dependent bound, with a sub-Gaussian specialization requiring no density. Exact evaluations of the regular NTK and its induced kernels and independent auxiliary covariates remain assumptions of the construction.

\begin{appendices}

\renewcommand{\baselinestretch}{1.33}
\selectfont

\section{Mathematical Tools}
\label{sec::math-tools}
Except where a general bounded kernel is specified, the appendices preceding Appendix~\ref{sec:proofs-unbounded} use the spherical construction on compact $\cX$. Appendix~\ref{sec:proofs-unbounded} supplies the changes on $\RR^d$, and Appendix~\ref{sec:proof-vanilla-bias} proves the biased vanilla comparison.
\subsection{Localized empirical processes}
\label{sec:concentration-sup-emp-process}
For an integrable function $f$, write $Pf=\int f\,d\mu$ and $P_nf=n^{-1}\sum_{i=1}^nf(\bbx_i)$.
\begin{definition}\label{def:RC}
For independent Rademacher signs $\sigma_i$, independent of the observations, the complexities are defined~by
\bals
\hat\cfrakR(\cF)=\Expect{\bsigma}{\sup_{f\in\cF}\frac1n\sum_{i=1}^n\sigma_i f(\bbx_i)},\qquad
\cfrakR(\cF)=\Expect{}{\hat\cfrakR(\cF)}.
\eals
For an arbitrary class we use outer expectation. Separable RKHS classes below have measurable suprema. The signs differ from the activation $\sigma(z)=\max\set{z,0}$. Write $R_n\cF=\sup_{f\in\cF}n^{-1}\sum_{i=1}^n\sigma_i f(\bbx_i)$.
\end{definition}
\begin{definition}[{\cite[Definition 3.1]{bartlett2005}}]\label{def:sub-root-function}
A function $\psi:[0,\infty)\to[0,\infty)$ is sub-root if it is nonnegative and nondecreasing, and $\psi(r)/\sqrt r$ is nonincreasing for $r>0$.
\end{definition}
We use contraction, $\Expect{\bsigma}{R_n(\phi\circ\cF)}\le L_\phi\Expect{\bsigma}{R_n\cF}$ for an $L_\phi$-Lipschitz $\phi$~\cite{Ledoux-Talagrand-Probability-Banach}, and the following localized comparison.
\begin{theorem}[{\cite[Theorem 3.3]{bartlett2005}}]\label{theorem:LRC-population}
Let $\cF$ be pointwise measurable with range in $[a,b]$, and suppose $\Var{f}\le T(f)\le\bar B Pf$ for some $\bar B>0$. Let $\psi$ be a nonzero sub-root function with fixed point $r^*$ satisfying $\psi(r)\ge\bar B\cfrakR\set{f:T(f)\le r}$ for $r\ge r^*$. For $K_0>1$ and $x>0$, with probability at least $1-\exp\pth{-x}$,
\bals
Pf\le\frac{K_0}{K_0-1}P_nf+\frac{704K_0}{\bar B}r^*
+\frac{x\set{11(b-a)+26\bar B K_0}}n,\qquad f\in\cF.
\eals
\end{theorem}
\section{More Details about the Theoretical Results}
\label{sec:Q-generation}
The analysis assumes $N$ independent observations from $\mu$, independent of the labeled sample. Reusing training covariates or a generator fitted to $\bS$ would require a corresponding dependence and distribution-approximation analysis. Exact evaluation of the spectral kernel $K^{(s)}$ induced by the regular NTK is an additional assumption beyond sampling~$\mu$.
\label{sec:difference-existing-kernel-theory}
The induced operator $T_K^{2s+3}$ defines the interpolation RKHS used by both Proposition~\ref{prop:kernel-baseline} and the network theorem. The latter additionally controls joint parameter motion and finite-sample approximation, with sufficient-size comparisons in Section~\ref{sec:comparison-pgd}.
\label{sec:better-network-width}
For fixed $\alpha,s'$, the sufficient width $m\gtrsim n^{6\alpha s'/(2\alpha s'+1)}d\log m$ has sample-size exponent less than $3$. This condition quantifies the width required by the proof, with comparisons made under the respective target, kernel, architecture and information assumptions.

\section{Detailed Proofs}
\label{sec:proofs}

We present the detailed proofs for the theoretical results, and basic notations are introduced in Section~\ref{sec:proofs-definitions}.

\subsection{Basic Definitions}
\label{sec:proofs-definitions}
We use the attention Gram matrix, its normalization and its eigendecomposition defined in Section~\ref{sec:setup-kernels-target-function}. Proposition~\ref{proposition:Kattn-gram-nonsingular} proves nonsingularity for the assumed non-antipodal sample. In particular all empirical attention eigenvalues in the denominators below are positive.
Throughout the conditional noise analysis, the training design is fixed and $T$ is a measurable positive integer function of that design and the fixed parameters. Thus $T$ is independent of the response noise, $\bQ$ and initialization, as is $T=\hat T$ whenever $\hat T\ge1$. The integers $m,N$ are deterministic. Statements with residual or motion hypotheses hold uniformly over trajectories satisfying those hypotheses on the specified probability event.

For $0<\tau\le1$ and a positive integer $T$, the residual and fixed quantities used below are defined~by $\bu(t)=\hat\by(t)-\by$, $c_{\bu}=\mu_0/\min\set{\sqrt{2e\eta},1}+\sigma_0+\tau+1$ and $R=\eta c_{\bu}(4d+6\log(2mn))T/\sqrt m$. Define
\bals
\cV_t&=\set{- (\bI_n-\eta\bKattn_n)^tf^*(\bS)},\\
\cE_{t,\tau}&=\set{\bbe_1+\bbe_2:\bbe_1=-(\bI_n-\eta\bKattn_n)^t\bw,\ \ltwonorm{\bbe_2}\le\sqrt n\tau}.
\eals
For $\delta\in(0,1)$, set $q=2d\log(1+2m)+\log(4/\delta)$ and define
\bal\label{eq:C1}
C_1(m,d,\delta)=C\kappa^2\left(\sqrt{q/m}+q/m+\frac{d+\log(2m/\delta)}m\right).
\eal
Here $C$ is the fixed absolute constant in Theorem~\ref{theorem:sup-hat-g}. The good initialization set for (\ref{eq:good-initialization-sup-hat-h}) and the Gaussian row bound is defined~by
\bals
\cW_0=\set{\bW(0):(\ref{eq:good-initialization-sup-hat-h})\text{ holds},\ \max_{r\in[m]}\ltwonorm{\bbw_r(0)}^2\le4d+6\log(2mn)}.
\eals
Lemma~\ref{lemma:empirical-loss-convergence} gives motion and residual bounds for arbitrary $\tau$. Under $\tau=T^{-1/2}$ and $T\le\hat T$, Lemma~\ref{lemma:early-stopped-residual-recursion} gives sharper bounds together with the bounds $R,2R$.
The actual GD iterates determine the sets
\bals
\cW(\bS,\bW(0),T)&=\set{(\bW(t),\ba(t)):1\le t\le T,\ \bu(k)\in\cV_k+\cE_{k,\tau}\text{ for }0\le k<t},\\
\cFnn(\bS,\bW(0),T)&=\set{f_t=f(\bW(t),\ba(t),\cdot):(\bW(t),\ba(t))\in\cW(\bS,\bW(0),T)}.
\eals
Define $\cF(B,w)=\set{h+e:h\in\cH_{\Kattn}(B),\ \supnorm{e}\le w}$. We leave the dependence of the trajectory sets on $\bQ$ and noise implicit and use the Minkowski convention for sums of sets. Theorem~\ref{theorem:bounded-NN-class} places the trained functions in $\cF(B_h,w)$, with radius defined~by (\ref{eq:B_h}).
\subsection{Uniform Convergence to the NTK (\ref{eq:kernel-two-layer})}
\label{sec:uniform-convergence-ntk-more}
We define the following functions with $\bW = \set{\bw_r}_{r=1}^m$:
\bals
h(\bw,\bu,\bv) \defeq \relu{\bw^{\top} \bu} \relu{\bw^{\top} \bv}, \quad \hat h(\bW,\bu,\bv) \defeq \frac {1}{m} \sum\limits_{r=1}^m h(\bbw_r,\bu,\bv), 
\eals
where $\bu,\bv \in \cX$. Then we have the following theorem stating the uniform convergence of $\hat h(\bW(0),\cdot,\cdot)$ to $K(\cdot,\cdot)$.

\begin{theorem}\label{theorem:good-random-initialization}
Suppose $m\ge2$, $m \gsim n^{1/(2d)}$ and $m/\log m \ge d$. Then with probability at least $1-1/n$ over the random initialization $\bW(0) = \set{\bbw_r(0)}_{r=1}^m$ with $\bbw_{r}(0) \overset{\textup{i.i.d.}}{\sim} \cN(\bzero,\kappa^2 \bI_d)$
for all $r \in [m]$,
\bal\label{eq:good-initialization-sup-hat-h}
\sup_{\bu \in \cX,\bv \in \cX} \abth{ K(\bu,\bv) -
\hat h(\bW(0),\bu,\bv) }
\le C_1(m,d,1/n) \lsim \sqrt{\frac{d \log m}{m}},
\eal
where $C_1$ is defined in (\ref{eq:C1}).
\end{theorem}
\begin{proof}
Apply Theorem~\ref{theorem:sup-hat-g} with $\delta=1/n$. Since $m\gtrsim n^{1/(2d)}$, $\log n\le2d\log m+\cO(d)$, so both $q$ and $d+\log(2mn)$ in (\ref{eq:C1}) are $\cO(d\log(2m))$. The assumption $m/\log m\ge d$ then gives
$C_1(m,d,1/n)\lesssim\sqrt{d\log m/m}$.
Under the hypotheses of this theorem, $m\gtrsim d\log m$ with a sufficiently large constant implies $C_1(m,d,1/n)\le1-b$, since $b<1/2$. The probability remains $1-1/n$.
\end{proof}

Theorem~\ref{theorem:good-random-initialization} and the Gaussian row bound give $\Pr\set{\bW(0)\in\cW_0}\ge1-2/n$. The movement estimates hold deterministically on this intersection.

\subsection{Proofs for the Main Result,
Theorem~\ref{theorem:minimax-nonparametric-regression-Kattn-general}
and
Theorem~\ref{theorem:minimax-nonparametric-regression-Kattn-poly-edr}
}
\label{sec:proofs-main-results}

The results below implement the roadmap in Section~\ref{sec:detailed-roadmap-key-results}.

\begin{proof}[\textbf{\textup{Proof of Theorem~\ref{theorem:minimax-nonparametric-regression-Kattn-general}}}]
\textbf{Outline:} Compare critical radii, verify the stopping and size conditions, then combine empirical and population bounds.
On the design event of Lemma~\ref{lemma:hat-eps-eps-relation}, $\hat\eps_n^2\asymp\eps_n^2$. For $T=\hat T\ge1$, the integer stopping identity gives
$\hat\eps_n^2\le (\eta T)^{-1}<2\hat\eps_n^2$,
since $T\le1/(\eta\hat\eps_n^2)<T+1\le2T$. Hence $T\asymp\eps_n^{-2}$ and $1/(\eta t)\lesssim\eps_n^2$ for $\lceil c_tT\rceil\le t\le T$.
Take $w=\eps_n$ and $\tau=T^{-1/2}$. The stated sizes ensure $m\ge2$, $m\gtrsim n^{1/(2d)}$, $m/\log m\ge d$ and $C_1\le1-b$. In (\ref{eq:m-cond-bounded-NN-class}) the two time/tolerance terms are both of order $d\log m\,\eps_n^{-6}$, and in (\ref{eq:N-cond-bounded-NN-class}) both are of order $\eps_n^{-6}\log(2/\delta)$. Thus all recursion, decomposition and population-comparison hypotheses hold on this radius event.
Theorems~\ref{theorem:empirical-loss-bound} and~\ref{theorem:LRC-population-NN-eigenvalue} give clean empirical risk $C/(\eta t)$ and population risk at most twice this quantity plus $C(\eps_n^2+w^2)$. This proves the risk bound. To calculate its probability, intersect the critical-radius and deterministic-class design events. On each such design, the size conditions hold at $T(\bS)$, the feature, row and auxiliary failures total at most $2/n+\delta$, and the noise failures total at most $C\exp\pth{-cn}+C\exp\pth{-cn\hat\eps_n^2}$. Radius comparison bounds the last term by $C\exp\pth{-c'n\eps_n^2}$. Integrating over these designs and adding their complement gives the stated probability. Uniformity through $T$ gives the simultaneous assertion.
\end{proof}

\begin{proof}[\textbf{\textup{Proof of Theorem~\ref{theorem:minimax-nonparametric-regression-Kattn-poly-edr}}}]
\textbf{Outline:} Compute the attention complexity from the spectrum and substitute its critical radius into the general theorem.
Theorem~\ref{theorem:spectrum-Kattn} gives $\lambdaattn_j=\lambda_j^{s'}\asymp j^{-\beta}$, $\beta=2\alpha s'>1$. Splitting at $J=\lfloor\eps^{-2/\beta}\rfloor$, the two-sided spectral bounds and integral comparison give
\bals
\sum_{j=1}^{\infty}\min\set{\lambdaattn_j,\eps^2}\asymp J\eps^2+\sum_{j=J+1}^{\infty}j^{-\beta}\asymp\eps^{2-2/\beta}.
\eals
The fixed-point equation is $\eps^4\asymp n^{-1}\eps^{2-2/\beta}$, so $\eps_n^2\asymp n^{-\beta/(\beta+1)}$. The general theorem then gives $\hat T\asymp n^{\beta/(\beta+1)}$, $m\gtrsim n^{3\beta/(\beta+1)}d\log m$ and $N\gtrsim n^{3\beta/(\beta+1)}\log(2/\delta)$. The initialization term is smaller because $\beta>5$ and $d\ge5$. Finally $n\eps_n^2\asymp n^{1/(\beta+1)}$ gives the displayed failure exponent.
\end{proof}

\begin{proposition}\label{prop:sharper-eps-n-eps-Kn}
We have $\eps_n^2 \le \eps_{K,n}^2$.
\end{proposition}
\begin{proof}
Since $0<\lambda_j\le b<1$ and $\lambdaattn_j=\lambda_j^{s'}\le\lambda_j$, coefficientwise comparison gives $R_{\Kattn}(\eps)\le R_K(\eps)$ for every $\eps\ge0$. At $\eps_n$, we have $\eps_n^2=\sigma_0R_{\Kattn}(\eps_n)\le\sigma_0R_K(\eps_n)$. Sub-root fixed-point crossing~\cite[Lemma 3.2]{bartlett2005} gives the claim.
\end{proof}

\subsection{Key Technical Results}
\label{sec:detailed-key-results}
The coarse recursion below allows arbitrary $0<\tau\le1$. Lemma~\ref{lemma:early-stopped-residual-recursion} controls the actual joint updates before stopping and uses spectral filtering to attain $\tau=T^{-1/2}$.

\begin{lemma}\label{lemma:empirical-loss-convergence}
Recall that $b=\kappa^2/2$. Fix the training design and suppose $\delta\in(0,1)$, $N\ge1$, $m \gsim d\log m$ and
\bals
m \gsim \max\set{T^2d\log m/\tau^2, T^4 (d^4+\log^4 m ) /\tau^2,n^{1/(2d)}},  \quad N \gsim T^2 \log(2/\delta)/\tau^2. 
\eals
Train the neural network $f(\bW(t),\ba(t),\cdot)$ by GD
using Algorithm~\ref{alg:GD} with the learning rate $\eta=\Theta(1)\in(0,1)$. The following conclusion is asserted on the good feature event $\bW(0)\in\cW_0$, whose failure probability is included below. Then with probability at least
$1-2/n-\delta-\exp\pth{-\Theta(n)}$ over the random initialization $\bW(0)$, the random noise $\bw$, and the random sample $\bQ$,
$(\bW(t),\ba(t)) \in \cW(\bS,\bW(0),T)$ for every $t \in [T]$.
Moreover, for every $t \in [0,T]$, $\bu(t) = \bv(t) + \be(t)$ where
$\bv(t) \in \cV_t$,
$\be(t) \in \cE_{t,\tau}$,  $\ltwonorm{\bu(t)} \le c_{\bu} \sqrt n$, and
$\ltwonorm{\bbw_r(t) - \bbw_r(0)} \le R$, $\abth{a_r(t) - a_r(0)} \le 2R$.
\end{lemma}

The stopping restriction gives additional control of the actual GD recursion. In the next lemma the errors arise from the preceding updates. Their spectral filtering permits $\tau=T^{-1/2}$, whereas the unrestricted errors in Lemma~\ref{lemma:bounded-Linfty-vt-sum-et} require its stronger tolerance condition.
\begin{lemma}\label{lemma:early-stopped-residual-recursion}
Recall that $b=\kappa^2/2$. Fix the training design, $1\le T\le\hat T$, $\delta\in(0,1)$ and $\tau=T^{-1/2}$. Suppose
\bal\label{eq:early-stopped-recursion-sizes}
m\gtrsim\max\set{T^3d\log m,n^{1/(2d)}},\qquad
N\gtrsim T^3\log(2/\delta).
\eal
Train the network by Algorithm~\ref{alg:GD} with $\eta=\Theta(1)\in(0,1)$. There is one event of conditional probability at least
$1-2/n-\delta-\exp\pth{-cn}-\exp\pth{-cn\hat\eps_n^2}$ over $\bW(0),\bw,\bQ$ on which, simultaneously for every $0\le t\le T$, $\bu(t)\in\cV_t+\cE_{t,\tau}$, $\ltwonorm{\bu(t)}\le c_{\bu}\sqrt n$ and
\bal\label{eq:early-stopped-weight-bounds}
\ltwonorm{\ba(t)}\lesssim T^{1/4},\qquad
\max_{r\in[m]}\ltwonorm{\bbw_r(t)-\bbw_r(0)}
\lesssim\frac{(d+\log(2mn))T^{5/4}}m.
\eal
The original movement bounds $\ltwonorm{\bbw_r(t)-\bbw_r(0)}\le R$ and $\abth{a_r(t)}\le2R$ also hold. For $0\le k<T$, the actual one-step errors in the residual recursion satisfy
\bal\label{eq:early-stopped-one-step-error}
\bu(k+1)=(I-\eta\bKattn_n)\bu(k)+\bE(k+1),\qquad
\max_{1\le j\le T}\ltwonorm{\bE(j)}\le\frac{\sqrt n}{4T^{3/2}}.
\eal
For $0\le t\le T$, define
\bal\label{eq:early-stopped-h-definition}
h_t=-\frac\eta n\sum_{k=0}^{t-1}\sum_{i=1}^n\Kattn(\cdot,\bbx_i)u_i(k),\qquad h_0=0.
\eal
On the same event, $\norm{h_t}{\cH_{\Kattn}}\le B_h$ for every $t\le T$.
\end{lemma}

The pointwise tolerance $w$ introduces additional size conditions, while stopping controls the RKHS noise component.
\begin{theorem}\label{theorem:bounded-NN-class}
Recall that $b=\kappa^2/2$. Set $\tau=T^{-1/2}$. Suppose $\delta\in(0,1)$, $w\in(0,1)$,
\bal
m &\gsim \max\set{T^3d\log m,T^2d\log m/w^2,n^{1/(2d)}}, \label{eq:m-cond-bounded-NN-class} \\
N &\gsim \max\set{T^3\log(2/\delta),T^2\log(2/\delta)/w^2},  \label{eq:N-cond-bounded-NN-class}
\eal
and the neural network
$f_t = f(\bW(t),\ba(t),\cdot) $ is trained by GD using Algorithm~\ref{alg:GD} with the learning rate $\eta=\Theta(1)\in(0,1)$. The following conclusion is asserted on the good feature event $\bW(0)\in\cW_0$, whose failure probability is included below. For each fixed training design and $T\le\hat T$, one event of conditional probability at least
$1-2/n-\delta-\exp\pth{-\Theta(n)}-\exp\pth{-\Theta(n \hat\eps_n^2)}$ over $\bW(0),\bw,\bQ$ gives, simultaneously for every $t\in[T]$,
$f_t\in \cFnn(\bS,\bW(0),T) \subseteq \cF(B_h,w)$, and $f_t$ has the following decomposition on $\cX$:
$f_t = h_t + e_t$,
where $h_t \in \cH_{\Kattn}(B_h)$ with $B_h$ defined in (\ref{eq:B_h}),
$e_t  \in L^{\infty}$ with $\supnorm{e_t } \le w$.
\end{theorem}

Arbitrary residual classes in the following bound use outer expectation. The risk proof applies concentration to a deterministic, separable RKHS ball and handles its residual deterministically.
\begin{lemma}[\hspace{-0.1pt}Adapted from {\cite[Lemma C.9]{Yang2025-generalization-two-layer-regression}}]\label{lemma:LRC-population-NN}
For every $B>0$, $w\ge0$ and $r>0$,
\bal\label{eq:LRC-population-NN}
\cfrakR
\pth{\set{f \in \cF(B,w) \colon \Expect{P}{f^2} \le r}}
\le \varphi_{B,w}(r),
\eal
where
\bal\label{eq:varphi-LRC-population-NN}
\varphi_{B,w}(r) \defeq
\min_{Q\in\set{0,1,2,\ldots}} \pth{({\sqrt r} + w) \sqrt{\frac{Q}{n}} +
B
\pth{\frac{\sum\limits_{q = Q+1}^{\infty}\lambdaattn_q}{n}}^{1/2}} + w.
\eal
\end{lemma}
\begin{theorem}\label{theorem:LRC-population-NN-eigenvalue}
Recall that $b=\kappa^2/2$. Fix $\delta\in(0,1)$, $w\in(0,1)$ and deterministic positive integers $m,N$. Train the network $f_t=f(\bW(t),\ba(t),\cdot)$ by Algorithm~\ref{alg:GD} with fixed $\eta=\Theta(1)\in(0,1)$. Let $T=T(\bS)$ be a measurable positive integer depending only on the training design and fixed parameters. For sufficiently large $n$, there is an event of probability at least
$1-C/n-\delta-C\exp\pth{-cn}-C\exp\pth{-cn\eps_n^2}$ over $\bW(0),\bw,\bS,\bQ$ on which the following implication holds. If $T\le\hat T$ and the bounds (\ref{eq:m-cond-bounded-NN-class}) and (\ref{eq:N-cond-bounded-NN-class}) hold with the sufficiently large constants in Theorem~\ref{theorem:bounded-NN-class}, then simultaneously for every $t\in[T]$,
\bals
\Expect{P}{(f_t-f^*)^2} - 2 \Expect{P_n}{(f_t-f^*)^2}
\lsim \eps_n^2 +w^2.
\eals
\end{theorem}

The clean empirical loss is controlled on one simultaneous noise event.
\begin{theorem}\label{theorem:empirical-loss-bound}
Suppose the neural network trained after the $t$-th step of GD, $f_t = f(\bW(t),\ba(t),\cdot)$, satisfies $\bu(t) = \bv(t) + \be(t)$
with $\bv(t) \in \cV_t$ and $\be(t) \in \cE_{t,\tau}$, and $t \in [T]$ with $T \le \hat T$. Assume
$\tau\le\min\set{1,(\eta T)^{-1/2}}$. For each fixed training design, there is a noise event of probability at least
$1-\exp\pth{-\Theta(n\hat \eps_n^2)}$ on which, simultaneously for every $t\in[T]$ satisfying the assumed residual representation, we have
\bals
\Expect{P_n}{(f_t-f^*)^2} \le
\frac{3}{\eta t} \pth{\frac{\mu_0^2}{2e}+3}.
\eals
\end{theorem}

\subsection{Proofs for Results in Section~\ref{sec:uniform-convergence-ntk-more} and Section~\ref{sec:detailed-key-results}}
\label{sec:proofs-key-results}

The following feature estimate is proved in Section~\ref{sec:proofs-uniform-convergence-sup-hat-g-V_R}.
\begin{theorem}\label{theorem:sup-hat-g}
Let ${\bW(0)} = \set{{{\bbw_r(0)}}}_{r=1}^m$, where the rows are independent and each ${{\bbw_r(0)}} \sim \cN(\bzero,\kappa^2 \bI_d)$ for $r \in [m]$. Then for any $\delta \in (0,1)$, with probability at least $1-\delta$ over ${\bW(0)}$,
\bal\label{eq:sup-hat-h}
\sup_{\bu \in \cX,\bv \in \cX} \abth{ K(\bu,\bv) - \hat h({\bW(0)},\bu,\bv) } \le C_1\pth{m,d,\delta},
\eal
where $C_1$ is defined in (\ref{eq:C1}).

\end{theorem}

\begin{proof}[\textbf{\textup{Proof of Lemma~\ref{lemma:empirical-loss-convergence}}}]
The size conditions imply $C_1\le1-b$ and $\log(2mn)\lesssim d\log m$. Intersect the feature, row-norm, covariance and Lemma~\ref{lemma:yt-y-bound} events, with total failure at most $2/n+\delta+\exp\pth{-cn}$. The coarse error bound (\ref{eq:empirical-loss-Et-bound-Em}) is
$\bE_{m,\eta,\delta}\lesssim\sqrt n\set{\frac{T(d^2+\log^2m)+\sqrt{d\log m}}{\sqrt m}+\sqrt{\frac{\log(2/\delta)}N}}.$
The three size conditions make each term at most $\tau/(3C_{\rm rec}T)$, where $C_{\rm rec}$ is the contraction-lemma constant. Thus every allowed preceding step has $\norm{\bE(k)}{}\le\tau\sqrt n/T$.
At zero $\bu(0)=-f^*(\bS)-\bw$, with zero accumulated error. Suppose the residual bound holds through $t<T$. Iterating the contraction recursion gives
\bals
\bu(t+1)=-\bS_{t+1}f^*(\bS)-\bS_{t+1}\bw+\bbe_2(t+1),\qquad
\bbe_2(t+1)=\sum_{k=1}^{t+1}\bS_{t+1-k}\bE(k),
\eals
where $\bS_j=(\bI_n-\eta\bKattn_n)^j$. Since $\norm{\bS_j}{}\le1$, $\norm{\bbe_2(t+1)}{}\le\tau\sqrt n$. The simultaneous signal/noise bound then gives $\norm{\bu(t+1)}{}\le c_{\bu}\sqrt n$, closing induction and proving the residual representation. The deterministic movement lemma gives the original $R,2R$ bounds through $T$, on the same event.
\end{proof}

\begin{proof}[\textbf{\textup{Proof of Lemma~\ref{lemma:early-stopped-residual-recursion}}}]
Fix the design and write $\bfsym{A}=\bKattn_n$, $\bS_j=(\bI_n-\eta \bfsym{A})^j$, $q_j(z)=1-(1-\eta z)^j$, $\bF_j=\bsigma(\bW(j),\bS)$, $\bG=\bF_0$, $C_1=C_1(m,d,1/n)$, $a_N=a_N(\delta)$ and $D=d+\log(2mn)$. Let $\Phi \bc=\sum_{i=1}^n c_i\Kattn(\cdot,\bbx_i)$, so $\Phi^*\Phi=n\bfsym{A}$. Intersect the feature, row-norm and covariance events. On this intersection, $\ltwonorm{\bA}\le1$, $\ltwonorm{\bG}\le\sqrt{mn}$, $\ltwonorm{\bA_r}\lesssim D/\sqrt m$, and the exact initialization-kernel comparison gives
\bal\label{eq:early-stopped-kernel-error}
\ltwonorm{\bG^\top\bA^2\bG/(nm)-\bfsym{A}}\le C\set{C_1+a_N(\delta)},\qquad
C_1+a_N(\delta)\le c_*T^{-3/2}.
\eal
The initialization-kernel comparison follows from Lemma~\ref{lemma:empirical-loss-convergence-contraction-bKattn-approx}, and $c_*$ can be made arbitrarily small by increasing the size constants. The initialization assumptions give $D\lesssim d\log m$.
The signal and filtered-noise calculations in the proof of Lemma~\ref{lemma:bounded-Linfty-vt-sum-et} do not use its tolerance condition. They give one further event on which, for all $t\le T$,
\bal\label{eq:early-stopped-signal-noise}
\norm{\frac\eta n\Phi\sum_{j=0}^{t-1}\bS_j\bw}{\cH_{\Kattn}}\le\sqrt2.
\eal
Since $q_T(z)^2/z\le\eta T\min\set{1,\eta Tz}$, stopping bounds the proxy trace by $1$ and the matrix norm by $\eta T/n$. Lemma~\ref{lemma:calibration-quadratic} with $x=cn/(\eta T)$ bounds the squared noise norm by $2$ with failure at most $\exp\pth{-cn\hat\eps_n^2}$. Monotonicity of the filters makes the bound simultaneous. Lemma~\ref{lemma:yt-y-bound} supplies the residual-noise bound with failure at most $\exp\pth{-cn}$.
Suppose the residual recursion has held through time $t-1$. Its accumulated error is $\bbe_2(j)=\sum_{i=1}^j\bS_{j-i}\bE(i)$. For each error at time $i$, the subsequent indices are $j=i,\ldots,t-1$, so its propagators sum to $\sum_{\ell=0}^{t-1-i}\bS_\ell$. The finite geometric identity $\eta\sum_{\ell=0}^{t-1-i}\bS_\ell=\bfsym{A}^{-1}(\bI_n-\bS_{t-i})$ is valid because $\bfsym{A}$ is nonsingular. Reversing these finite sums gives
\bal\label{eq:early-stopped-filtered-error}
\frac\eta n\Phi\sum_{j=0}^{t-1}\bbe_2(j)\stackrel{\circled{1}}{=}\frac1n\Phi\sum_{i=1}^{t-1}\bfsym{A}^{-1}(\bI_n-\bS_{t-i})\bE(i).
\eal
To estimate the contribution of each error in the RKHS norm, diagonalize $\bfsym{A}$. For $\bc\in\RR^n$ and an integer $1\le\ell\le T$, the identity $\Phi^*\Phi=n\bfsym{A}$ gives
\bals
\norm{n^{-1}\Phi \bfsym{A}^{-1}(\bI_n-\bS_\ell)\bc}{}^2=\frac1n\sum_{i=1}^n\frac{q_\ell(\hlambdaattn_i)^2}{\hlambdaattn_i}([{\bUattn}^\top \bc]_i)^2\stackrel{\circled{2}}{\le}\frac{\eta\ell}n\norm{\bc}{}^2.
\eals
Here \circled{2} follows from $q_\ell(z)\le\min\set{1,\eta\ell z}$ and $q_\ell(z)^2/z\le\eta\ell$. For $\eta\ell z\le1$, the latter is bounded by $(\eta\ell)^2z\le\eta\ell$, and for $\eta\ell z\ge1$ it is bounded by $z^{-1}\le\eta\ell$. Thus the filtered operator in (\ref{eq:early-stopped-filtered-error}) is bounded independently of the minimum empirical eigenvalue. Under (\ref{eq:early-stopped-one-step-error}), the triangle inequality gives
$\norm{\frac\eta n\Phi\sum_{j=0}^{t-1}\bbe_2(j)}{}\le \frac{\sqrt\eta}{4T^{3/2}}\sum_{i=1}^{t-1}\sqrt{t-i}\le\frac14.$
The final bound follows from $t\le T$, $\eta<1$ and $\sum_{i=1}^{t-1}\sqrt{t-i}\le T^{3/2}$. Combining (\ref{eq:early-stopped-filtered-error}) with (\ref{eq:early-stopped-signal-noise}) gives $\norm{h_t}{}\le B_h$, using only errors strictly before $t$.
Choose $C_a>2(B_h+1)$ and induct on the actual updates. At zero the output weights and motion vanish, while Lemma~\ref{lemma:yt-y-bound} gives $\norm{\bu(0)}{}\le c_{\bu}\sqrt n$. Suppose through $k<T$ that this residual bound holds, $\norm{\ba(j)}{}\le C_aT^{1/4}$ and all preceding errors have the asserted size. The first-layer update gives
\bal\label{eq:early-stopped-single-step-motion}
\max_{r\in[m]}\norm{\bbw_r(k+1)-\bbw_r(k)}{}\le \frac{\eta c_{\bu}}{\sqrt m}\max_{r\in[m]}\abth{[\bA\ba(k)]_r}\lesssim C_aDT^{1/4}/m.
\eal
Here unit inputs and $n^{-1}\sum_{i=1}^n\abth{u_i(k)}\le c_{\bu}$ justify this increment bound. Summing (\ref{eq:early-stopped-single-step-motion}) bounds total motion through $k+1$ by $CC_aDT^{5/4}/m\le1$.
We next bound the output weights using the residuals through time $k$. Let $\bc=\sum_{j=0}^k\bu(j)$. Telescoping the output-weight updates and separating the initialization features gives
\bal\label{eq:early-stopped-output-weight}
\norm{\frac\eta{n\sqrt m}\bA \bG\bc}{}^2\stackrel{\circled{3}}{\le}\norm{h_{k+1}}{}^2+\frac{C\eta^2}n(C_1+a_N)\norm{\bc}{}^2\le B_h^2+CT^2(C_1+a_N).
\eal
The exact quadratic form $\norm{\Phi \bc}{}^2=n\bc^\top \bfsym{A}\bc$ and (\ref{eq:early-stopped-kernel-error}) prove \circled{3}. Meanwhile ReLU Lipschitz continuity gives $\norm{\bF_j-\bG}{}\le\sqrt{mn}\max_{r\in[m]}\norm{\bbw_r(j)-\bbw_r(0)}{}$, so
\bals
\norm{\frac\eta{n\sqrt m}\bA\sum_{j=0}^k(\bF_j-\bG)\bu(j)}{}\le \eta c_{\bu}\sum_{j=0}^k\max_{r\in[m]}\norm{\bbw_r(j)-\bbw_r(0)}{}\lesssim C_aDT^{9/4}/m.
\eals
We now choose the size constants to close the output-weight bound in (\ref{eq:early-stopped-output-weight}). The kernel and width conditions can simultaneously enforce $T\sqrt{C_1+a_N}\le\sqrt{c_*}T^{1/4}$ and $DT^{9/4}/m\le c_*T^{-3/4}$. Choose $c_*$ so that $C\sqrt{c_*}\le1$ and $Cc_*\le1/2$. Since $T\ge1$, the frozen contribution is at most $(B_h+1)T^{1/4}$ and the moving contribution is at most $C_aT^{1/4}/2$. Thus their sum is at most $C_aT^{1/4}$ by $C_a>2(B_h+1)$. This establishes the output-weight bound at $k+1$ from the induction hypotheses through $k$.
The simultaneous-update identity is now
\bals
\hat\by(k+1)-\hat\by(k)=-\eta \bF_k^\top\bA^2\bF_k\bu(k)/(nm)+\bfm{r}_k,\qquad
\bfm{r}_k=m^{-1/2}(\bF_{k+1}-\bF_k)^\top\bA\ba(k+1).
\eals
The new output-weight bound and the single-step motion bound give $\norm{\bfm{r}_k}{}/\sqrt n\le CC_a^2DT^{1/2}/m$. Telescoping the two outer feature factors gives $\norm{(\bF_k^\top\bA^2\bF_k-\bG^\top\bA^2\bG)/(nm)}{}\le CC_aDT^{5/4}/m$, using $\norm{\bG}{}\le\sqrt{mn}$ and motion at most one. Therefore
$\frac{\norm{\bE(k+1)}{}}{\sqrt n}\le C\set{DT^{5/4}/m+C_1+a_N(\delta)}\le\frac1{4T^{3/2}}$.
The last bound uses $D\lesssim d\log m$, $T^{11/4}\le T^3$ and (\ref{eq:early-stopped-recursion-sizes}), with constants depending only on the fixed $C_a$.
Iterating this new recursion gives $\bu(k+1)=-\bS_{k+1}f^*(\bS)-\bS_{k+1}\bw+\bbe_2(k+1)$, with
$\norm{\bbe_2(k+1)}{}\le \sum_{i=1}^{k+1}\norm{\bE(i)}{}\le\sqrt n/(4\sqrt T)\le\sqrt n\tau$.
Lemma~\ref{lemma:yt-y-bound} completes the residual induction. The first-layer motion is at most $R$, and part (ii) of Lemma~\ref{lemma:second-layer-weight-vector-movement} gives $\abth{a_r(t)}\le2R$ under unit motion. The trajectory definitions yield membership in $\cW$ and the decomposition. Intersecting the feature, row, covariance and two noise events gives the stated probability uniformly through time $T$.
\end{proof}

\begin{proof}[\textbf{\textup{Proof of Theorem~\ref{theorem:bounded-NN-class}}}]
We apply Lemma~\ref{lemma:early-stopped-residual-recursion} and put $C_1=C_1(m,d,1/n)$, $a_N=a_N(\delta)$, $D=d+\log(2mn)$, $\bF_k=\bsigma(\bW(k),\bS)$ and $g_t(\bx)=m^{-1/2}\bsigma(\bW(0),\bx)^\top\bA\ba(t)$. Its event supplies (\ref{eq:early-stopped-weight-bounds}) and bounds the RKHS sum (\ref{eq:early-stopped-h-definition}) by $B_h$. ReLU Lipschitz continuity and $\norm{\bA}{}\le1$ therefore give
\bals
\abth{f_t(\bx)-g_t(\bx)}\le \max_{r\in[m]}\norm{\bbw_r(t)-\bbw_r(0)}{}\,\norm{\ba(t)}{}\lesssim DT^{3/2}/m.
\eals
Telescoping the exact second-layer updates from zero gives
$g_t(\bx)=-\frac\eta{nm}\sum_{k=0}^{t-1}\bsigma(\bW(0),\bx)^\top\bA^2\bF_k\bu(k).$
Replacing $\bF_k$ by $\bF_0$ introduces an error of at most $\eta c_{\bu}\max_{r\in[m]}\norm{\bbw_r(k)-\bbw_r(0)}{}$ per time, since $\norm{\bsigma(\bW(0),\bx)}{}\le\sqrt m$ and $\norm{\bF_k-\bF_0}{}\le\sqrt{mn}\max_{r\in[m]}\norm{\bbw_r(k)-\bbw_r(0)}{}$. Summation gives $CDT^{9/4}/m$. Lemma~\ref{lemma:bounded-NN-class-func} replaces the remaining frozen kernel on the same covariance event, with error at most $CT(C_1+a_N)$. Thus $e_t=f_t-h_t$ satisfies
\bal\label{eq:refined-NN-pointwise-error}
\supnorm{e_t}\stackrel{\circled{1}}{\le}C\set{\frac{DT^{9/4}}m+T\sqrt{\frac{d\log m}m}+T\sqrt{\frac{\log(2/\delta)}N}}.
\eal
In (\ref{eq:refined-NN-pointwise-error}), the tolerance-dependent size conditions bound each square-root term by $w/3$. The geometric mean of the two width conditions gives $m\gtrsim d\log m\,T^{5/2}/w$, so $D\lesssim d\log m$ and $T^{9/4}\le T^{5/2}$ also bound the first term by $w/3$. The estimates hold uniformly in $\bx,t$. Combining them with the RKHS bound proves the function-class inclusion on the stated event.
\end{proof}

\begin{proof}[\textbf{\textup{Proof of Theorem~\ref{theorem:LRC-population-NN-eigenvalue}}}]
Set $B=B_h+\mu_0$, $M=\sqrt bB$ and $\cG=\set{g:\norm{g}{\cH_{\Kattn}}\le B}$, a fixed class before sampling. From (\ref{eq:LRC-population-NN})--(\ref{eq:varphi-LRC-population-NN}) with $w=0$, splitting at the number of eigenvalues exceeding $r/B^2$ and applying Cauchy--Schwarz gives
\bals
\cfrakR\set{g\in\cG:Pg^2\le r}\stackrel{\circled{1}}{\le}\sqrt2 B R_{\Kattn}(\sqrt r/B)\le C_B R_{\Kattn}(\sqrt r),
\eals
where $C_B=\sqrt2\max\set{1,B}$. The last inequality uses $R_{\Kattn}(a\eps)\le\max\set{1,a}R_{\Kattn}(\eps)$, obtained directly from each truncated eigenvalue.
The loss class $\cL=\set{g^2:g\in\cG}$ is bounded by $M^2$. For $\cT(g^2)=M^2Pg^2$, $\Var{g^2}\le\cT(g^2)$ and the Bernstein constant is $\bar B=M^2$. Squaring is $2M$-Lipschitz on $[-M,M]$, so contraction gives the sub-root majorant
\bals
\psi(r)=2M^3C_BR_{\Kattn}(\sqrt r/M)\stackrel{\circled{2}}{\ge}\bar B\cfrakR\set{\ell\in\cL:\cT(\ell)\le r}.
\eals
The scaling inequality gives the following bound with a fixed constant:
\bals
A_0=\max\set{1,(2C_B/\sigma_0)\max\set{M^3,M^2}},\qquad \psi(r)\le A_0\sigma_0R_{\Kattn}(\sqrt r).
\eals
Lemma~\ref{lemma:sub-root-fix-point-properties}(3) therefore bounds its fixed point by $A_0^2\eps_n^2$. Theorem~\ref{theorem:LRC-population}, with $K_0=5$, yields with probability $1-\exp\pth{-x}$,
$Pg^2\le \tfrac54P_ng^2+C(\eps_n^2+x/n)$ uniformly for $g\in\cG$.
The localized suprema are measurable by separability. Indeed, finite Mercer sums approximate in RKHS norm and uniformly, and shrinking coefficients before rational approximation preserves the norm and positive localization constraints. This gives a countable dense subclass to which the concentration theorem applies.
On the decomposition event write $f_t-f^*=g_t+e_t$, with $g_t\in\cG$ and $\supnorm{e_t}\le w$. Young's inequality, $2\abth{ab}\le a^2/4+4b^2$, gives
$P(g_t+e_t)^2\le \tfrac54Pg_t^2+5w^2$, and $P_ng_t^2\le\tfrac54P_n(g_t+e_t)^2+5w^2$.
Combining these with the population comparison produces three factors $5/4$, and thus
$P(f_t-f^*)^2\le \tfrac{125}{64}P_n(f_t-f^*)^2+C(\eps_n^2+x/n+w^2)$.
Since $125/64<2$, choose $x=n\eps_n^2$. This uniform deterministic-class event applies simultaneously to every iterate on its decomposition event.
On the design event of Lemma~\ref{lemma:hat-eps-eps-relation}, $\hat\eps_n^2\ge c\eps_n^2$. For each design satisfying the stopping and size hypotheses, Theorem~\ref{theorem:bounded-NN-class} gives conditional failure at most $2/n+\delta+C\exp\pth{-cn}+C\exp\pth{-c'n\eps_n^2}$. Integrating this bound over those designs and adding the design-event complements proves the stated implication. The population comparison holds on the fixed RKHS ball, so it applies to every iterate in the decomposition.
\end{proof}

\begin{proof}[\textbf{\textup{Proof of Theorem~\ref{theorem:empirical-loss-bound}}}]
Write $\bfsym{A}=\bKattn_n$, $\bS_t=(\bI_n-\eta \bfsym{A})^t$ and $q_t(z)=1-(1-\eta z)^t$. The assumed representation gives the clean prediction error
$f_t(\bS)-f^*(\bS)=-\bS_tf^*(\bS)+(\bI_n-\bS_t)\bw+\bbe_2(t)$.
Lemma~\ref{lemma:bounded-Ut-f-in-RKHS} and $z(1-\eta z)^{2t}\le1/(2e\eta t)$ bound the signal squared norm divided by $n$ by $\mu_0^2/(2e\eta t)$. Put $\bR_T=\bUattn\diag(q_T(\hlambdaattn_i)^2){\bUattn}^\top$. Monotonicity bounds every noise form at $t\le T$ by $\bw^\top \bR_T\bw/n$. Since $q_T(z)^2\le\min\set{1,\eta Tz}$, the stopping inequality gives
\bals
\frac{\sigma_0^2}n\tr{\bR_T}\stackrel{\circled{1}}{\le}\sigma_0^2\eta T\hat \bR_{\Kattn}^2((\eta T)^{-1/2})\le(\eta T)^{-1}.
\eals
Also $\norm{\bR_T}{}\le1$ and $\tr{\bR_T^2}\le\tr{\bR_T}$. Lemma~\ref{lemma:calibration-quadratic} with $x=cn/(\eta T)$, for a sufficiently small fixed $c>0$, gives
$\bw^\top \bR_T\bw/n\le 2/(\eta T)\le2/(\eta t)$.
This one event fails with probability at most $\exp\pth{-cn/(\eta T)}\le \exp\pth{-cn\hat\eps_n^2}$. The accumulated-error squared norm divided by $n$ is at most $\tau^2\le1/(\eta T)\le1/(\eta t)$. Adding the three squared contributions using $\norm{a+b+c}{}^2\le3(\norm{a}{}^2+\norm{b}{}^2+\norm{c}{}^2)$ proves the stated clean-loss bound simultaneously.
\end{proof}

\subsection{Proof of the Lemmas Required for the Proofs in Section~\ref{sec:proofs-key-results}}
\label{sec:lemmas-main-results}

\begin{lemma}\label{lemma:yt-y-bound}
Let $t \in [0,T]$, $\bv = -\pth{\bI-\eta \bKattn_n}^{t} f^*(\bS)$,
 $\be = -\pth{\bI-\eta \bKattn_n}^{t} \bw$, and $\eta=\Theta(1)\in(0,1)$. Then with probability at least
$1-\exp\pth{-\Theta(n)}$
over the random noise $\bw$,
\bals
\ltwonorm{\bv} + \ltwonorm{\be} \le \pth{\frac{\mu_0}{\min\set{\sqrt{2e \eta},1}}+\sigma_0+1} \cdot
{\sqrt n}.
\eals
\end{lemma}
\begin{proof}
Write $A=\bKattn_n$ and $\bS_t=(\bI_n-\eta A)^t$. For $t\ge1$, the empirical eigenbasis and Lemma~\ref{lemma:bounded-Ut-f-in-RKHS} give

\bals
\norm{\bS_tf^*(\bS)}{}^2\le \frac1{2e\eta t}\sum_{i=1}^n\frac{([{\bUattn}^\top f^*(\bS)]_i)^2}{\hlambdaattn_i}\le\frac{n\mu_0^2}{2e\eta t}.
\eals
The scalar bound is $z(1-\eta z)^{2t}\le z\exp\pth{-2\eta tz}\le1/(2e\eta t)$. At zero reproduction gives $\norm{f^*(\bS)}{}\le\mu_0\sqrt{bn}\le\mu_0\sqrt n$. Lemma~\ref{lemma:calibration-quadratic} with $\bM=\bI_n$ and $x=cn$ bounds $\norm{\bw}{}^2$ by $\sigma_0^2n(1+2\sqrt c+2c)\le n(\sigma_0+1)^2$, choosing $c>0$ sufficiently small for the fixed proxy. Since $\norm{\bS_t}{}\le1$, this one event bounds the noise for all $t$. Adding both bounds proves the assertion simultaneously. The proxy is used as an upper bound on variance.
\end{proof}

\begin{lemma}
\label{lemma:empirical-loss-convergence-contraction}
Recall that $b=\kappa^2/2$. Fix the training design and an initialization $\bW(0)\in\cW_0$. Let $\delta\in(0,1)$, $\eta\in(0,1)$ and $T\ge1$ be an integer. Suppose $N\ge1$ and
$m\gtrsim\max\set{\eta c_{\bu}(d^2+\log^2m)T^2,d\log m,n^{1/(2d)}}$.
There is one event of probability at least $1-\delta$ over $\bQ$ on which, simultaneously for every $0\le t<T$, the following implication holds. If $\ltwonorm{\hat\by(k)-\by}\le c_{\bu}\sqrt n$ for every $0\le k\le t$, then
\bals
\hat \by(t+1) - \by  = \pth{\bI- \eta \bKattn_n }\pth{\hat \by(t) - \by} +\bE(t+1),
\eals
where $\ltwonorm {\bE(t+1)} \lsim \bE_{m,\eta,\delta}$,
and $\bE_{m,\eta,\delta}$ is defined by
\bal\label{eq:empirical-loss-Et-bound-Em}
\bE_{m,\eta,\delta} \defeq   R^2\sqrt {mn}/T
+ \eta c_{\bu}  {\sqrt n} \pth{R+C_1(m,d,1/n) +  \sqrt{\frac{\log (2/{\delta})}{N}}}.
\eal
\end{lemma}

\begin{proof}
The residual hypothesis and the deterministic movement lemma give motion at most $R$ and $\abth{a_r}\le2R$ through $t+1$. Put $\bF_t=\bsigma(\bW(t),\bS)$ and $\bG=\bF_0$. The good event gives $\norm{\bG}{}\le\sqrt{mn}$, $\norm{\bF_t-\bG}{}\le R\sqrt{mn}$ and $\norm{\bA}{}\le(b+C_1)b\le1$. The size condition ensures $R\le1$. The exact simultaneous increment is
\bals
\hat\by(t+1)-\hat\by(t)=-\eta \bF_t^\top\bA^2\bF_t\bu(t)/(nm)+\bfm{r}_t,\\
(\bfm{r}_t)_i=m^{-1/2}(\bsigma(\bW(t+1),\bbx_i)-\bsigma(\bW(t),\bbx_i))^\top\bA\ba(t+1).
\eals
The one-step bound (\ref{eq:first-layer-single-increment}) is $R/T$, while $\norm{\ba(t+1)}{}\le2R\sqrt m$. ReLU Lipschitz continuity and Cauchy--Schwarz therefore give $\norm{\bfm{r}_t}{}\le2R^2\sqrt{mn}/T$. These inequalities use the one-step movement and operator norm bounds.
Telescoping the two outer feature factors yields
$\norm{(\bF_t^\top\bA^2\bF_t-\bG^\top\bA^2\bG)/(nm)}{}\le 2R+R^2\le3R$.
Lemma~\ref{lemma:empirical-loss-convergence-contraction-bKattn-approx} bounds the initialization-kernel error by $C(C_1+a_N(\delta))$. Multiplying by $\eta\norm{\bu(t)}{}\le\eta c_{\bu}\sqrt n$ and adding $\bfm{r}_t$ proves (\ref{eq:empirical-loss-Et-bound-Em}). The event is uniform over the times and residual vectors in the hypothesis for every $N\ge1$.
\end{proof}

\begin{lemma}\label{lemma:empirical-loss-convergence-contraction-bKattn-approx}
Recall that $b=\kappa^2/2$. Fix $\delta\in(0,1)$. Suppose $\bW(0)\in\cW_0$ and $m\gtrsim\max\set{d\log m,n^{1/(2d)}}$. Then with probability at least $1-\delta$ over $\bQ$,
\bals
\ltwonorm{\frac{1}{nm}\bsigma^{\top}(\bW(0),\bS)
\bA^2 \bsigma(\bW(0),\bS) - \bKattn_n} \lsim C_1(m,d,1/n) +  \sqrt{\frac{\log (2/{\delta})}{N}}.
\eals
\end{lemma}
\begin{proof}

\smallskip
\noindent Let $\bF=\bsigma(\bW(0),\bQ)$, $\bG=\bsigma(\bW(0),\bS)$ and $\bL=(K^{(s)}(\bbq_i,\bbq_j))_{i,j\le N}$. Put $\bH=\bF^\top \bF/m$ and $\bJ=\bG^\top \bF/m$. The attention matrix is $\bA=\bF\bL\bF^\top/(mN^2)$. Hence the exact normalized initialization Gram matrix is
\bals
\frac{\bG^\top\bA^2\bG}{nm}\stackrel{\circled{1}}{=}\frac{\bJ \bL \bH \bL \bJ^\top}{nN^4}.
\eals
Here \circled{1} uses the exact attention-matrix normalization.

On the good feature event each entry of $\bH-K(\bQ,\bQ)$ and $\bJ-K(\bS,\bQ)$ is at most $C_1$. The entries of all kernel factors have absolute value at most $1$, by the width condition. Telescoping the three random-feature factors and summing the $N^4$ products gives an entrywise error at most $3C_1$ for the unnormalized kernel. The normalized Gram error is at most $3C_1$ in operator norm.

\smallskip
\noindent The population-feature empirical product has kernel
$\tK_N(\bx,\bx')=N^{-4}K(\bx,\bQ)\bL K(\bQ,\bQ)\bL K(\bQ,\bx').$
Section~\ref{sec:calibration} proves the uniform estimate $\sup_{\bx,\bx'}\abth{\tK_N(\bx,\bx')-\Kattn(\bx,\bx')}\le4b^4a_N(\delta)$ for the four averaging positions using the same sample. It also bounds the normalized Gram error. Combining the feature and covariance estimates with $a_N(\delta)\lesssim\sqrt{\log(2/\delta)/N}$ proves the result uniformly over the training points.
\end{proof}

\begin{lemma}[Movement bounds]\label{lemma:two-layer-weight-vector-movement}\label{lemma:second-layer-weight-vector-movement}\label{lemma:first-layer-weight-vector-movement}
Fix $T\ge1$, $0\le t<T$, $N\ge1$, $\eta=\Theta(1)\in(0,1)$ and $\bW(0)\in\cW_0$. Suppose $\ltwonorm{\bu(k)}\le c_{\bu}\sqrt n$ for $0\le k\le t$.
The following assertions are deterministic for every sample $\bQ$.
\begin{enumerate}[label=(\roman*),leftmargin=18pt]
\item If $\abth{a_r(k)}\le1$ for $k\le t$ and every $r$, then
\bal\label{eq:first-layer-weight-bound}
\max_{r\in[m]}\ltwonorm{\bbw_r(k)-\bbw_r(0)}\le R,\qquad 0\le k\le t+1.
\eal
\item If first-layer motion is at most one for $k\le t$ and $m\gtrsim\max\set{d\log m,n^{1/(2d)}}$, then
\bal\label{eq:second-layer-weight-bound}
\max_{r\in[m]}\abth{a_r(k)}\le\frac{2\eta c_{\bu}(4d+6\log(2mn))T}{\sqrt m}=2R,\qquad 0\le k\le t+1.
\eal
\item If $m\gtrsim\max\set{(d^2+\log^2m)T^2,d\log m,n^{1/(2d)}}$, both conclusions hold without either motion hypothesis.
\end{enumerate}
\end{lemma}
\begin{proof}
Put $D=4d+6\log(2mn)$, $\bF=\bsigma(\bW(0),\bQ)$ and $\bG=\bsigma(\bW(0),\bS)$. Under the width condition in (ii) or (iii), the good feature event gives $\ltwonorm{\bG}\le\sqrt{mn}$ and $\ltwonorm{\bA}\le(b+C_1(m,d,1/n))b\le1$. Part (i) uses only the following entry bound, which follows from the row-norm event and the attention factorization for every $m,N\ge1$:
\bal\label{eq:second-layer-weight-vector-movement-seg1-1}
\abth{\bA_{rr'}}\le D/m,\qquad \ltwonorm{\bA_r}\le D/\sqrt m.
\eal
Indeed $\ltwonorm{\bF_r}\le\sqrt{ND}$ and $\ltwonorm{\bK_N^{(s)}}\le1$, so Cauchy--Schwarz in $\bA=\bF\bK_N^{(s)}\bF^\top/(mN)$ proves the entry bound. If $\abth{a_{r'}(k)}\le1$, then $\abth{[\bA\ba(k)]_r}\le D$. Each first-layer update block therefore satisfies
\bal\label{eq:first-layer-single-increment}
\ltwonorm{\bbw_r(k+1)-\bbw_r(k)}\le \frac{\eta c_{\bu}D}{\sqrt m}=R/T.
\eal
The increment estimate uses unit inputs and $n^{-1}\sum_{i=1}^n\abth{u_i(k)}\le c_{\bu}$. Summing at most $T$ updates proves (\ref{eq:first-layer-weight-bound}).
Under unit first-layer motion, ReLU Lipschitz continuity gives $\ltwonorm{\bsigma(\bW(k),\bS)}\le2\sqrt{mn}$. The exact second-layer update and (\ref{eq:second-layer-weight-vector-movement-seg1-1}) give
$\abth{a_r(k+1)-a_r(k)}\le \frac\eta{n\sqrt m}\ltwonorm{\bA_r}\ltwonorm{\bsigma(\bW(k),\bS)}\ltwonorm{\bu(k)}\le2\eta c_{\bu}D/\sqrt m$.
Summing from $a_r(0)=0$ proves (\ref{eq:second-layer-weight-bound}). For (iii), $\log(2mn)\lesssim d\log m$ and the size condition makes $2R\le1$. Both bounds hold at zero. If they hold through $k$, (i) uses only output weights through $k$ and (ii) uses only first-layer motion through $k$ to give the two new bounds at $k+1$. This proves (iii) by simultaneous induction.
\end{proof}

\begin{lemma}\label{lemma:bounded-Linfty-vt-sum-et}
Fix $\eta\in(0,1)$ and a positive integer $T\le\hat T$. For $0\le t'<T$, define
\bals
v(\cdot,t')=\frac\eta n\sum_{j=1}^n\Kattn(\bbx_j,\bx)\bv_j(t'),\quad \hat e(\cdot,t')=\frac\eta n\sum_{j=1}^n\Kattn(\bbx_j,\bx)\be_j(t').
\eals
For $t\in[T]$, let $h(\cdot,t')=v(\cdot,t')+\hat e(\cdot,t')$ and $h(\cdot)=\sum_{t'=0}^{t-1}h(\cdot,t')$. Assume $\bv(t')\in\cV_{t'}$ and $\be(t')\in\cE_{t',\tau}$ for every $0\le t'\le t-1$.
Suppose $\tau\le1/(\eta T)$.
For each fixed training design, one noise event of probability at least $1 - \exp\pth{-\Theta(n \hat\eps_n^2)}$ bounds every such sum simultaneously, and on this event $\norm{h}{\cHKattn}\le B_h$, with $B_h$ defined in (\ref{eq:B_h}).
\end{lemma}
\begin{proof}
Write $\bfsym{A}=\bKattn_n$, $\bS_t=(\bI_n-\eta \bfsym{A})^t$, $\Phi \bc=\sum_{i=1}^n c_i\Kattn(\cdot,\bbx_i)$ and $q_t(z)=1-(1-\eta z)^t$. Then $\Phi^*\Phi=n\bfsym{A}$. The signal sum has squared norm
$n^{-1}\sum_{i=1}^n\frac{q_t(\hlambdaattn_i)^2}{\hlambdaattn_i}([{\bUattn}^\top f^*(\bS)]_i)^2\le \mu_0^2$,
by $0\le q_t\le1$ and Lemma~\ref{lemma:bounded-Ut-f-in-RKHS}. The noise quadratic form has eigenvalues $q_t(\hlambdaattn_i)^2/(n\hlambdaattn_i)$, increasing with $t$. For $a=\eta T$, $q_T(z)^2/z\le a\min\set{1,az}$ and the stopping inequality give
\bals
\frac{\sigma_0^2}n\sum_{i=1}^n\frac{q_T(\hlambdaattn_i)^2}{\hlambdaattn_i}
\stackrel{\circled{1}}{\le}\sigma_0^2a^2\hat R_{\Kattn}^2(a^{-1/2})\le1.
\eals
The matrix norm is at most $a/n$ and its squared Frobenius norm is at most $(a/n)$ times its trace. Lemma~\ref{lemma:calibration-quadratic} with $x=cn/a$ bounds this form by $2$, for a sufficiently small fixed $c>0$. Since $a^{-1}\ge\hat\eps_n^2$, failure is at most $\exp\pth{-cn\hat\eps_n^2}$, and monotonicity gives one event for all $t\le T$.
Every remaining residual vector has norm at most $\sqrt n\tau$. As $\norm{\Phi}{}\le\sqrt{nb}$, their total RKHS contribution is at most $\eta T\sqrt b\tau\le1$. The triangle inequality yields $\mu_0+\sqrt2+1=B_h$. The estimate is uniform over the allowed residual sums, including those depending on the observations.
\end{proof}

\begin{lemma}\label{lemma:bounded-NN-class-func}
Recall that $b=\kappa^2/2$ and write $J_m(\bx,\bx')=m^{-1}\bsigma(\bW(0),\bx)^\top\bA^2\bsigma(\bW(0),\bx')$. Fix $\delta\in(0,1)$. Suppose $\bW(0)\in\cW_0$, $m\gtrsim\max\set{d\log m,n^{1/(2d)}}$ and $\ltwonorm{\bu(t')}\le c_{\bu}\sqrt n$ with $c_{\bu}=\Theta(1)$. Then with probability at least $1-\delta$ over $\bQ$,
\bals
\supnorm{\frac1n\sum_{j=1}^n[J_m(\bx,\bbx_j)-\Kattn(\bx,\bbx_j)]\bu_j(t')}\lsim C_1(m,d,1/n)+\sqrt{\frac{\log(2/\delta)}N}.
\eals
\end{lemma}
\begin{proof}
The proof of Lemma~\ref{lemma:empirical-loss-convergence-contraction-bKattn-approx} applies with arbitrary $\bx,\bx'$, because the good feature event and the covariance concentration event are uniform on $\cX$. It gives $\sup_{\bx,\bx'}\abth{J_m(\bx,\bx')-\Kattn(\bx,\bx')}\le  C(C_1+a_N(\delta))$. Therefore the expression in the lemma is bounded by
$\frac1n\sum_{j=1}^n\abth{u_j(t')}\,C(C_1+a_N(\delta))
\le c_{\bu}C(C_1+a_N(\delta))$.
The estimate is uniform over residual vectors satisfying the norm bound, including sample- and noise-dependent vectors.
\end{proof}

\begin{lemma}[In the proof of
{\cite[Lemma 8]{RaskuttiWY14-early-stopping-kernel-regression}}]
\label{lemma:bounded-Ut-f-in-RKHS}
For any $f \in \cH_{K}(\mu_0)$, we have
\bals
\frac 1n \sum_{i=1}^n \frac{\bth{\bU^{\top}f(\bS)}_i^2}{\hlambda_i} \le \mu_0^2.
\eals
\end{lemma}
\begin{proof}
Let $E:\cH_K\to\RR ^n$ be $Ef=(f(\bbx_i))_i$. Then $EE^*=\bK=n\bU\diag(\hlambda_i)\bU^\top$. The orthogonal projection of $f$ onto $\operatorname{ran}(E^*)$ is $E^*\bK^{-1}Ef$ and has squared norm $(Ef)^\top\bK^{-1}Ef\le \norm{f}{\cH_K}^2$. Expanding the inverse yields the displayed inequality. The same proof applies with $K$ replaced by $\Kattn$. If a Gram matrix is singular, the Moore--Penrose inverse gives the identical assertion on positive eigenvalues and $Ef$ lies in its range.
\end{proof}

\begin{lemma}
\label{lemma:hat-eps-eps-relation}
For sufficiently large $n$ depending on the fixed law and model parameters, with probability at least $1-4\exp(-\Theta(n\eps_n^2))$,
$\eps_n^2 \lsim\hat \eps_n^2, \quad \hat \eps_n^2 \lsim \eps_n^2$.
Similarly, with probability at least $1-4\exp(-\Theta(n\eps_{K,n}^2))$,
$\eps_{K,n}^2 \lsim \hat \eps_{K,n}^2, \quad \hat \eps_{K,n}^2 \lsim \eps_{K,n}^2$.
\end{lemma}

\begin{proof}
We prove the attention assertion. Let $u_i=\Kattn(\cdot,\bbx_i)$. Define $G=\Expect{}{u_i\otimes u_i}$ and $G_n=n^{-1}\sum_{i=1}^n u_i\otimes u_i$. These positive trace-class covariances act on the separable RKHS $\cHKattn$, with $\norm{u_i}{}^2\le b^{s'}$. Their positive spectra are respectively $(\lambdaattn_j)$ and $(\hlambdaattn_i)$: the inclusion $\cHKattn\to L^2(\mu)$ and the normalized evaluation map have these covariances and the corresponding integral or Gram operators as their opposite-order compositions.
Put $r=\eps_n^2$, $\hat r=\hat\eps_n^2$, $S(v)=\sum_{j=1}^{\infty}\min\set{\lambdaattn_j,v}$ and $S_n(v)=\sum_{i=1}^n\min\set{\hlambdaattn_i,v}$. Then $S(r)=nr^2/\sigma_0^2$. Define $d(r)=\tr{G(G+rI)^{-1}}$ and $F=\tr{G_n(G_n+rI)^{-1}}$. Scalar spectral calculus gives
$d(r)\le S(r)/r\le2d(r)$, and $F\le S_n(r)/r\le2F$.
Here $z/(z+r)\le\min\set{z/r,1}\le2z/(z+r)$. By Proposition~\ref{prop:infinite-spectrum-radius}, $nr\to\infty$, so $\ell=b^{s'}/(nr)\le1$ eventually.
Remove observation $i$, retaining normalization $n$, and write $G_{-i}=G_n-n^{-1}u_i\otimes u_i$ and $\ell_i=n^{-1}\iprod{u_i}{(G_{-i}+rI)^{-1}u_i}\le\ell$. The rank-one resolvent formula gives
\bals
n^{-1}\iprod{u_i}{(G_n+rI)^{-1}u_i}\stackrel{\circled{1}}{=}\frac{\ell_i}{1+\ell_i},\qquad F=\sum_{i=1}^n\frac{\ell_i}{1+\ell_i}.
\eals
Independence of $u_i,G_{-i}$ and inverse Jensen give
\bals
\Expect{}{\sum_{i=1}^n\ell_i}=\tr{G\Expect{}{(G_{-1}+rI)^{-1}}} \ge\tr{G((1-1/n)G+rI)^{-1}}\ge d(r).
\eals
Inverse Jensen follows from $\iprod{v}{A^{-1}v}=\sup_h\set{2\iprod{v}{h}-\iprod{h}{Ah}}$ for $A\ge rI$. To obtain the upper expectation bound, apply inverse Jensen to $G_n+rI$, multiply the bounded inverse inequality by $r$ and subtract it from $I$. In quadratic-form order this gives $\Expect{}{G_n(G_n+rI)^{-1}}\stackrel{\circled{2}}{\le}G(G+rI)^{-1}$. Take diagonal quadratic forms in a fixed orthonormal basis and sum them by Tonelli to obtain $\Expect{}{F}\le d(r)$. These are traces of positive operators and are finite because $G(G+rI)^{-1}\le G/r$. Likewise, nonnegative diagonal terms of $G^{1/2}A^{-1}G^{1/2}$ justify the earlier expectation interchange by Tonelli. Thus
$\frac{d(r)}{1+\ell}\le \Expect{}{F}\le d(r)$.
For concentration define $F_i=\tr{G_{-i}(G_{-i}+rI)^{-1}}$. A second resolvent calculation yields
\bals
\Delta_i=F-F_i=
\frac{r\norm{(G_{-i}+rI)^{-1}u_i}{}^2}{n(1+\ell_i)}
=\frac{r(1+\ell_i)}n\norm{(G_n+rI)^{-1}u_i}{}^2.
\eals
The identity $(G_{-i}+rI)^{-1}u_i=(1+\ell_i)(G_n+rI)^{-1}u_i$ gives the second expression. Since $r(A+rI)^{-2}\le(A+rI)^{-1}$ for positive $A$,
\bals
0\le\Delta_i\le \frac{\ell_i}{1+\ell_i}\le\ell,\qquad
\sum_{i=1}^n\Delta_i\stackrel{\circled{3}}{\le}(1+\ell)F\le2F.
\eals
The second inequality sums the last expression for $\Delta_i$ and uses the trace identity for $F$. Hence $F/\ell$, with omitted-observation versions $F_i/\ell$, has increments in $[0,1]$ summing to at most $2F/\ell$. Their squared increments have the same upper bound, since squaring decreases each increment. These facts verify both the self-bounding and weak self-bounding hypotheses with parameters $(2,0)$. The self-bounding inequalities~\cite[Chapter 6]{BLM13} therefore give, for a deviation $u>0$ in the upper tail and $0<u\le\Expect{}{F}$ in the lower tail,
\bals
\Pr\set{F-\Expect{}{F}\ge u}\le\exp\left(-\frac{u^2}{4\ell\Expect{}{F}+(5/3)\ell u}\right),\qquad
\Pr\set{\Expect{}{F}-F\ge u}\le\exp\left(-\frac{u^2}{4\ell\Expect{}{F}}\right).
\eals
The upper choice $u=\Expect{}{F}$ gives exponent $3\Expect{}{F}/(17\ell)$, and the lower choice $u=\Expect{}{F}/2$ gives exponent $\Expect{}{F}/(16\ell)$. A union bound consequently yields
$\Pr\set{F<\Expect{}{F}/2\text{ or }F>2\Expect{}{F}}\le 2\exp\pth{-\Expect{}{F}/(16\ell)}$.

Measurability and integrability hold since $F$ is a continuous function of a finite Gram matrix and $F\le n$. From the scalar comparisons, $d(r)\ge nr/(2\sigma_0^2)$, and the expectation bound gives $\Expect{}{F}/\ell\ge n^2r^2/(4\sigma_0^2b^{s'})$. As $nr\to\infty$, the failure is at most $2\exp\pth{-cnr}$. On the complementary event $d(r)/4\le F\le2d(r)$, so
$S(r)/8\le S_n(r)\le4S(r),$ and $1/\sqrt8\le q(r)\defeq\sigma_0\sqrt{S_n(r)/n}/r\le2.$
Now $S_n(v)/v$ is nonincreasing, so $q(v)\sqrt v$ is nonincreasing. It follows that
$q(r/8)\ge \sqrt8q(r)\ge1$, and $q(4r)\le q(r)/2\le1$.
The nonzero empirical spectrum makes $q$ continuous and strictly decreasing from infinity to zero, and its unique unit crossing is $\hat r$. Consequently $r/8\le\hat r\le4r$. Replacing the attention kernel, its spectra, $r$ and $b^{s'}$ by $K$, its spectra, $\eps_{K,n}^2$ and $b$ proves the second assertion. The two assertions hold on their respective design events.
\end{proof}

\begin{lemma}
[{\cite[Lemma B.9]{yang2024gradientdescentfindsoverparameterized}}]
\label{lemma:sub-root-fix-point-properties}
Suppose $\psi \colon [0,\infty) \to [0,\infty)$ is a sub-root function that is not identically zero, with unique positive fixed point $r^*$. Then the following properties hold.

\begin{itemize}[leftmargin=8pt]
\item[(1)] Let $a \ge 0$. Then $\psi(r) + a$ as a function of $r$ is also a sub-root function with fixed point $r^*_a$, and
$r^* \le r^*_a \le r^* + 2a$.
\item[(2)] Let $b \ge 1$, $c \ge 0$. Then $\psi(br+c)$ as a function of $r$ is also a sub-root function with fixed point $r^*_b$, and
$r^*_b \le br^* +2c/b$.
\item[(3)] Let $b \ge 1$. Then $\psi_b(r) = b\psi(r)$ is also a sub-root function with fixed point $r^*_b$, and
$r^*_b \le b^2r^* $.
\end{itemize}
\end{lemma}
\begin{proof}
Sub-root monotonicity gives $\psi(u)\le\sqrt{ur^*}$ for $u\ge r^*$. The ratios of $\psi(r)+a$ and $b\psi(r)$ to $\sqrt r$ are nonincreasing. For the composition, write the ratio as $[\psi(br+c)/\sqrt{br+c}]\sqrt{b+c/r}$, a product of nonnegative nonincreasing functions. Each modification is also nondecreasing and nonzero, so its positive fixed point is unique.

For (1), take $u=r^*+2a$. Since $\sqrt{ur^*}\le r^*+a$, we have $\psi(u)+a\le u$. The fixed-point crossing therefore gives $r_a^*\le u$, while $\psi+a\ge\psi$ gives $r_a^*\ge r^*$.
For (2), take $u=br^*+2c/b$. Then $bu+c\ge r^*$ and $u^2-r^*(bu+c)=cr^*+4c^2/b^2\ge0$, so
$\psi(bu+c)\le \sqrt{r^*(bu+c)}\le u$.
The crossing gives $r_b^*\le u$.
For (3), $u=b^2r^*$ gives $b\psi(u)\le b\sqrt{ur^*}=u$, and the same crossing argument proves $r_b^*\le b^2r^*$.
\end{proof}

\subsection{Proofs for the Approximate Uniform Convergence for the Kernel $\Kattn$}

The operator event in Section~\ref{sec:calibration} controls the kernel approximations before training points are selected.
\begin{theorem}
\label{theorem:hatKattn-close-to-Kattn-supnorm}
For every fixed $\bx' \in \cX$ and every $\delta \in (0,1)$, with probability at least
$1-\delta$ over $\bQ = \set{\bbq_i}_{i=1}^N$, we have
\bals
\supnorm{\hKattn(\bx,\bx') - \Kattn(\bx,\bx')} \lsim \sqrt{\frac{\log (2/\delta)}{N}}.
\eals
As a result, with probability at least
$1-\delta$ over $\bQ$,
\bals
\sup_{\bx \in \cX,i \in [n]}\abth{\hKattn(\bx,\bbx_i) - \Kattn(\bx,\bbx_i)} \lsim  \sqrt{\frac{\log (2/{\delta})}{N}},  \\
\ltwonorm{\hbKattn - \bKattn} \lsim n \sqrt{\frac{\log (2/{\delta})}{N}}.
\eals
\end{theorem}

The empirical intermediate kernel $\hKint$ defined in the setup concentrates on the same covariance event.

\begin{theorem}
\label{theorem:hatKint-close-to-Kint-supnorm}
For every fixed $\bx' \in \cX$ and every $\delta \in (0,1)$, with probability at least
$1-\delta$ over $\bQ = \set{\bbq_i}_{i=1}^N$, we have
\bals
\supnorm{\hKint(\bx,\bx') - \Kint(\bx,\bx')} \lsim \sqrt{\frac{\log (2/\delta)}{N}}.
\eals
As a result, with probability at least
$1-\delta$ over $\bQ$,
\bals
\sup_{\bx \in \cX,i \in [n]}\abth{\hKint(\bx,\bbx_i) - \Kint(\bx,\bbx_i)} &\lsim  \sqrt{\frac{\log (2/{\delta})}{N}},  \\
\ltwonorm{\hbKint - \bKint} &\lsim n \sqrt{\frac{\log (2/{\delta})}{N}}.
\eals
\end{theorem}
\begin{proof}[\textbf{\textup{Proof of both kernel approximation theorems}}]
Use $k_{\bx},C,C_N$ from Section~\ref{sec:calibration}, and put $B=C^{s-1}C^2C^{s-1}$. Substituting the section representation into the kernel definitions gives multipliers $C_NBC_N$ and $CBC$ for $\hKattn,\Kattn$, and $C^{s-1}C_N^2C^{s-1}$ and $B$ for $\hKint,\Kint$. On the common covariance event,
$\norm{C_NBC_N-CBC}{}\le 2b^3a_N(\delta)$, and $\norm{C_N^2-C^2}{}\le 2ba_N(\delta)$.
Ordered telescoping gives these bounds from $\norm{C_N}{},\norm{C}{}\le b$, $\norm{C^{s-1}}{}\le1$ and $\norm{B}{}\le b^2$. The section norms $\sqrt b$ give uniform kernel errors $2b^4a_N$ and $2b^2a_N$. Since entrywise error $a$ gives matrix operator error at most $na$, the same event proves the Gram bounds.
\end{proof}

\section{More Results about Uniform Convergence and Integral Operators}
\label{sec:RKHS-Kattn-more-results}

\subsection{Results about Eigenvalues of the Integral Operators}
\label{sec:RKHS-Kattn-more-results-repeat-1}

\begin{theorem}
\label{theorem:Ks-bounded}
The RHS of (\ref{eq:Ks-def}), $\sum\limits_{j=1}^{\infty} \lambda_j^{s} e_j(\bx)e_j(\bx')$,
converges uniformly on $\cX \times \cX$ with $s \ge 1$, and $K^{(s)}$ is well-defined. Moreover, $\sup_{\bx,\bx' \in \cX} \abth{K^{(s)}(\bx,\bx')} < 1$.
\end{theorem}
\begin{proof}
Mercer's theorem gives diagonal tails $D_J(\bx)=\sum_{j=J+1}^{\infty}\lambda_je_j(\bx)^2$ tending uniformly to zero. Since $0<\lambda_j\le b<1$, Cauchy--Schwarz gives
$\sup_{\bx,\bx'}\abth{\sum_{j=J+1}^{\infty}\lambda_j^se_j(\bx)e_j(\bx')}\le b^{s-1}\sup_{\bx} D_J(\bx)\longrightarrow0.$
Thus $K^{(s)}$ converges uniformly and absolutely, is continuous and positive semidefinite as a limit of finite positive Gram sums. Its diagonal is at most $b^{s-1}K(\bx,\bx)=b^s$, and Cauchy--Schwarz for the full series gives $\abth{K^{(s)}(\bx,\bx')}\le b^s<1$.
\end{proof}

\begin{theorem}
\label{theorem:spectrum-Kattn}
On the positive spectral subspace of $T_K$, the operators associated with $K^{(s)}$, $\Kint$ and $\Kattn$ are respectively $T_K^s$, $T_K^{2s+1}$ and $T_K^{2s+3}$. In particular $T_{\Kattn}e_j=\lambda_j^{s'}e_j$ and $\lambdaattn_j=\lambda_j^{s'}$, where $s'=2s+3$. The null space of $T_K$ remains the null space of $T_{\Kattn}$.
\end{theorem}
\begin{proof}
Bounded kernels and $\int\abth{f}\,d\mu\le\norm{f}{L^2(\mu)}$ make every defining iterated integral absolutely integrable, so Fubini applies. Uniform Mercer convergence identifies $T_{K^{(s)}}=T_K^s$. The two intermediate and two outer integrations then give
\bals
T_{\Kint}=T_K^sT_KT_K^s=T_K^{2s+1},\qquad
T_{\Kattn}\stackrel{\circled{1}}{=}T_KT_{\Kint}T_K=T_K^{2s+3}.
\eals
Applying these identities to positive eigenfunctions gives the asserted spectrum. Functional calculus maps zero to zero and positive eigenvalues to positive values, preserving the null space and positive spectral subspace.
\end{proof}

\begin{proposition}
\label{proposition:Kattn-gram-nonsingular}
\looseness=-1
If $\bx_1,\ldots,\bx_n\in\cX$ satisfy $\bbx_i\ne\pm \bbx_j$ for $i\ne j$, the Gram matrices of $K$ and $\Kattn$ are nonsingular.
\end{proposition}
\begin{proof}

\smallskip
\noindent Suppose $\bc^\top\bK \bc=0$. The feature covariance representation implies $\sum_{i=1}^{n} c_i\sigma(\bw^\top \bbx_i)=0$ for Gaussian-almost every $\bw$. The left side is continuous and the Gaussian density is strictly positive everywhere, so it vanishes for every $\bw$. For a fixed $i$, the hyperplane $H_i=\set{\bw:\bw^\top \bbx_i=0}$ differs from each $H_j$. Choose $\bw_0\in H_i\setminus\bigcup_{j\ne i}H_j$. Such a point exists because finitely many proper linear subspaces cannot cover $H_i$. Along $\bw_0+t \bbx_i$, all terms except the $i$-th are affine for sufficiently small $\abth{t}$. The $i$-th term has derivative jump $c_i\ltwonorm{\bbx_i}^2=c_i$. The identically zero function has no derivative jump, whence $c_i=0$. This holds for every $i$.

\smallskip
\noindent Mercer expansion gives $\bc^\top\bKattn \bc=\sum_{j=1}^{\infty}\lambda_j^{s'}(\sum_{i=1}^{n} c_i e_j(\bbx_i))^2$. If this is zero, every coordinate sum is zero, because each $\lambda_j$ is positive. It follows that $\bc^\top\bK \bc=\sum_{j=1}^{\infty}\lambda_j(\sum_{i=1}^{n} c_i e_j(\bbx_i))^2=0$. The preceding argument gives $\bc=0$.
\end{proof}

\begin{proposition}\label{prop:infinite-spectrum-radius}
Under the stated non-atomic covariate law, $K$ and $\Kattn$ have infinitely many positive eigenvalues. The critical population rate $\eps_n$ of $\Kattn$ satisfies $\eps_n\to0$ and $n\eps_n^2\to\infty$. The same assertions hold for the critical population rate $\eps_{K,n}$ of $K$.
\end{proposition}
\begin{proof}
Non-atomicity makes every finite set of antipodal pairs null, so the support contains arbitrarily large non-antipodal samples. Proposition~\ref{proposition:Kattn-gram-nonsingular} makes both Gram matrices nonsingular at every such size. Both RKHSs, and therefore their Mercer spectra, are infinite dimensional.
Trace summability and the fixed-point equation give
$\eps_n^4\le \frac{\sigma_0^2}n\sum_{j=1}^{\infty}\lambdaattn_j\to0$, and $n\eps_n^2=\sigma_0^2\sum_{j=1}^{\infty}\min\set{\lambdaattn_j/\eps_n^2,1}$.
Thus $\eps_n\to0$. For every fixed $J$, eventually $\eps_n^2\le\lambdaattn_J$, making the second expression at least $\sigma_0^2J$. Since $J$ is arbitrary, $n\eps_n^2\to\infty$. The same proof applies to $K$.
\end{proof}

\subsection{Proofs of Theorem~\ref{theorem:sup-hat-g} }
\label{sec:proofs-uniform-convergence-sup-hat-g-V_R}

\begin{lemma}[Gaussian feature product and net bounds]\label{lemma:Gaussian-feature-net}
Let $\bw_r\sim\cN(0,\kappa^2\bI_d)$ be independent, with $d\ge1$, and put $F_m(\bu,\bv)=m^{-1}\sum_{r=1}^m\sigma(\bw_r^\top \bu)\sigma(\bw_r^\top \bv)$ for unit $\bu,\bv$. For every $z>0$,
\bals
\Pr\set{\abth{F_m(\bu,\bv)-K(\bu,\bv)}>\kappa^2(8\sqrt{z/m}+4z/m)}\le2\exp\pth{-z}.
\eals
If $\delta\in(0,1)$ and $m\ge\max\set{2,\log(4/\delta)}$, set $\ell=2d\log(1+2m)+\log(8/\delta)$. On one event of probability at least $1-\delta/2$,
\bals
\frac{\fnorm{\bW(0)}^2}{m}\le5\kappa^2d,\qquad
\sup_{\bu,\bv}\abth{F_m(\bu,\bv)-K(\bu,\bv)}\le\kappa^2\left(8\sqrt{\ell/m}+\frac{4\ell+20d}m\right).
\eals
This event is independent of the training design and measurable. The statement applies in any ambient dimension, in particular after replacing $d$ by $d+1$.
\end{lemma}
\begin{proof}
For $A=\bw^\top \bu$, $B=\bw^\top \bv$ and $X=\sigma(A)\sigma(B)$, Cauchy--Schwarz and the Gaussian even moments give, for every integer $k\ge2$,
\bals
\Expect{}{\abth{X}^k}\stackrel{\circled{1}}{\le}\kappa^{2k}(2k-1)!!\le k!(2\kappa^2)^k,\qquad
\Expect{}{\abth{X-\Expect{}{X}}^k}\le k!(4\kappa^2)^k.
\eals
Here \circled{1} needs no independence of $A,B$, and the centered estimate follows by convexity and centering. With $Y=X-\Expect{}{X}$ and $L=4\kappa^2$, expansion of the exponential series yields
$\log\Expect{}{\exp\pth{\theta Y}}\le \theta^2L^2/(1-\abth{\theta}L)$ for $\abth{\theta}L<1$.
Independence of the rows multiplies the numerator by $m$. Exponential Markov with $\theta=\sqrt{z/(mL^2)}/(1+L\sqrt{z/(mL^2)})$, and then the same argument for $-Y$, gives the scalar bound.
An $m^{-1}$-net of the unit sphere has at most $(1+2m)^d$ points, by comparing volumes of disjoint radius-$1/(2m)$ balls in a maximal separated set. A union bound over ordered pairs with $z=\ell$ fails with probability at most $\delta/4$.
For $v=\log(4/\delta)$, the chi-square tail gives, with failure at most $\delta/4$,
$\fnorm{\bW(0)}^2\le \kappa^2(md+2\sqrt{mdv}+2v)\le5\kappa^2md$.
The last inequality uses $v\le m$ and $d\ge1$. ReLU Lipschitz continuity implies
$\abth{\sigma(\bw^\top \bu)\sigma(\bw^\top \bv)-\sigma(\bw^\top \bu_0)\sigma(\bw^\top \bv_0)} \le \norm{\bw}{}^2(\norm{\bu-\bu_0}{}+\norm{\bv-\bv_0}{}).$
For net points within $m^{-1}$ of both inputs, the empirical and population extension errors sum to at most $2m^{-1}(\fnorm{\bW(0)}^2/m+\Expect{}{\norm{\bw}{}^2})\le12\kappa^2d/m\le20\kappa^2d/m$. This proves the joint bound. Continuity equates the supremum with that over a countable dense subset, proving measurability.
To identify the mean, put $\theta=\arccos(\bu^\top \bv)$. A two-dimensional Gaussian polar representation gives $A=\kappa R\cos t$, $B=\kappa R\cos(t-\theta)$, where $t$ is uniform and independent of $R$ with $\Expect{}{R^2}=2$. Integrating over the positive overlap yields
\bals
\Expect{}{X}\stackrel{\circled{2}}{=}\frac{\kappa^2}{\pi}\int_{\theta-\pi/2}^{\pi/2}\cos t\cos(t-\theta)\,dt
=\frac{\kappa^2}{2\pi}\set{(\pi-\theta)\cos\theta+\sin\theta}=K(\bu,\bv).
\eals
The endpoints follow by continuity.
\end{proof}
\begin{proof}[\textbf{\textup{Proof of Theorem~\ref{theorem:sup-hat-g}}}]
For arbitrary $\delta\in(0,1)$, use the scalar bound above on an $m^{-1}$-net with $q$ from (\ref{eq:C1}), giving failure at most $\delta/2$. The row-norm tail and a union bound give $\max_{r\in[m]}\norm{\bw_r}{}^2\le C\kappa^2(d+\log(2m/\delta))$ with failure at most $\delta/2$. The product Lipschitz bound, together with $\Expect{}{\norm{\bw}{}^2}=\kappa^2d$, adds at most $C\kappa^2(d+\log(2m/\delta))/m$. This proves (\ref{eq:sup-hat-h}) with the explicit formula (\ref{eq:C1}). All population expectations are untruncated.
\end{proof}

\begin{lemma}\label{lemma:uniform-bound-gaussian}
For every $\delta\in(0,1)$, with probability at least $1-\delta$ over $\bW(0)$,
\bals
\max_{r\in[m]}\ltwonorm{\bbw_r(0)}^2\le\kappa^2\set{d+2\sqrt{d\log(m/\delta)}+2\log(m/\delta)}.
\eals
\end{lemma}
\begin{proof}
Each $\ltwonorm{\bbw_r(0)}^2/\kappa^2$ is chi-square with $d$ degrees of freedom. Apply Lemma~\ref{lemma:chi-square-concentration} with $x=\log(m/\delta)$ and sum the $m$ tail probabilities.
\end{proof}

\begin{lemma}\label{lemma:chi-square-concentration}
{\rm (\cite[Lemma 1]{Laurent2000-chi-square-concentration})} Let $k\ge1$ be an integer and let $\set{X_i}_{i=1}^k$ be i.i.d. standard Gaussian random variables and put $X = \sum\limits_{i=1}^k X_i^2$. For every $x>0$,
\bals
\Prob{X - k \ge 2 \sqrt{kx} + 2x} \le \exp\pth{-x}, \quad \Prob{k - X \ge 2 \sqrt{kx} } \le \exp\pth{-x}.
\eals
\end{lemma}

\section{Further Details for the Kernel Approximation and Regression Risk}
\label{sec:calibration}

We give operator representations for the attention NTK, its empirical outer approximation and the empirical product induced by the network. The operators act on $\cH_K$, with operator norm $\ltwonorm{\cdot}$ and Hilbert--Schmidt norm $\norm{\cdot}{\textup{HS}}$, and we retain $b=\kappa^2/2$ and $s'=2s+3$.

\subsection{Kernel sections, covariance and spectral multipliers}
Write $k_{\bx}=K(\cdot,\bx)$. Continuity gives $\norm{k_{\bx}-k_{\by}}{\cH_K}^2=K(\bx,\bx)+K(\by,\by)-2K(\bx,\by)$, so the sections above a countable dense subset of compact $\cX$ have dense span. Thus $\cH_K$ is separable and all bounded operator expectations below are Bochner integrals. The operators $C,C_N$ are those defined in Section~\ref{sec:setup-kernels-target-function}.
Here $u\otimes v$ has Hilbert--Schmidt norm $\norm{u}{}\norm{v}{}$. Since $\norm{k_{\bx}}{}^2=b$, these covariances are positive trace-class operators with trace $b$ and operator norm at most $b$. Positivity is quadratic-form positivity and imposes no entrywise sign condition.
In the orthonormal RKHS basis $v_j=\sqrt{\lambda_j}e_j$, the reproducing identity and $L^2(\mu)$ orthogonality give $Cv_j=\lambda_jv_j$. The spectral power $C^{s-1}$ defined in Section~\ref{sec:setup-kernels-target-function} has norm at most $b^{s-1}\le1$. The uniform Mercer expansion in Theorem~\ref{theorem:Ks-bounded} gives $K^{(s)}(\bx,\bx')=\iprod{k_{\bx}}{C^{s-1}k_{\bx'}}$ on the entire support. Substitution into the defining integrals yields
\bal\label{eq:calibration-attention}
\begin{aligned}
\Kint(\bx,\bx')&=\iprod{k_{\bx}}{C^{s-1}C^2C^{s-1}k_{\bx'}},\\
\Kattn(\bx,\bx')&=\iprod{k_{\bx}}{CC^{s-1}C^2C^{s-1}Ck_{\bx'}}.
\end{aligned}
\eal
The population factors are powers of $C$, so their attention multiplier is $C^{2s+2}$, and its kernel integral operator has eigenvalues $\lambda_j^{2s+3}$, including the additional section-coordinate factor $\lambda_j$. Covariance operators act on $\cH_K$, while integral operators act on $L^2(\mu)$.

\subsection{Concentration without treating repeated samples as independent}

\begin{lemma}\label{lemma:calibration-covariance}
Recall that $b=\kappa^2/2$. For every $\delta\in(0,1)$, with probability at least $1-\delta$,
\bals
\ltwonorm{C_N-C}\le\norm{C_N-C}{\textup{HS}}
\le a_N(\delta),\qquad
a_N(\delta)=\frac{\sqrt{b^2-\sum_{j=1}^{\infty}\lambda_j^2}+b\sqrt{\log(1/\delta)}}{\sqrt N}.
\eals
\end{lemma}
\begin{proof}
For $Z_i=k_{\bbq_i}\otimes k_{\bbq_i}$, Hilbert--Schmidt centering and independence give
\bals
\Expect{}{\norm{C_N-C}{\textup{HS}}^2}=N^{-1}\Expect{}{\norm{Z_1-C}{\textup{HS}}^2} =N^{-1}(b^2-\norm{C}{\textup{HS}}^2)=N^{-1}\left(b^2-\sum_{j=1}^{\infty}\lambda_j^2\right).
\eals
Jensen bounds the mean norm by its square root. The exact rank-one identity is
$\norm{Z_{\bx}-Z_{\by}}{\textup{HS}}^2=2b^2-2K(\bx,\by)^2\le2b^2$.
Thus replacing one observation changes $\norm{C_N-C}{\textup{HS}}$ by at most $\sqrt2b/N$. Bounded differences bounds the upper deviation above its mean by $\exp\pth{-Nu^2/b^2}$ at threshold $u$. Take $u=b\sqrt{\log(1/\delta)/N}$ and use $\ltwonorm{C_N-C}\le\norm{C_N-C}{\textup{HS}}$.
\end{proof}

The covariance event controls the kernel arguments and Gram matrices uniformly. We apply it to the following product, whose averaging positions contain the same auxiliary observations.

The actual population-feature empirical attention kernel is
\bal
\tK_N(\bx,\bx')={}\frac1{N^4}\sum_{i,j,k,\ell=1}^N
K(\bx,\bbq_i)K^{(s)}(\bbq_i,\bbq_j)K(\bbq_j,\bbq_k)
K^{(s)}(\bbq_k,\bbq_\ell)K(\bbq_\ell,\bx').
\label{eq:calibration-five-factors}
\eal
Repeated indices and diagonal terms are included in (\ref{eq:calibration-five-factors}). Its exact section representation is
\bals
\tK_N(\bx,\bx')=\iprod{k_{\bx}}{C_NC^{s-1}C_N^2C^{s-1}C_Nk_{\bx'}}.
\eals
Replacing its four empirical covariance factors by $C$ gives the population representation (\ref{eq:calibration-attention}).
\begin{lemma}\label{lemma:calibration-product}
Recall that $b=\kappa^2/2$. On the event of Lemma~\ref{lemma:calibration-covariance},
\bals
\sup_{\bx,\bx'\in\cX}\abth{\tK_N(\bx,\bx')-\Kattn(\bx,\bx')}
\le4b^4a_N(\delta).
\eals
\end{lemma}
\begin{proof}
Replace the four occurrences of $C_N$ in $M_N=C_NC^{s-1}C_NC_NC^{s-1}C_N$ one at a time by $C$. The four resulting ordered products each contain one factor $C_N-C$, three covariances of norm at most $b$, and two deterministic factors of norm at most one. No empirical factors are commuted. Thus, for $M=CC^{s-1}C^2C^{s-1}C$,
\bals
\ltwonorm{M_N-M}\stackrel{\circled{1}}{\le}4b^3\ltwonorm{C_N-C}\le4b^3a_N(\delta),\qquad
\abth{\iprod{k_{\bx}}{(M_N-M)k_{\bx'}}}\le 4b^4a_N(\delta).
\eals
The first bound uses submultiplicativity and the second uses $\norm{k_{\bx}}{}^2=b$.
\end{proof}

The outer approximation has multiplier $C_NC^{s-1}C^2C^{s-1}C_N$, while the actual product uses $C_N$ in its middle positions. Their difference is $C_NC^{s-1}(C_N^2-C^2)C^{s-1}C_N$, whose norm is at most $2b^3a_N(\delta)$. The initialization approximation therefore uses the empirical product with all four covariance factors.

\subsection{Initialization derivatives and zero output weights}

The zero output initialization explains why the kernel in Lemma~\ref{lemma:empirical-loss-convergence-contraction-bKattn-approx} is the full initialization NTK. Since $\bA$ is fixed and symmetric, the chosen ReLU subgradient gives
\bals
\nabla_{\ba}f(\bW,\ba,\bx)=m^{-1/2}\bA\bsigma(\bW,\bx),\qquad
\nabla_{\bbw_r}f(\bW,\ba,\bx)=m^{-1/2}(\bA\ba)_r\indict{\bbw_r^\top \bx\ge0}\bx.
\eals
At $\ba(0)=0$, the first-layer derivative vanishes. The inner product of the second-layer derivatives therefore gives
$J_m(\bx,\bx')=m^{-1}\bsigma(\bW(0),\bx)^\top\bA^2\bsigma(\bW(0),\bx').$
By symmetry of $\bA$ and the zero first-layer derivative, Lemmas~\ref{lemma:empirical-loss-convergence-contraction-bKattn-approx} and~\ref{lemma:calibration-product} give $\sup_{\bx,\bx'}\abth{J_m(\bx,\bx')-\Kattn(\bx,\bx')}\le3C_1+4b^4a_N(\delta)$. This also bounds the normalized Gram error, with an additional factor $n$ for the unnormalized Gram matrix.

\subsection{Concentration of positive quadratic forms at the fixed noise scale}

The noise assumption in the setup implies that the vector $\bw$ satisfies
$\Expect{}{\exp(\iprod{ \bv}{\bw})}\le \exp(\sigma_0^2\ltwonorm{\bv}^2/2)$ for every $\bv\in\RR ^n$.
Indeed independence permits factorization of the moment-generating function, and the scalar assumption bounds each factor. This argument is valid conditional on the training design and the sample $\bQ$ because the noise is independent of both. Accordingly, the conditional assertion below applies to matrices determined by the covariates and independent auxiliary randomness.

\begin{lemma}\label{lemma:calibration-quadratic}
Let $\bM$ be a deterministic positive semidefinite $n\times n$ matrix and let $x>0$. Then
\bals
\Pr\set{\bw^\top \bM\bw>
\sigma_0^2\bigl(\tr{\bM}+
2\sqrt{\tr{\bM^2}x}+2\ltwonorm{\bM}x\bigr)}\le \exp\pth{-x}.
\eals
The same assertion holds conditionally when $\bM$ depends only on the covariates and independent auxiliary randomness.
\end{lemma}
\begin{proof}

\smallskip
\noindent Let $\bg$ be standard Gaussian in $\RR ^n$, independent of $\bw$. For $\theta>0$, the Gaussian exponential identity gives
$\exp\pth{\theta\bw^\top \bM\bw} =\Expect{\bg}{\exp\pth{\sqrt{2\theta}\iprod{ \bg}{\bM^{1/2}\bw}}}.$
Tonelli's theorem applies because the integrand is nonnegative. Taking expectation in $\bw$ and using the vector sub-Gaussian bound yields
\bals
\Expect{}{\exp\pth{\theta\bw^\top \bM\bw}}
\stackrel{\circled{1}}{\le}\Expect{\bg}{\exp\pth{\theta\sigma_0^2\bg^\top \bM\bg}}
=\prod_{i=1}^n(1-2\theta\sigma_0^2m_i)^{-1/2},
\eals
Here \circled{1} uses the vector sub-Gaussian moment-generating bound, and the equality uses diagonalization of the Gaussian quadratic form.

Here $m_i$ are the eigenvalues of $\bM$, and $2\theta\sigma_0^2\ltwonorm{\bM}<1$.

\smallskip
\noindent To control the logarithm, we use the elementary power-series inequality
$-\log(1-u)-u\le u^2/[2(1-u)]$ for $0\le u<1$.

Applying this inequality to the eigenvalues gives
\bals
\log\Expect{}{\exp\set{\theta(\bw^\top \bM\bw-\sigma_0^2\tr{\bM})}}
\stackrel{\circled{2}}{\le}\frac{\theta^2\sigma_0^4\tr{\bM^2}}
{1-2\theta\sigma_0^2\ltwonorm{\bM}}.
\eals
\smallskip
\noindent Write $v=\sigma_0^4\tr{\bM^2}$ and $c=2\sigma_0^2\ltwonorm{\bM}$. If $v=0$, positivity implies $\bM=0$ and the conclusion is immediate. Otherwise take $\theta=\sqrt{x/v}/(1+c\sqrt{x/v})$. Substitution in the exponential Markov bound shows that a deviation $2\sqrt{vx}+cx$ has probability at most $\exp\pth{-x}$. Conditioning on a covariate-measurable matrix preserves every step.
\end{proof}

The expectation is bounded by the proxy $\sigma_0^2\tr{\bM}$, and the argument uses the joint moment-generating bound for the original noise. For the prediction and RKHS forms, take $x=c_qn/(\eta T)$ with $c_q=1/[64(1+\sigma_0^2)]$. Since $2\sigma_0\sqrt{c_q}+2\sigma_0^2c_q<1$, the bounds are $2/(\eta T)$ and $2$, respectively, simultaneously for $t\le T$ by filter monotonicity.

\subsection{Interpolation Spaces and Population Norm Comparison}
The interpolation norm is $\norm{f}{[\cH_K]^r}^2=\sum_{j=1}^{\infty}\beta_j^2/\lambda_j^r$ for $f=\sum_{j=1}^{\infty}\beta_je_j$. The spectral identity $\lambdaattn_j=\lambda_j^{s'}$ gives equality of Hilbert balls $[\cH_K]^{s'}(\mu_0)=\cH_{\Kattn}(\mu_0)$. For $r\ge1$, Mercer tails provide canonical continuous representatives.
For $r_2\ge r_1>0$ and $0<\lambda_j\le b<1$,
$\norm{f}{[\cH_K]^{r_1}}^2\le b^{r_2-r_1}\norm{f}{[\cH_K]^{r_2}}^2,$ and $\abth{h(\bx)}\le \norm{h}{\cH_{\Kattn}}\sqrt{\Kattn(\bx,\bx)}.$
The first bound is coefficientwise spectral comparison. The second is reproduction, with $\Kattn(\bx,\bx)\le b^{s'}\le b$. Thus the deterministic RKHS ball has uniformly bounded clean squared loss even with unbounded response noise. The quadratic residual comparison is proved in Theorem~\ref{theorem:LRC-population-NN-eigenvalue}.
\subsection{Critical Rates and the Early Stopping Time}
Trace summability and dominated convergence make $R_{\Kattn}(\eps)^2=n^{-1}\sum_{j=1}^{\infty}\min\set{\lambdaattn_j,\eps^2}$ continuous. For $\eps>0$, the square of $R_{\Kattn}(\eps)/\eps^2$ has summands $\min\set{\lambdaattn_j/\eps^4,1/\eps^2}$, strictly decreasing when the spectrum is nonzero. The ratio tends to infinity at zero and to zero at infinity. Hence the critical-radius equation has exactly one positive solution, as does its empirical counterpart. The empirical trace is at most $b^{s'}$, so $\hat\eps_n^4\le\sigma_0^2b^{s'}/n$, and the stopping set is nonempty for every design when $n\ge\eta^2\sigma_0^2b^{s'}$.
The stopping inequality $\hat R_{\Kattn}((\eta t)^{-1/2})\le(\sigma_0\eta t)^{-1}$ is equivalent by this ratio monotonicity to $\eta t\hat\eps_n^2\le1$. Therefore $\hat T=\lfloor1/(\eta\hat\eps_n^2)\rfloor$, and for $T=\hat T\ge1$ the integer inequalities give $(\eta T)^{-1}\asymp\hat\eps_n^2$.
Ordered Gram eigenvalues are continuous in the covariates, so $\set{\hat T\ge k}$ is Borel for each integer $k$. The positive empirical radius makes $\hat T$ finite. Kernel evaluations, ReLU and the chosen zero subgradient are Borel, so each explicit GD update is measurable. The stopped function $f_{\hat T}=\sum_{k\ge0}\indict{\hat T=k}f_k$ is jointly measurable. Continuous sample paths and Tonelli make its regression risk measurable.
The choice $T=\hat T$ ensures that the signal bias reaches the critical-radius scale. More generally, this requires a lower bound $T\ge c\hat T$ within the stopping interval. Proposition~\ref{prop:infinite-spectrum-radius} gives $n\eps_n^2\to\infty$ under the non-atomic law, and polynomial spectra give $n\eps_n^2\asymp n^{1/(2\alpha s'+1)}$.

\smallskip
\noindent\textbf{A baseline with exactly the assumed kernel evaluations.}
Ordinary early-stopped kernel GD~\cite{RaskuttiWY14-early-stopping-kernel-regression} attains the same regression risk rate with the same kernel evaluations as the network. The proof below verifies its probability, regression risk and measurability.
\begin{proposition}\label{prop:kernel-baseline}
Recall that $b=\kappa^2/2$. Under the covariate, target-function and sub-Gaussian noise assumptions of the general main theorem, define $g_0=0$ and, for integers $k\ge0$,
\bals
g_{k+1}(\bx)=g_k(\bx)-\frac\eta n\sum_{i=1}^n\Kattn(\bx,\bbx_i)\set{g_k(\bbx_i)-y_i}.
\eals
Use the same $\eta\in(0,1)$ and $T=\hat T$ as in that theorem. For every fixed $c_t\in(0,1)$ and sufficiently large $n$, with probability at least $1-C\exp\pth{-cn\eps_n^2}$, simultaneously for $\lceil c_tT\rceil\le t\le T$, $\Expect{P}{(g_t-f^*)^2}\le C\eps_n^2$. If $\lambda_j\asymp j^{-2\alpha}$ with $\alpha>1/2$, the same event gives regression risk at most $Cn^{-2\alpha s'/(2\alpha s'+1)}$ throughout this interval. All constants are uniform over the stated target ball. This estimator uses only the labeled observations and the kernel evaluations already supplied to the network.
\end{proposition}
\begin{proof}
Fix the design and put $A=\bKattn_n$, $S_k=(I-\eta A)^k$ and $r_k=g_k(\bS)-\by$. Evaluation of the update gives $r_{k+1}=(I-\eta A)r_k$ and $r_k=-S_kf^*(\bS)-S_k\bw$. The function update is
$g_t=-\frac\eta n\sum_{k=0}^{t-1}\sum_{i=1}^n\Kattn(\cdot,\bbx_i)r_{k,i}.$
Thus Theorem~\ref{theorem:empirical-loss-bound} applies with zero recursion error. The telescoped function update is the general sum of Lemma~\ref{lemma:bounded-Linfty-vt-sum-et} up to sign. Choose $\tau=\min\set{1,1/(\eta T)}$ so its hypotheses hold. Intersecting these conditional noise events yields failure at most $C\exp\pth{-cn\hat\eps_n^2}$ and, for all $t\le T$,
$\norm{g_t}{\cH_{\Kattn}}\le B_h$, and $P_n(g_t-f^*)^2\le \frac3{\eta t}\left(\frac{\mu_0^2}{2e}+2\right)$.
The clean-loss calculation has no accumulated error here. The deterministic-ball inequality proved for Theorem~\ref{theorem:LRC-population-NN-eigenvalue}, with $x=n\eps_n^2$, gives $P(g_t-f^*)^2\le2P_n(g_t-f^*)^2+C\eps_n^2$ on a design event. Intersect also Lemma~\ref{lemma:hat-eps-eps-relation}. Then $\hat\eps_n^2\asymp\eps_n^2$ and $T\asymp\eps_n^{-2}$, so $1/(\eta t)\le C\eps_n^2$ throughout the stated interval. Integrating the conditional noise failure over this design event and adding its complement proves the simultaneous population bound with failure $C\exp\pth{-cn\eps_n^2}$.
Each fixed-time update is jointly Borel, and the finite design-measurable stopping time preserves measurability by the countable-time construction given above. Under $\lambda_j\asymp j^{-2\alpha}$ with $\alpha>1/2$, the polynomial critical-radius calculation gives $\eps_n^2\asymp n^{-2\alpha s'/(2\alpha s'+1)}$. Substitution proves the polynomial regression risk bound on the same event. Every constant is uniform over the target ball.
\end{proof}

\section{Proofs for the Unbounded Input Space}
\label{sec:proofs-unbounded}
\begingroup
We normalize the Euclidean construction, verify the bounded-input GD hypotheses and restore the response scale. Constants may depend on $M_2$ and fixed parameters, but not on sample sizes or failure probabilities.

\subsection{Normalization and the target space}
\begin{lemma}\label{lemma:unbounded-normalization}
Under the assumptions of Section~\ref{sec:setup-unbounded}, the kernels in (\ref{eq:unbounded-powered-kernels}) are well-defined and $T_{\Kattn}=T_K^{s'}$. The normalized attention kernel in (\ref{eq:normalized-attention-kernel}) extends continuously to $\operatorname{supp}(\tmu)\subseteq\unitsphere d$, with diagonal bounded by $D_0$. The map $f\mapsto f/q$ is an isometry of the associated RKHSs, with the argument changed by $\phi$. Their restrictions to the respective supports have infinite positive spectra, and their sampled Gram matrices are strictly positive definite almost surely.
\end{lemma}
\begin{proof}
Continuity makes $\cH_K$ separable. With $I:\cH_K\to L^2(\mu)$ the inclusion, \newline
$\norm{Ih}{\mu}^2\le bM_2\norm{h}{\cH_K}^2$ and $\norm{K_{\bx}\otimes K_{\bx}}{\rm tr}=bq(\bx)^2$ make $C$ positive trace class, with $\tr{C}=bM_2$. The reproducing identity gives $C=I^*I$ and $T_K=II^*$. Singular-value decomposition then gives, also for real $s\ge1$,
\bals
T_{\Kattn}=IC^{2s+2}I^*\stackrel{\circled{1}}{=}(II^*)^{2s+3},\qquad T_{K^{(s)}}=T_K^s.
\eals
The bounds $\abth{K(\bx,\bx')}\le bq(\bx)q(\bx')$ and $\abth{K^{(s)}(\bx,\bx')}\le b\lambda_1^{s-1}q(\bx)q(\bx')$ give one $q^2$ factor for each integrated variable. The absolute five-factor integral is at most $b^5\lambda_1^{2s-2}q(\bx)q(\bx')M_2^4<\infty$, justifying Fubini and the source coefficient description.

Represent $K_0$ by $h_{\bz}(\bbw)=\sigma(\bbw^\top \bz)$ in $L^2(\cN(0,\kappa^2\bI_{d+1}))$. ReLU Lipschitz continuity gives $\ltwonorm{h_{\bz}-h_{\bz'}}\le\kappa\ltwonorm{\bz-\bz'}$ and $\ltwonorm{h_{\bz}}^2=b$ on the sphere. The Euclidean section has feature vector $q(\bx)h_{\phi(\bx)}$. Transport $C$ to their closed span and extend it by zero on its orthogonal complement. The normalized kernel has features $C^{s+1}h_{\bz}$, continuous on the whole sphere, and
$\barKattn(\bz,\bz)=\ltwonorm{C^{s+1}h_{\bz}}^2
\le \lambda_1^{2s+2}b=D_0$.
This canonically extends the kernel to $\operatorname{supp}(\tmu)$. To verify the target norm, map each finite linear combination $\sum_{i=1}^k c_i\Kattn(\cdot,\bx_i)$ to $\sum_{i=1}^k c_iq(\bx_i)\barKattn(\cdot,\phi(\bx_i))$, where $k\ge1$, $c_i\in\RR$ and $\bx_i\in\operatorname{supp}(\mu)$. The defining relation (\ref{eq:normalized-attention-kernel}) gives
\bals
\norm{\sum_{i=1}^k c_iq(\bx_i)\barKattn(\cdot,\phi(\bx_i))}{\cH_{\barKattn}}^2
=\sum_{i,j=1}^kc_ic_j\Kattn(\bx_i,\bx_j)
=\norm{\sum_{i=1}^kc_i\Kattn(\cdot,\bx_i)}{\cH_{\Kattn}}^2.
\eals
Thus the map is well-defined even when a function has several finite-span representations, and extends by completion as an isometry. Its range is dense because every normalized section at an image point is the image of an original section divided by the positive scalar $q(\bx)$. Continuity of normalized sections in the RKHS norm adds sections at boundary points of $\operatorname{supp}(\tmu)$, so the extension is onto that RKHS. Reproduction identifies the image of any $f$ at $\phi(\bx)$ with $f(\bx)/q(\bx)$. Applying this to $f^*$ proves the source bound for $g^*$. Its continuous representative also specifies values at boundary points that have no Euclidean preimage. Thus normalization of the completed attention kernel preserves the stated source class.

Injectivity and the absence of antipodal image pairs let Proposition~\ref{proposition:Kattn-gram-nonsingular} give strict Gram positivity, preserved for $K$ by positive $q$ factors. On the closed span of support sections, $\iprod{h}{Ch}=0$ implies $h=0$ almost surely, hence on the support by continuity, hence in that span. $C^{s+1}$ therefore preserves finite linear independence, giving strict Gram positivity for $\Kattn,\barKattn$. Non-atomicity supplies arbitrarily many distinct sampled points, whose strict Gram positivity contradicts finite operator rank as in Proposition~\ref{prop:infinite-spectrum-radius}. Both spectra are infinite.
\end{proof}

\subsection{Covariance and initialization bounds}
\begin{lemma}\label{lemma:unbounded-covariance}
Use $C_N$ from Section~\ref{sec:setup-kernels-target-function} and define $M_{2,N}=N^{-1}\sum_{i=1}^Nq(\bbq_i)^2$. For each $\delta\in(0,1)$, with probability at least $1-\delta$ over $\bQ$,
\bal\label{eq:unbounded-covariance-bound}
\ltwonorm{C_N-C}\le\norm{C_N-C}{\textup{HS}}\le C a_{N,\Phi}(\delta),\qquad
\abth{M_{2,N}-M_2}\le C a_{N,\Phi}(\delta).
\eal
For a fixed sufficiently small $c>0$, define $N_0(P,n,\delta)$ to be the least positive integer such that, for every deterministic $N\ge N_0(P,n,\delta)$, with probability at least $1-\delta$ over $\bQ$,
\bal\label{eq:unbounded-auxiliary-accuracy}
\norm{C_N-C}{\textup{HS}}+\abth{M_{2,N}-M_2}\le c\eps_{0,n}^3.
\eal
This threshold is finite for every integer $n\ge1$ and $\delta\in(0,1)$. The probability statement is for each chosen $N$, without simultaneous control over all sample sizes. If $\Phi(R)\le2\exp\pth{-aR^2}$ for a fixed $a>0$, then, with probability at least $1-\delta$,
\bal\label{eq:unbounded-subgaussian-covariance}
\norm{C_N-C}{\textup{HS}}+\abth{M_{2,N}-M_2}
\le C\pth{\sqrt{\frac{\log(2/\delta)}N}+\frac{\log(2/\delta)}N}.
\eal
Consequently, for $\eps_{0,n}\le1$, $N_0(P,n,\delta)\lsim\eps_{0,n}^{-6}\log(2/\delta)$. The constant may depend on $a$ and the fixed model parameters, but not on $n,N,\delta$.
\end{lemma}
\begin{proof}
Fix a deterministic $R\ge1$, put $Z(\bx)=K_{\bx}\otimes K_{\bx}$, $Z_R(\bx)=Z(\bx)\indict{\ltwonorm{\bx}\le R}$ and $H=1+R^2$. Since $\norm{Z_R}{\textup{HS}}\le bH$, independence removes centered cross terms and gives
\bals
\Expect{}{\norm{N^{-1}\sum_{i=1}^NZ_R(\bbq_i)-\Expect{}{Z_R(\bX)}}{\textup{HS}}^2} =N^{-1}\Expect{}{\norm{Z_R-\Expect{}{Z_R}}{\textup{HS}}^2} \le b^2H^2/N.
\eals
The change per observation is at most $2bH/N$. Combining the mean bound with bounded differences~\cite{Wainwright2019-high-dim-statistics} gives error at most $CbH\sqrt{\log(8/\delta)/N}$ with failure $\delta/4$. Positivity and Markov give the empirical tail bound, with failure $\delta/4$,
\bals
\norm{N^{-1}\sum_{i=1}^N(Z-Z_R)(\bbq_i)}{\textup{HS}}
\le \frac bN\sum_{i=1}^Nq(\bbq_i)^2\indict{\ltwonorm{\bbq_i}>R}
\le 4b\Psi(R)/\delta.
\eals
The bias is at most $b\Psi(R)$. For the scalar mean, apply bounded differences to $q(\bX)^2\indict{\ltwonorm{\bX}\le R}\in[0,H]$, with failure probability $\delta/4$. The full mean difference is at most this truncated-mean error plus the empirical tail and $\Psi(R)$. The same Markov event above bounds the empirical tail by $4\Psi(R)/\delta$. Thus the scalar estimate uses no additional tail event and no fourth moment. Choosing a deterministic radius within twice the positive infimum in (\ref{eq:unbounded-tail-definitions}) proves (\ref{eq:unbounded-covariance-bound}), without a union over radii or lost independence from reusing $\bQ$.

We first justify the threshold in (\ref{eq:unbounded-auxiliary-accuracy}) under a finite second moment. For each fixed $n,\delta$, choose a deterministic $R\ge1$ so that $\Psi(R)\le c'\delta\eps_{0,n}^3$, where $c'>0$ is sufficiently small relative to $c$ and the constants in (\ref{eq:unbounded-covariance-bound}). Such an $R$ exists because $\Psi(R)\to0$. Taking $N\gsim(1+R^2)^2\eps_{0,n}^{-6}\log(8/\delta)$ with a sufficiently large constant makes the sum of the two errors in (\ref{eq:unbounded-covariance-bound}) at most $c\eps_{0,n}^3$. The same choice works for every larger deterministic $N$. Hence a finite least threshold exists. The resulting threshold depends on the law through its integrable weighted tail.

For the sub-Gaussian improvement, we concentrate the full observations directly. Continuity of $K$ on the separable domain $\RR^d$ makes $\cH_K$ separable and its kernel sections continuous in RKHS norm. The space of Hilbert--Schmidt operators on $\cH_K$ is therefore a separable Hilbert space. In its Hilbert direct sum with $\RR$, put $\bZ(\bx)=(K_{\bx}\otimes K_{\bx},q(\bx)^2)$. Its norm is $\sqrt{b^2+1}\,q(\bx)^2$, and its expectation is $(C,M_2)$. The tail assumption implies, by integration,
\bals
\Expect{P}{\exp\pth{a\ltwonorm{\bX}^2/2}}\le3,\qquad
\Expect{P}{q(\bX)^{2k}}\le3\exp\pth{a/2}\,k!(2/a)^k,\quad k\ge2.
\eals
Here $k$ is an integer. Centering and Jensen's inequality therefore give a fixed $L>0$, depending only on $a,b$, such that the independent centered observations $\bY_i=\bZ(\bbq_i)-\Expect{P}{\bZ(\bX)}$ satisfy $\Expect{}{\norm{\bY_i}{}^k}\le k!L^k$ for every integer $k\ge2$.

We apply a Hilbert-space Bernstein argument to obtain the confidence dependence. For $0<\theta<L^{-1}$, define $F(\bu)=\cosh(\theta\norm{\bu}{})$ on this direct-sum Hilbert space. The second Fr\'echet derivative $D^2F$, with its continuous extension at zero, satisfies $\abth{D^2F(\bu)[\bv,\bv]}\le\theta^2F(\bu)\norm{\bv}{}^2$ and $F(\bu+t\bv)\le F(\bu)\exp\pth{\theta t\norm{\bv}{}}$ for $0\le t\le1$. Taylor's formula with integral remainder, centering and the moment bound yield
\bals
\Expect{}{F(\bu+\bY_i)}
&\le F(\bu)\pth{1+\sum_{k=2}^{\infty}\frac{\theta^k\Expect{}{\norm{\bY_i}{}^k}}{k!}}
\le F(\bu)\exp\pth{\frac{\theta^2L^2}{1-\theta L}}.
\eals
The first inequality follows by integrating the Hessian bound along $\bu+t\bY_i$, expanding the nonnegative exponential remainder and using $\Expect{}{\bY_i}=0$. All terms are integrable because $\theta L<1$. Conditional application and independence then give $\Expect{}{F(\sum_{i=1}^N\bY_i)}\le\exp\pth{N\theta^2L^2/(1-\theta L)}$. Since $\cosh u\ge\exp(u)/2$ for $u\ge0$, exponential Markov and the choice $\theta=r/(2NL^2+Lr)$ give, for every $r>0$,
\bals
\Prob{\norm{\sum_{i=1}^N\bY_i}{}\ge r}
\le2\exp\pth{-\frac{r^2}{4NL^2+2Lr}}.
\eals
Taking $r=2L\sqrt{N\log(2/\delta)}+2L\log(2/\delta)$ gives failure at most $\delta$. Cauchy--Schwarz bounds the sum of the two component norms by $\sqrt2$ times their joint norm, proving (\ref{eq:unbounded-subgaussian-covariance}). For $\eps_{0,n}\le1$, a sufficiently large constant in $N\gsim\eps_{0,n}^{-6}\log(2/\delta)$ makes the joint error at most $c\eps_{0,n}^3$ and gives the asserted bound for $N_0$.
\end{proof}

\begin{lemma}\label{lemma:normalized-initialization}
Suppose $m\ge2$, $m\gsim n^{1/(2(d+1))}$ and $m/\log m\ge d+1$. Put $C_1=C_1(m,d+1,1/n)$. Fix $\delta\in(0,1)$ and a deterministic $\eps\in(0,1]$. Suppose $N$ is chosen so that $\norm{C_N-C}{\textup{HS}}+\abth{M_{2,N}-M_2}\le\eps$ holds with probability at least $1-\delta$ over $\bQ$. There is an initialization event of probability at least $1-2/n$ on which the uniform feature covariance error on $\unitsphere d$ is at most $C_1\lesssim\sqrt{(d+1)\log m/m}$ and $\max_{r\in[m]}\ltwonorm{\bbw_r(0)}\le C\sqrt{d+1+\log(2mn)}$. On its intersection with this covariance event, if $C_1\le1$, the matrix in (\ref{eq:normalized-attention-network}) satisfies, for a fixed constant $L\ge1$,
\bal\label{eq:normalized-matrix-bounds}
\ltwonorm{\bA}\le L,\qquad
\max_{r\in[m]}\ltwonorm{\bA_r}\le\frac{C(d+1+\log(2mn))}{\sqrt m}.
\eal
Furthermore, the initialization kernel $J_m(\bz,\bz')=m^{-1}\bsigma(\bW(0),\bz)^\top\bA^2\bsigma(\bW(0),\bz')$ obeys
\bal\label{eq:normalized-initialization-kernel}
\sup_{\bz,\bz'\in\operatorname{supp}(\tmu)}\abth{J_m(\bz,\bz')-\barKattn(\bz,\bz')}
\le C\set{C_1+\eps}.
\eal
\end{lemma}
\begin{proof}
Theorem~\ref{theorem:good-random-initialization} and Lemma~\ref{lemma:uniform-bound-gaussian}, in dimension $d+1$, give both initialization bounds on the full sphere with total failure $2/n$. Let $\bH=\bG_{\bQ}\bD$. The covariance event, diagonal bounds and $\sigma(\bbw^\top\phi(\bx))\le\ltwonorm{\bbw}$ give
$\frac{\ltwonorm{\bH}^2}{Nm}\le (b+C_1)M_{2,N}$, $\ltwonorm{\bK_N^{(s)}}\le b\lambda_1^{s-1}M_{2,N}$, and $\ltwonorm{\bH_r}\le \ltwonorm{\bbw_r(0)}\sqrt{NM_{2,N}}$.
The stated covariance event gives $M_{2,N}\le M_2+1$. Substitution into $\bA=\bH\bK_N^{(s)}\bH^\top/(Nm)$ proves (\ref{eq:normalized-matrix-bounds}) and fixes $L$ deterministically.

We expand $\bA^2$ and replace its three empirical feature covariances by their expectations. Retaining the same auxiliary sample in the four averaging positions gives
\bals
J_N(\phi(\bx),\phi(\bx'))\stackrel{\circled{1}}{=}
\iprod{\frac{K_{\bx}}{q(\bx)}}{C_NC^{s-1}C_N^2C^{s-1}C_N\frac{K_{\bx'}}{q(\bx')}}_{\cH_K}.
\eals
As in Lemma~\ref{lemma:calibration-product}, the end norms are $\sqrt b$, the four covariance norms at most $bM_{2,N}$ and the two power norms at most $\lambda_1^{s-1}$. Telescoping gives
$\sup_{\bx,\bx'\in\operatorname{supp}(\mu)}\abth{J_N(\phi(\bx),\phi(\bx'))-\barKattn(\phi(\bx),\phi(\bx'))} \le C\ltwonorm{C_N-C}.$
On the covariance event, telescoping the three feature positions uses errors $C_1$, remaining normalized factors bounded by $b+C_1$, and $\abth{K^{(s)}(\bx,\bx')}\le b\lambda_1^{s-1}q(\bx)q(\bx')$. Each summand therefore has weight $\prod_{\ell=1}^4q(\bbq_{i_\ell})^2$, including repeated indices. It follows that
$\sup_{\bz,\bz'\in\operatorname{supp}(\tmu)}\abth{J_m(\bz,\bz')-J_N(\bz,\bz')} \le CC_1\left(N^{-1}\sum_{i=1}^Nq(\bbq_i)^2\right)^4 \le CC_1.$
Together with the assumed covariance accuracy, these bounds prove (\ref{eq:normalized-initialization-kernel}). Continuity covers boundary points of the image support.
\end{proof}

\subsection{The bounded-input training argument after normalization}
\begin{lemma}\label{lemma:normalized-training}
Fix $\delta\in(0,1)$, the training design and $1\le T\le\hat T_0$. Let $w\in(0,1)$ and fix a deterministic $\eps\in(0,1]$. Suppose the deterministic auxiliary size $N$ is chosen so that $\norm{C_N-C}{\textup{HS}}+\abth{M_{2,N}-M_2}\le\eps$ holds with probability at least $1-\delta$ over $\bQ$, and suppose
\bal\label{eq:normalized-training-sizes}
m\gsim\max\set{(d+1)\log m\,T^3,(d+1)\log m\,T^2/w^2,n^{1/(2(d+1))}},\qquad
\eps\le c\min\set{T^{-3/2},w/T}.
\eal
Choose a fixed $\eta\in(0,\eta_0)$, where $\eta_0=\min\set{1,(2D_0)^{-1},[2L^2(b+2)]^{-1}}$ and $L$ is as in Lemma~\ref{lemma:normalized-initialization}. With conditional probability at least $1-2/n-\delta-C\exp\pth{-cn}-C\exp\pth{-cn\hat\eps_{0,n}^2}$, the iterates in (\ref{eq:normalized-attention-loss}) admit, simultaneously for $t\le T$, the representation
\bal\label{eq:normalized-training-decomposition}
f_t^{(\textup{nor})}(\bx)=h_t(\phi(\bx))+e_t(\bx),\qquad
\norm{h_t}{\cH_{\barKattn}}\le B_h,\qquad \sup_{\bx\in\operatorname{supp}(\mu)}\abth{e_t(\bx)}\le w.
\eal
On the same event, for every $1\le t\le T$,
\bal\label{eq:normalized-clean-empirical-error}
\frac1n\sum_{i=1}^n\left(f_t^{(\textup{nor})}(\bbx_i)-g^*(\phi(\bbx_i))\right)^2\le\frac C{\eta t}.
\eal
\end{lemma}
\begin{proof}
Put $\bz_i=\phi(\bbx_i)$, $\xi_i=w_i/q(\bbx_i)$, $\bu(t)=(f_t^{(\textup{nor})}(\bbx_i)-y_i/q(\bbx_i))_{i=1}^n$, $\bfsym{A}=(\barKattn(\bz_i,\bz_j))_{i,j\in[n]}/n$, $\bS_j=(\bI_n-\eta \bfsym{A})^j$, $\bF_j=\bsigma(\bW(j),(\bz_i)_{i=1}^n)$ and $\bG=\bF_0$. Write $\rho=C\set{C_1+\eps}$, $D_m=d+1+\log(2mn)$ and $\Gamma \bc=\sum_{i=1}^nc_i\barKattn(\cdot,\bz_i)$, so $\Gamma^*\Gamma=n\bfsym{A}$.

The bound $\ltwonorm{\bfsym{A}}\le D_0$ and the stipulated $\eta_0$ give $\ltwonorm{\bS_j}\le1$. Lemma~\ref{lemma:unbounded-normalization} gives $\bfsym{A}^{-1}$. Conditional independence of the original noise gives the heterogeneous joint bound
\bals
\Expect{}{\exp\pth{\bv^\top\bxi}\mid\bS}\stackrel{\circled{1}}{\le}
\exp\left(\frac{\sigma_0^2}{2}\sum_{i=1}^n\frac{v_i^2}{q(\bbx_i)^2}\right)
\le \exp\pth{\sigma_0^2\ltwonorm{\bv}^2/2}.
\eals
Lemma~\ref{lemma:calibration-quadratic} applies conditionally. For $q_j(z)=1-(1-\eta z)^j$, $q_j(z)^2/z\le\eta j\min\set{1,\eta jz}$, so normalized stopping bounds the noise trace by $1$ and operator norm by $\eta T/n$. Equation (\ref{eq:early-stopped-signal-noise}), with $\Gamma,g^*,\bxi$, gives norms $\mu_0,\sqrt2$ simultaneously, with failure $C\exp\pth{-cn\hat\eps_{0,n}^2}$. The residual-noise estimate has failure probability $C\exp\pth{-cn}$. In this normalized proof, choose the fixed residual constant
\bals
c_{\bu}=\mu_0\max\set{1,\sqrt{D_0},(2e\eta)^{-1/2}}+\sigma_0+2.
\eals
Reproduction bounds the time-zero signal by $\mu_0\sqrt{nD_0}$. For $t\ge1$, the empirical RKHS projection and $z(1-\eta z)^{2t}\le(2e\eta t)^{-1}$ give $\mu_0\sqrt{n/(2e\eta t)}$. On the residual-noise event, $\ltwonorm{\bS_t\bxi}\le(\sigma_0+1)\sqrt n$. Adding an accumulated error of norm at most $\sqrt nT^{-1/2}\le\sqrt n$ gives $\ltwonorm{\bu(t)}\le c_{\bu}\sqrt n$, also when $D_0>1$.

The filtered-error identity (\ref{eq:early-stopped-filtered-error}) remains exact with $\Gamma,\bfsym{A},\bS_j$. If preceding errors satisfy $\ltwonorm{\bE(i)}\le\sqrt n/(4T^{3/2})$, spectral calculus gives $\norm{n^{-1}\Gamma \bfsym{A}^{-1}(\bI_n-\bS_\ell)}{}\le\sqrt{\eta\ell/n}$. Consequently,
\bals
\norm{\frac\eta n\Gamma\sum_{j=0}^{t-1}\sum_{i=1}^j\bS_{j-i}\bE(i)}{}
\le \frac{\sqrt\eta}{4T^{3/2}}\sum_{i=1}^{t-1}\sqrt{t-i}\le \frac14.
\eals
Hence $h_t=-\eta n^{-1}\Gamma\sum_{j=0}^{t-1}\bu(j)$ has norm at most $B_h$, using errors strictly before $t$.

The changed matrix bounds needed in the update induction are $\ltwonorm{\bA}\le L$, $\max_{r\in[m]}\ltwonorm{\bA_r}\le CD_m/\sqrt m$, $\ltwonorm{\bG}\le C\sqrt{mn}$ and $\ltwonorm{\bG^\top\bA^2\bG/(nm)-\bfsym{A}}\le\rho$. Suppose the residual and output-weight bounds hold through $k<T$ and choose fixed large $C_a$. The actual updates, using $\ltwonorm{\bz_i}=1$, give
\bals
\max_{r\in[m]}\ltwonorm{\bbw_r(k+1)-\bbw_r(k)}\le CC_aD_mT^{1/4}/m,\qquad
\max_{r\in[m]}\ltwonorm{\bbw_r(k+1)-\bbw_r(0)}\le CC_aD_mT^{5/4}/m,\\
\ltwonorm{\frac\eta{n\sqrt m}\bA \bG\sum_{j=0}^k\bu(j)}^2\le B_h^2+CT^2\rho,\qquad
\ltwonorm{\frac\eta{n\sqrt m}\bA\sum_{j=0}^k(\bF_j-\bG)\bu(j)}\le CC_aD_mT^{9/4}/m.
\eals
The first pair uses the row bound and summed increments. The second uses $\Gamma^*\Gamma=n\bfsym{A}$ and ReLU Lipschitz continuity, which bounds $\ltwonorm{\bF_j-\bG}$ by $\sqrt{mn}$ times the largest row movement. Here the constants in the four bounds are fixed once $L,D_0,\eta$ and the residual bound have been fixed. Choose $C_a>2(B_h+1)$ before choosing the size constants. The covariance and first width conditions in (\ref{eq:normalized-training-sizes}) can make $\rho\le c_*T^{-3/2}$ and $D_mT^{9/4}/m\le c_*T^{-3/4}$ for an arbitrarily small fixed $c_*>0$, since $D_m\lesssim(d+1)\log m$. The frozen output contribution is then at most $(B_h+C\sqrt{c_*})T^{1/4}$ and the moving contribution is at most $CC_ac_*T^{1/4}$. Taking $C\sqrt{c_*}\le1$ and $Cc_*\le1/2$ makes their sum at most $C_aT^{1/4}$. This proves the output-weight bound at $k+1$ from the induction hypotheses through $k$, including the bound for $h_{k+1}$. We can therefore estimate the next residual using the new output-weight bound.

The simultaneous-update identity bounds the feature-increment term by $CC_a^2D_mT^{1/2}\sqrt n/m$ and the outer-feature approximation by $CC_aD_mT^{5/4}/m$. Thus
$\ltwonorm{\bE(k+1)}/\sqrt n\le C(D_mT^{5/4}/m+\rho)
\le (4T^{3/2})^{-1}$.
The last step uses $T^{11/4}\le T^3$. Contraction gives accumulated error $\sqrt n/(4\sqrt T)$, closing the induction from zero with the modified $L,D_0,\eta$, including sign changes.

Freezing both feature occurrences in the telescoped output update, as in Theorem~\ref{theorem:bounded-NN-class}, now gives
\bal\label{eq:normalized-pointwise-comparison}
\sup_{\bx\in\operatorname{supp}(\mu)}\abth{f_t^{(\textup{nor})}(\bx)-h_t(\phi(\bx))}
\le C\set{D_mT^{9/4}/m+TC_1+T\eps}
\le w.
\eal
Movement and (\ref{eq:normalized-initialization-kernel}) give the feature and kernel approximation errors. The geometric mean of the width conditions gives $m\gsim(d+1)\log m\,T^{5/2}/w$. The other terms use (\ref{eq:normalized-training-sizes}), proving (\ref{eq:normalized-training-decomposition}) via (\ref{eq:normalized-pointwise-comparison}).

Theorem~\ref{theorem:empirical-loss-bound}'s calculation with the verified filter, source, joint noise and accumulated error gives contributions $C/(\eta t)$, $C/(\eta T)$ and $1/(16T)$ respectively. The prediction-noise estimate has failure probability $C\exp\pth{-cn\hat\eps_{0,n}^2}$ and proves (\ref{eq:normalized-clean-empirical-error}). Filter monotonicity makes the noise events simultaneous. Intersecting with initialization and covariance gives the stated probability.
\end{proof}

\subsection{Population regression risk and the tail transfer}
\begin{proof}[\textbf{\textup{Proof of Theorem~\ref{theorem:regression-unbounded}}}]
\textbf{Proof outline.} Verify normalized stopping and training, transfer empirical error to population error, then restore the response scale.
The compact normalized support and diagonal bound $D_0$ permit the proof of Lemma~\ref{lemma:hat-eps-eps-relation}: deleting one observation changes the covariance-resolvent trace by at most $D_0/(n\eps_{0,n}^2)$, its mean is comparable to the population effective dimension, and concentration followed by fixed-point bracketing gives $\hat\eps_{0,n}^2\asymp\eps_{0,n}^2$ with failure $C\exp\pth{-cn\eps_{0,n}^2}$. The diagonal bound gives $\hat\eps_{0,n}^2\le\sigma_0\sqrt{D_0/n}$. Thus $\eta\sigma_0\sqrt{D_0/n_0}\le1$ ensures $\hat T_0\ge1$ for every training design when $n\ge n_0$. Infinite positive spectrum and the radius comparison then give $T=\hat T_0\asymp\eps_{0,n}^{-2}$ on the design event.

Choose $n_0\ge2$ sufficiently large for this radius comparison, $\eta\sigma_0\sqrt{D_0/n_0}\le1$ and, for every $n\ge n_0$, $\eps_{0,n}<1$, $c\eps_{0,n}^3\le1$. These requirements depend only on the fixed law and model parameters. Take $w=\eps_{0,n}$ and use the deterministic covariance tolerance $\eps=c\eps_{0,n}^3$ in Lemma~\ref{lemma:normalized-training}. The choice $N\ge N_0(P,n,\delta)$ gives (\ref{eq:unbounded-auxiliary-accuracy}) with failure at most $\delta$, independently of the training design. On the radius event, both width terms in (\ref{eq:normalized-training-sizes}) are at most $C(d+1)\log m\,\eps_{0,n}^{-6}$, and both $T^{-3/2}$ and $w/T$ are comparable to $\eps_{0,n}^3$. Taking the fixed $c$ sufficiently small therefore verifies the training conditions. Integrating Lemma~\ref{lemma:normalized-training}'s conditional bound is valid on every such design. There $\exp\pth{-cn\hat\eps_{0,n}^2}\le \exp\pth{-c'n\eps_{0,n}^2}$.

The verified bounds permit the local-complexity argument of Theorem~\ref{theorem:LRC-population-NN-eigenvalue} for the kernel $\barKattn$ under $\tmu$, with diagonal bound $D_0$. Young's inequality gives, uniformly over the deterministic class,
\bals
\Expect{P}{(f_t^{(\textup{nor})}-f^*/q)^2}
\le 2\Expect{P_n}{(f_t^{(\textup{nor})}-f^*/q)^2}+C(\eps_{0,n}^2+w^2)
\le C\eps_{0,n}^2,\\
\abth{f_t^{(\textup{nor})}(\bx)-f^*(\bx)/q(\bx)}
\le (B_h+\mu_0)\sqrt{\barKattn(\phi(\bx),\phi(\bx))}+w\le B.
\eals
The additional failure is $C\exp\pth{-cn\eps_{0,n}^2}$. The second step uses the empirical clean-error bound and $t\ge c_tT$. The third is the reproducing inequality. This proves (\ref{eq:normalized-regression-risk}) simultaneously on the stated interval.

Write $e=f_t^{(\textup{nor})}-f^*/q$. Restoring scale and splitting the domain gives, on the same event and for every $R\ge0$,
\bals
\Expect{P}{(f_t-f^*)^2}
&=\Expect{P}{q(\bX)^2e(\bX)^2\indict{\ltwonorm{\bX}\le R}}+\Expect{P}{q(\bX)^2e(\bX)^2\indict{\ltwonorm{\bX}>R}}\\*
&\stackrel{\circled{1}}{\le}(1+R^2)\Expect{P}{e(\bX)^2}+B^2\Psi(R).
\eals
The last step uses the bounded-radius weight and the uniform error envelope. Substitution proves (\ref{eq:unbounded-regression-tail}) for all radii on the same event. Adding the radius, training and population failures gives the stated probability.
\end{proof}

\begin{proof}[\textbf{\textup{Proof of Corollary~\ref{corollary:regression-unbounded-subgaussian}}}]
\textbf{Proof outline.} Apply the sub-Gaussian tail to the auxiliary requirement and risk transfer, then substitute the spectral upper bound.
The direct concentration estimate (\ref{eq:unbounded-subgaussian-covariance}) and a sufficiently large constant in (\ref{eq:unbounded-subgaussian-N}) give (\ref{eq:unbounded-auxiliary-accuracy}) for every larger deterministic $N$. Hence the finite least threshold satisfies $N_0(P,n,\delta)\lsim\eps_{0,n}^{-6}\log(2/\delta)$, and Theorem~\ref{theorem:regression-unbounded} applies. The tail identity in (\ref{eq:unbounded-tail-definitions}) gives $\Psi(R)\le2(1+R^2+a^{-1})\exp\pth{-aR^2}$. For $n\ge2$, take $R^2=3a^{-1}\log n$. On the simultaneous event of the theorem, this gives
$\Expect{P}{(f_t-f^*)^2}\le C(\log n)\eps_{0,n}^2+C(\log n)n^{-3}
\le C'(\log n)\eps_{0,n}^2$.
The second step uses $n\eps_{0,n}^2\to\infty$. The tail implies $M_2<\infty$, so Lemma~\ref{lemma:euclidean-base-eigenvalue-upper} and Proposition~\ref{prop:normalized-eigenvalue-upper}, with $\alpha=(d+3)/(2d)$, give
$\eps_{0,n}^2\le Cn^{-2\alpha s'/(2\alpha s'+1)}
=Cn^{-s'(d+3)/(s'(d+3)+d)}$.
This deterministic substitution proves (\ref{eq:unbounded-subgaussian-risk}) without an additional failure probability.
\end{proof}

\subsection{Eigenvalue upper bounds under a finite second moment}
A spherical harmonic approximation bounds the Euclidean spectrum without a density assumption. Operator comparison then handles normalization, which does not commute with powers of $K_0$.

\begin{lemma}\label{lemma:euclidean-base-eigenvalue-upper}
Let $\mu$ be a Borel probability measure on $\RR^d$ with $M_2=\int_{\RR^d}(1+\ltwonorm{\bx}^2)\,d\mu(\bx)<\infty$. For $K$ in (\ref{eq:kernel-unbounded}), let $(\lambda_j)_{j\ge1}$ be the eigenvalues of $T_K$ in nonincreasing order, counted with multiplicity and completed by zeros if necessary. Then
\bal\label{eq:euclidean-base-eigenvalue-upper}
\lambda_j\le C_d\kappa^2 M_2\,j^{-\frac{d+3}{d}},\qquad j\ge1,
\eal
where $C_d$ depends only on $d$. In particular, $\lambda_j\lesssim j^{-2\alpha}$ with $\alpha=(d+3)/(2d)>1/2$. No density or additional covariate tail assumption is required.
\end{lemma}
\begin{proof}
Under normalized surface measure $\sigma_d$, let $(Y_{\ell,k})_{k=1}^{h_\ell}$ be an orthonormal basis of real degree-$\ell$ harmonics and $\theta_\ell$ the corresponding eigenvalue of $T_{K_0}$. The ReLU calculation~\cite[Theorem~1, Section~3.2 and Appendix~A]{BiettiB21}, in ambient dimension $d+1$, gives $0\le\theta_\ell\le C_d\kappa^2(1+\ell)^{-(d+3)}$, $h_\ell\le C_d(1+\ell)^{d-1}$ and $\sum_{k=1}^{h_\ell}Y_{\ell,k}(\bz)^2=h_\ell$. Hence
$K_0(\bz,\bz')=\sum_{\ell=0}^\infty\theta_\ell\sum_{k=1}^{h_\ell}Y_{\ell,k}(\bz)Y_{\ell,k}(\bz').$
Cauchy--Schwarz gives $\sum_{k=1}^{h_\ell}\abth{Y_{\ell,k}(\bz)Y_{\ell,k}(\bz')}\le h_\ell$, so summability of $\theta_\ell h_\ell$ proves absolute uniform convergence. Odd-degree coefficients above degree one vanish. The covariance kernel's first nonpolynomial endpoint term has exponent $3/2$, giving degree decay $\ell^{-(d+3)}$ in ambient dimension $d+1$. Multiplicity has order $\ell^{d-1}$, so the diagonal-tail summand is $\ell^{-4}$. The ordered index $j$ in (\ref{eq:euclidean-base-eigenvalue-upper}) counts these multiplicities.

Let $K_{0,L}$ retain degrees $0$ through an integer $L\ge1$. Nonnegative coefficients make both the retained and omitted kernels positive semidefinite, while the addition formula gives
$(K_0-K_{0,L})(\bz,\bz)=\sum_{\ell=L+1}^\infty\theta_\ell h_\ell \le C_d\kappa^2\sum_{\ell=L+1}^\infty(1+\ell)^{-4} \le C_d\kappa^2L^{-3}.$
Define $K_L(\bx,\bx')=q(\bx)q(\bx')K_{0,L}(\phi(\bx),\phi(\bx'))$, $A_L=T_{K_L}$ and $B_L=T_{K-K_L}$. Bounded harmonics and $M_2<\infty$ make every retained feature $q(\bx)Y_{\ell,k}(\phi(\bx))$ square integrable. Thus $T_K=A_L+B_L$ with positive trace-class summands and $\tr{T_K}=bM_2$. Tonelli and the uniform omitted diagonal yield
\bals
\operatorname{rank}(A_L)\le r_L=\sum_{\ell=0}^Lh_\ell\le C_d(L+1)^d,\qquad
\tr{B_L}=\int q(\bx)^2(K_0-K_{0,L})(\phi(\bx),\phi(\bx))\,d\mu(\bx)
\le C_d\kappa^2M_2L^{-3}.
\eals
If $L^2(\mu)$ is finite dimensional, indices beyond its dimension have zero eigenvalues and satisfy the bound. For the remaining large $j$, choose $L=\lfloor c_dj^{1/d}\rfloor$ with fixed small $c_d>0$, so $L\ge1$, $r_L\le j/2$ and $L\asymp j^{1/d}$. To retain the extra factor $j^{-1}$, combine the rank bound with the trace of the omitted operator. For \circled{1} below, restrict to the orthogonal complement of $\operatorname{ran}(A_L)$ and the first $j-r_L-1$ eigenvectors of $B_L$. This subspace has codimension at most $j-1$, the $A_L$ quadratic form vanishes on it, and the $B_L$ Rayleigh quotient is at most $\lambda_{j-r_L}(B_L)$. The min--max principle gives the first inequality. Positivity gives the next one because the first $j-r_L$ eigenvalues of $B_L$ are each at least $\lambda_{j-r_L}(B_L)$. Thus
\bals
\lambda_j(T_K)&\stackrel{\circled{1}}{\le}\lambda_{j-r_L}(B_L)\le \frac{\tr{B_L}}{j-r_L} \le C_d\kappa^2M_2j^{-1}j^{-3/d}=C_d\kappa^2M_2j^{-(d+3)/d}.
\eals
For the finitely many remaining indices, $\lambda_j\le\tr{T_K}=bM_2$ gives the same bound after increasing $C_d$. This proves (\ref{eq:euclidean-base-eigenvalue-upper}) without a density or additional tail hypothesis.
\end{proof}

\begin{proposition}\label{prop:normalized-eigenvalue-upper}
Let $(\nu_j)_{j\ge1}$ be the positive eigenvalues of $T_{\barKattn}$ under $\tmu$, in nonincreasing order. Then $\nu_j\le\lambda_j^{s'}$. If $\lambda_j\lesssim j^{-2\alpha}$ with $\alpha>1/2$, then $\eps_{0,n}^2\lesssim n^{-2\alpha s'/(2\alpha s'+1)}$.
\end{proposition}
\begin{proof}
Changing variables by $\phi$ is unitary between the two $L^2$ spaces. In $L^2(\mu)$ let $M$ multiply by $q^{-1}$, a self-adjoint contraction. The normalized operator becomes $MT_K^{s'}M$. Its nonzero spectrum equals that of $T_K^{s'/2}M^2T_K^{s'/2}$, using $U^*U$ and $UU^*$ for $U=T_K^{s'/2}M$. Since $0\le M^2\le I$, min-max principal gives
$\nu_j \le\lambda_j(T_K^{s'})=\lambda_j^{s'}$.

Under the polynomial upper bound, split the complexity sum at $J=\lceil\eps^{-1/(\alpha s')}\rceil$. The head is at most $J\eps^2$ and the tail at most $CJ^{1-2\alpha s'}$, hence
$R_{\barKattn}(\eps)^2\le Cn^{-1}\eps^{2-1/(\alpha s')}$.
Taking $\eps^2=C'n^{-2\alpha s'/(2\alpha s'+1)}$ with large $C'$ satisfies the critical-radius inequality. Combining this estimate with (\ref{eq:unbounded-regression-tail}) proves (\ref{eq:unbounded-polynomial-tail-risk}). Since the spectral bound is one-sided, the training conditions remain expressed in terms of the normalized critical radius.
\end{proof}

\section{Proof of the Biased Vanilla Comparison}
\label{sec:proof-vanilla-bias}

We verify (\ref{eq:vanilla-li-regression-risk}) through kernel statistics and all-time coupling, adapting the concentration and filtering argument of~\cite[Proposition~13]{Li2024-edr-general-domain} to the specified architecture.

\begin{proposition}\label{prop:vanilla-bias-regression}
Let $P$ be non-atomic on $\RR^d$ and satisfy $\Phi(R)\le2\exp\pth{-aR^2}$ for all $R\ge0$, with fixed $a>0$. Retain the noise and independence assumptions of Section~\ref{sec:setup}, and suppose $f^*\in[\cH_K]^{s'}(\mu_0)$ with $s'=2s+3\ge5$. Consider (\ref{eq:vanilla-bias-network}), with $\ba(0)=0$ and independent $\cN(0,\kappa^2\bI_{d+1})$ first-layer rows, trained by the gradient flow specified in the comparison paragraph. Let $\lambda_0=\lambda_{\min}((K(\bbx_i,\bbx_j))_{i,j\in[n]})$ and choose deterministic $t_{\rm op}\asymp n^{(d+3)/(s'(d+3)+d)}$. There is a sufficiently large fixed constant $C_{\rm v}>0$, depending only on the covariate law and fixed model parameters, with the following property. For every fixed $\delta\in(0,1)$, every deterministic positive integer $m$ and sufficiently large $n$, an event of probability at least $1-\delta$ over $\bS,\bw,\bW(0)$ exists on which the width condition
\bal\label{eq:vanilla-comparison-width}
m\ge C_{\rm v}\bigl(1+n+\log(1/\delta)+\lambda_0^{-1}\bigr)^{32}
\eal
implies
\bals
\Expect{P}{(f_{t_{\rm op}}^{\textup{(vanilla,bias)}}-f^*)^2}
\le C\log^2(12/\delta)\,n^{-\frac{s'(d+3)}{s'(d+3)+d}}.
\eals
The constants are uniform over the stated source ball and independent of $n,m,\delta$. The lower threshold on $n$ is independent of $\lambda_0$. The displayed polynomial is a deliberately coarse sufficient width condition. The guarantee holds for every gradient-flow trajectory, so any universally measurable trajectory selection has the stated regression risk.
\end{proposition}

\subsection{Statistical bound for the biased vanilla kernel flow}
\label{sec:vanilla-biased-kernel-statistics}

The covariance and filtered-noise bounds use factorial moments to handle the unbounded diagonal $bq(\bX)^2$.
\begin{lemma}\label{lemma:vanilla-moment-bernstein}
Let $\bZ_1,\ldots,\bZ_n$ be independent, centered random vectors in a real separable Hilbert space with norm $\norm{\cdot}{}$. Suppose that $B>0$ and $v_i^2\ge0$ satisfy
\bals
\Expect{}{\norm{\bZ_i}{}^p}\le\frac{p!}{2}v_i^2B^{p-2},\qquad p=2,3,\ldots,\quad i\in[n].
\eals
Put $V=\sum_{i=1}^n v_i^2$. For every $u>0$, with probability at least $1-2\exp\pth{-u}$,
\bal\label{eq:vanilla-moment-bernstein}
\norm{\sum_{i=1}^n\bZ_i}{}\le\sqrt{2Vu}+Bu.
\eal
\end{lemma}
\begin{proof}
For $0<\theta<B^{-1}$, put $F(\bz)=\cosh(\theta\norm{\bz}{})$. Its radial second derivative is $\theta^2F(\bz)$ and its tangential coefficient is $\theta\sinh(\theta\norm{\bz}{})/\norm{\bz}{}\le\theta^2F(\bz)$, also at zero. The triangle inequality gives $F(\bz+t\bh)\le F(\bz)\exp\pth{t\theta\norm{\bh}{}}$. Taylor's integral remainder therefore yields
\bals
F(\bz+\bh)\stackrel{\circled{1}}{\le}F(\bz)+DF(\bz)[\bh]+F(\bz)(\exp\pth{\theta\norm{\bh}{}}-1-\theta\norm{\bh}{}).
\eals
Centering removes the linear term. Tonelli and the moment assumption sum the remaining series. Successive conditioning then gives
\bals
\Expect{}{F(\bz+\bZ_i)}\le F(\bz)\exp\left(\frac{\theta^2v_i^2}{2(1-\theta B)}\right),\qquad
\Expect{}{\cosh\left(\theta\norm{\sum_{i=1}^n\bZ_i}{}\right)}
\stackrel{\circled{2}}{\le}\exp\left(\frac{\theta^2V}{2(1-\theta B)}\right).
\eals
If $V\allowbreak=0$ every vector vanishes almost surely. Otherwise Markov and $\cosh(a)\allowbreak\ge \exp\pth{a}/2$ give
$\Pr\left(\norm{\sum_{i\allowbreak=1}^n\bZ_i}{}>a\right)
\allowbreak\le 2\exp\left(-\theta a\allowbreak+\frac{\theta^2V}{2(1-\theta B)}\right)$.
Substituting $\theta\allowbreak=\sqrt{2u/V}/(1\allowbreak+B\sqrt{2u/V})$ and $a\allowbreak=\sqrt{2Vu}\allowbreak+Bu$ makes the exponent $-u$, proving (\ref{eq:vanilla-moment-bernstein}).
\end{proof}

\begin{lemma}\label{lemma:vanilla-source-power-perturbation}
Let $A$ and $B$ be positive compact operators on a separable Hilbert space, with $\ltwonorm{A},\ltwonorm{B}\le M$, where $M>0$. If $A-B$ is Hilbert--Schmidt and $r\ge1$, then
\bal\label{eq:vanilla-source-power-perturbation}
\norm{A^r-B^r}{\textup{HS}}\le rM^{r-1}\norm{A-B}{\textup{HS}}.
\eal
\end{lemma}
\begin{proof}
Let $\cJ$ index a finite or countable complete orthonormal basis. Choose orthonormal eigenbases $(u_i)_{i\in\cJ}$ and $(v_j)_{j\in\cJ}$ for $A,B$, including their null spaces, with eigenvalues $a_i,b_j$. For $D_r(a,b)=(a^r-b^r)/(a-b)$ and $D_r(a,a)=ra^{r-1}$, the mean-value theorem gives $\abth{D_r(a,b)}\le rM^{r-1}$. When $r=1$ its value at zero is $1$. Evaluating the left and right eigenvectors gives
$\iprod{u_i}{(A^r-B^r)v_j}=D_r(a_i,b_j)\iprod{u_i}{(A-B)v_j}$.
When $a_i=b_j$, both sides vanish. Parseval, including the finite or countable basis sums, gives
\bals
\norm{A^r-B^r}{\textup{HS}}^2\stackrel{\circled{1}}{=}\sum_{i\in\cJ}\sum_{j\in\cJ}\abth{D_r(a_i,b_j)}^2\abth{\iprod{u_i}{(A-B)v_j}}^2
\le r^2M^{2r-2}\norm{A-B}{\textup{HS}}^2.
\eals
The bounds on finite partial sums and monotone convergence establish the Hilbert--Schmidt property and (\ref{eq:vanilla-source-power-perturbation}) for every real $r\ge1$.
\end{proof}

\begin{proposition}\label{prop:vanilla-biased-kernel-statistics}
Suppose that the covariates and noise satisfy the independence and regression assumptions in Section~\ref{sec:setup}, that $\Phi(R)\le2\exp\pth{-aR^2}$ for every $R\ge0$ and a fixed $a>0$, and that $f^*\in[\cH_K]^{s'}(\mu_0)$, with $s'=2s+3\ge5$. Define the training-sample covariance operator
$C_n=\frac1n\sum_{i=1}^n K_{\bbx_i}\otimes K_{\bbx_i}$.
For $t\ge0$, let $g_t(z)=(1-\exp\pth{-tz})/z$ for $z>0$ and $g_t(0)=t$, and let the zero-start empirical kernel gradient flow be
\bal\label{eq:vanilla-biased-kernel-flow}
\hat f_t^K=g_t(C_n)\frac1n\sum_{i=1}^ny_iK_{\bbx_i}.
\eal
Choose a deterministic stopping time
\bal\label{eq:vanilla-biased-stopping-time}
t_{\mathrm{op}}\asymp n^{\frac{d+3}{s'(d+3)+d}},
\eal
where the two comparison constants are fixed. For every fixed $\delta\in(0,1)$ and all sufficiently large $n$, with probability at least $1-\delta$ over the labeled observations,
\bal\label{eq:vanilla-biased-kernel-risk}
\norm{\hat f_{t_{\mathrm{op}}}^K-f^*}{\mu}^2
\le C\log^2(12/\delta)\,n^{-\frac{s'(d+3)}{s'(d+3)+d}}.
\eal
The constant $C$ is independent of $n$ and $\delta$, and is uniform over the stated target ball. No eigenvalue lower bound is assumed.
\end{proposition}
\begin{proof}
Set $\gamma=d/(d+3)$, $r=(s'-1)/2=s+1$, $t=t_{\mathrm{op}}$, $\lambda=t^{-1}$ and $L=\log(12/\delta)$. Thus (\ref{eq:vanilla-biased-stopping-time}) gives $\lambda\asymp n^{-1/(s'+\gamma)}$. Take $n$ large enough that $\lambda\le1$. Scalars added to operators multiply the identity on $\cH_K$.

Put $Q=q(\bX)^2$ and $A_q=4(1+a^{-1})$. Tail integration and $(1+u)^p\le2^{p-1}(1+u^p)$ give, for integers $p\ge1$,
$\Expect{}{\ltwonorm{\bX}^{2p}}=p\int_0^\infty u^{p-1}\Pr(\ltwonorm{\bX}^2>u)\,du
\le 2p!a^{-p}$, and $\Expect{}{Q^p}\le p!A_q^p$.
Lemma~\ref{lemma:unbounded-normalization} gives separability, $\tr{C}=b\Expect{}{Q}$ and $\norm{h}{\mu}=\norm{C^{1/2}h}{\cH_K}$. The source expansion $f^*=\sum_{j=1}^\infty c_j\lambda_j^{s'/2}e_j$ supplies $u=\sum_{j=1}^\infty c_jv_j$, $v_j=\sqrt{\lambda_j}e_j$, with $\norm{u}{\cH_K}\le\mu_0$ and canonical representative $f^*=C^ru$. Equality holds in $L^2(\mu)$ and at the observations almost surely. Reproduction and spectral calculus give
\bal\label{eq:vanilla-source-envelope}
\abth{f^*(\bx)}\le \norm{C^ru}{\cH_K}\sqrt{K(\bx,\bx)}
\le \mu_0\ltwonorm{C}^r\sqrt b\,q(\bx).
\eal
Equation (\ref{eq:vanilla-source-envelope}) supplies the label moments.

For $Z=K_\bX\otimes K_\bX-C$, the rank-one norm and Jensen give
$\Expect{}{\norm{Z}{\textup{HS}}^p}\le 2^{p-1}(b^p\Expect{}{Q^p}+\norm{C}{\textup{HS}}^p) \le 2^pb^p\Expect{}{Q^p} \le p!(2bA_q)^p.$
Lemma~\ref{lemma:vanilla-moment-bernstein}, with $B=2bA_q$ and $v_i^2=2B^2$, gives a design event $E_C$, of failure $2\exp\pth{-L}$, on which
\bal\label{eq:vanilla-covariance-HS}
\ltwonorm{C_n-C}\le \norm{C_n-C}{\textup{HS}}
\le C(\sqrt{L/n}+L/n)
\le CL/\sqrt n.
\eal
Since $\lambda\sqrt n\asymp n^{1/2-1/(s'+\gamma)}\to\infty$, require (\ref{eq:vanilla-covariance-HS}) at most $\lambda/2$. Ordering and inversion give
\bal\label{eq:vanilla-covariance-comparison}
C\le C_n+\lambda,\qquad
\ltwonorm{C^{1/2}(C_n+\lambda)^{-1/2}}\le 1,\qquad
(C_n+\lambda)^{-1}\stackrel{\circled{1}}{\le}\tfrac32(C+\lambda)^{-1}.
\eal
The last follows from $C+\lambda\le C_n+3\lambda/2\le(3/2)(C_n+\lambda)$. Also $\ltwonorm{C_n}\le M=\ltwonorm{C}+1$, a fixed bound for the real-power lemma.

Define $\cN(\lambda)=\tr{C(C+\lambda)^{-1}}$, $\cN_n(\lambda)=\tr{C_n(C_n+\lambda)^{-1}}$ and $\ell_\lambda(\bx)=\iprod{K_{\bx}}{(C+\lambda)^{-1}K_{\bx}}$. Lemma~\ref{lemma:euclidean-base-eigenvalue-upper} bounds the nonzero covariance eigenvalues by $Aj^{-1/\gamma}$. Splitting at $J=\lceil\lambda^{-\gamma}\rceil$ gives
\bal\label{eq:vanilla-population-effective-dimension}
\cN(\lambda)=\sum_{j=1}^\infty\frac{\lambda_j}{\lambda_j+\lambda}
\le J+\frac A\lambda\sum_{j=J+1}^\infty j^{-1/\gamma}
\le C\lambda^{-\gamma}.
\eal
Tonelli gives $\Expect{}{\ell_\lambda(\bX)}\allowbreak=\cN(\lambda)$, and $0\allowbreak\le\ell_\lambda(\bX)\allowbreak\le bQ/\lambda$. Centering and Jensen therefore give
$\Expect{}{\abth{\ell_\lambda(\bX)-\cN(\lambda)}^p}\allowbreak\le 2^p\Expect{}{\ell_\lambda(\bX)^p}
\allowbreak\le p!(2bA_q/\lambda)^p$.
The scalar moment lemma, with $B\allowbreak=2bA_q/\lambda$, $v_i^2\allowbreak=2B^2$, yields $E_\ell$, of failure $2\exp\pth{-L}$, satisfying
\bal\label{eq:vanilla-leverage-trace-concentration}
\frac1n\sum_{i=1}^n\ell_\lambda(\bbx_i)
\le \cN(\lambda)+\frac{CL}{\lambda\sqrt n}
\le C\lambda^{-\gamma}.
\eal
For the last step use (\ref{eq:vanilla-population-effective-dimension}) and require $L\allowbreak\le c\sqrt n\lambda^{1-\gamma}$, which holds eventually since $n^{1/2-(1-\gamma)/(s'\allowbreak+\gamma)}\to\infty$. Sandwiched ordering in (\ref{eq:vanilla-covariance-comparison}) and the rank-one trace give, on $E_C\cap E_\ell$,
\bal\label{eq:vanilla-empirical-effective-dimension}
\cN_n(\lambda)\stackrel{\circled{2}}{\le}\tfrac32\tr{C_n(C+\lambda)^{-1}}
=\frac{3}{2n}\sum_{i=1}^n\ell_\lambda(\bbx_i)
\le C\lambda^{-\gamma}.
\eal
Neither event needs an eigenvalue lower bound.

Let $\xi=n^{-1}\sum_{i=1}^nw_iK_{\bbx_i}$. The clean empirical response equals $C_nf^*$ and $g_t(C_n)C_n=I-\exp\pth{-tC_n}$, including its null space. Hence (\ref{eq:vanilla-biased-kernel-flow}) gives
\bal\label{eq:vanilla-biased-flow-decomposition}
\hat f_t^K-f^*=-\exp\pth{-tC_n}C^ru+g_t(C_n)\xi.
\eal
For $B_t=\norm{C^{1/2}\exp\pth{-tC_n}C^ru}{\cH_K}$, norm comparison, insertion of $C_n^r$ and spectral calculus give
\bals
B_t&\le \norm{(C_n+\lambda)^{1/2}\exp\pth{-tC_n}C^ru}{\cH_K}\\
&\stackrel{\circled{3}}{\le}\mu_0\sup_{z\ge0}(z+\lambda)^{1/2}z^r\exp\pth{-tz}
+\mu_0\sup_{z\ge0}(z+\lambda)^{1/2}\exp\pth{-tz}\ltwonorm{C^r-C_n^r}\\
&\le C\lambda^{r+1/2}+C\sqrt\lambda\,\norm{C_n-C}{\textup{HS}}.
\eals
Substitute $z=\lambda v$. The two scalar suprema then have orders $\lambda^{r+1/2}$ and $\sqrt\lambda$, because $(1+v)^{1/2}v^r\exp\pth{-v}$ and $(1+v)^{1/2}\exp\pth{-v}$ are bounded on $[0,\infty)$. Lemma~\ref{lemma:vanilla-source-power-perturbation} applies with $r=s+1\ge2$ and the fixed bound $M$ on both covariances. It controls $C^r-C_n^r$ in Hilbert--Schmidt norm and hence in operator norm. Inserting $C_n^r$ separates the empirical spectral filter from the source-operator perturbation and permits the estimate for noncommuting covariances. Since $r+1/2=s'/2$, (\ref{eq:vanilla-covariance-HS}) gives for real $s$,
\bal\label{eq:vanilla-biased-flow-bias}
B_t\le C\lambda^{s'/2}+CL\sqrt{\lambda/n}\qquad\text{on }E_C.
\eal
Condition on the design and put $a_i=n^{-1}(C_n+\lambda)^{1/2}g_t(C_n)K_{\bbx_i}$, $S_t=\sum_{i=1}^n\norm{a_i}{\cH_K}^2$ and $a_{\max}=\max_i\norm{a_i}{\cH_K}$. From $1-\exp\pth{-v}\le\min\set{v,1}$, separately for $z\le\lambda$ and $z\ge\lambda$, including $g_t(0)=\lambda^{-1}$,
$(z+\lambda)\abth{g_t(z)}\le 2$, and $(z+\lambda)g_t(z)^2z\le \frac{4z}{z+\lambda}$.
Summing the empirical rank-one traces and applying (\ref{eq:vanilla-empirical-effective-dimension}) yields
\bal\label{eq:vanilla-filtered-noise-trace}
S_t\stackrel{\circled{4}}{=}\frac1n\tr{(C_n+\lambda)g_t(C_n)^2C_n}
\le \frac4n\cN_n(\lambda)
\le \frac{C}{n\lambda^\gamma}.
\eal
The original $w_i$ remain independent and centered conditionally. Integrating their sub-Gaussian tails gives $\Expect{}{\abth{w_i}^p\mid\bS}\allowbreak\le p!A_w^p$ for $p\allowbreak\ge2$, with $A_w\allowbreak=4\sigma_0$. The estimate follows from $\exp\pth{-v^2/(2\sigma_0^2)}\allowbreak\le \exp\pth{1/2}\exp\pth{-v/\sigma_0}$. Therefore
$\Expect{}{\norm{w_ia_i}{\cH_K}^p\mid\bS} \allowbreak\le \frac{p!}{2}(2A_w^2\norm{a_i}{\cH_K}^2)(A_wa_{\max})^{p-2}.$
If $a_{\max}\allowbreak=0$ the noise vanishes. Otherwise the conditional Hilbert-space moment lemma, with $V\allowbreak=2A_w^2S_t$ and $B\allowbreak=A_wa_{\max}$, gives an event $E_w$ of conditional failure $2\exp\pth{-L}$ on which
$\norm{\sum_{i\allowbreak=1}^nw_ia_i}{\cH_K} \allowbreak\le 2A_w\sqrt{S_tL}\allowbreak+A_wa_{\max}L \allowbreak\le 3A_wL\sqrt{S_t}.$
The last step uses $a_{\max}\allowbreak\le\sqrt{S_t}$ and $L\allowbreak\ge1$. Since the sum equals $(C_n\allowbreak+\lambda)^{1/2}g_t(C_n)\xi$, the norm comparison and (\ref{eq:vanilla-filtered-noise-trace}) give
\bal\label{eq:vanilla-biased-flow-noise}
\norm{C^{1/2}g_t(C_n)\xi}{\cH_K}
\le \norm{\sum_{i=1}^nw_ia_i}{\cH_K}
\le \frac{CL}{\sqrt{n\lambda^\gamma}}.
\eal
Only the original noises are assumed independent.

Combining (\ref{eq:vanilla-biased-flow-decomposition}), (\ref{eq:vanilla-biased-flow-bias}) and (\ref{eq:vanilla-biased-flow-noise}) gives
$\norm{\hat f_t^K-f^*}{\mu}\le C\lambda^{s'/2}+CL\sqrt{\lambda/n}+\frac{CL}{\sqrt{n\lambda^\gamma}} \le CLn^{-s'/(2(s'+\gamma))}.$
Here $\lambda^{s'/2}\asymp(n\lambda^\gamma)^{-1/2}$ and $\sqrt{\lambda/n}\le(n\lambda^\gamma)^{-1/2}$. Squaring proves the $\log^2(12/\delta)$ risk in (\ref{eq:vanilla-biased-kernel-risk}). The two design failures and the conditional noise failure total $6\exp\pth{-L}=\delta/2\le\delta$. The sample-size conditions hold for all sufficiently large $n$ at fixed $\delta$ or $\delta=1/n$, with thresholds independent of $\lambda_0$.
\end{proof}

\subsection{Finite-width comparison for the biased vanilla network}
\label{sec:vanilla-finite-width-comparison}
We verify the finite-width comparison for (\ref{eq:vanilla-bias-network}) by uniform feature concentration and a parameter bootstrap.

Write $p=d+1$ and retain the augmented inputs, $q$, $b$ and ReLU activation from Section~\ref{sec:setup}.
We use unnormalized Gram matrices on the augmented inputs. For a fixed training sample $(\bbx_i,y_i)_{i=1}^n$, define
\bal\label{eq:vanilla-design-constants}
\bK=(K(\bbx_i,\bbx_j))_{i,j=1}^n,\qquad
\lambda_0=\lambda_{\min}(\bK)>0,\qquad
q_{\max}=\max_{i\in[n]}q(\bbx_i),\qquad
Y_n=\frac{\ltwonorm{(y_i)_{i=1}^n}}{\sqrt n}.
\eal
The positive-eigenvalue requirement is the almost-sure conclusion of Lemma~\ref{lemma:unbounded-normalization}. Let $\bW(t)$ have rows $\bbw_r(t)^\top$, let $\ba(0)=0$ and let the rows $\bbw_r(0)$ be independent $\cN(0,\kappa^2\bI_p)$ vectors. We abbreviate the network in (\ref{eq:vanilla-bias-network}) by $f_t^{\textup{(vanilla,bias)}}$. The kernel gradient flow $\hat f_t^K$ starts from zero and solves
\bal\label{eq:vanilla-zero-kernel-flow}
\frac{d}{dt}\hat f_t^K=-\frac1n\sum_{i=1}^n(\hat f_t^K(\bbx_i)-y_i)K_{\bbx_i},\qquad
\hat f_0^K=0.
\eal
If $\by=(y_i)_{i=1}^n$ and $\bk(\bx)=(K(\bx,\bbx_i))_{i=1}^n$, then
$\hat f_t^K(\bx)=\bk(\bx)^\top \bK^{-1}(\bI_n-\exp\pth{-\bK t/n})\by,$ and $\bu_K(t)=(\hat f_t^K(\bbx_i)-y_i)_{i=1}^n=-\exp\pth{-\bK t/n}\by.$

\begin{lemma}\label{lemma:vanilla-uniform-initial-covariance}
Fix $\delta_*\in(0,1)$ and a deterministic integer $m\ge\max\set{2,\log(4/\delta_*)}$. Define
\bal\label{eq:vanilla-initial-event-constants}
B_{\rm init}=3\kappa\sqrt p,\qquad
u_* =\log(8/\delta_*),\qquad
\ell_m=2p\log(1+2m)+u_*,\qquad
\epsilon_m=\kappa^2\left(8\sqrt{\frac{\ell_m}{m}}+\frac{4\ell_m+20p}{m}\right).
\eal
With probability at least $1-\delta_*$ over $\bW(0)$, the event $\cE_{m,\delta_*}$ consisting of the following two inequalities holds:
\bal\label{eq:vanilla-initial-covariance-event}
\frac{\fnorm{\bW(0)}}{\sqrt m}\le B_{\rm init},\qquad
\sup_{\bx,\bx'\in\RR^d}
\frac{\abth{K_m^0(\bx,\bx')-K(\bx,\bx')}}{q(\bx)q(\bx')}\le\epsilon_m,
\eal
where $K_m^0(\bx,\bx')=m^{-1}\sum_{r=1}^m\sigma(\bbw_r(0)^\top\tilde \bx)\sigma(\bbw_r(0)^\top\tilde \bx')$. In particular,
\bal\label{eq:vanilla-initial-covariance-coarse}
\epsilon_m\le100\kappa^2(p+u_*+1)m^{-1/4}.
\eal
The event does not depend on the training sample. The supremum in (\ref{eq:vanilla-initial-covariance-event}) is unchanged when restricted to a countable dense subset of $\RR^d\times\RR^d$, by continuity, so this event is measurable.
\end{lemma}

\begin{proof}
Apply the common Gaussian lemma~\ref{lemma:Gaussian-feature-net} underlying Theorem~\ref{theorem:sup-hat-g}, with dimension $p=d+1$, failure $\delta_*$ and $\ell=\ell_m$. Its width hypothesis holds. With probability $1-\delta_*/2$, $\fnorm{\bW(0)}^2/m\le5\kappa^2p\le B_{\rm init}^2$ and
$\sup_{\ltwonorm{\bu}=\ltwonorm{\bv}=1}\abth{F_m(\bu,\bv)-K(\bu,\bv)} \le \kappa^2\left(8\sqrt{\frac{\ell_m}{m}}+\frac{4\ell_m+20p}{m}\right)=\epsilon_m.$
With $B_{\rm init}=3\kappa\sqrt p$, homogeneity at $\bu=\tilde \bx/q(\bx)$, $\bv=\tilde \bx'/q(\bx')$ gives (\ref{eq:vanilla-initial-covariance-event}). Since $\log(1+2m)\le3\sqrt m$,
$\epsilon_m\le \kappa^2\set{8\sqrt{6p+u_*}+4(6p+u_*)+20p}m^{-1/4}
\le 100\kappa^2(p+u_*+1)m^{-1/4}$.
This proves (\ref{eq:vanilla-initial-covariance-coarse}).
\end{proof}

\begin{proposition}\label{prop:vanilla-all-time-coupling}
Retain the fixed sample in (\ref{eq:vanilla-design-constants}) and the constants in (\ref{eq:vanilla-initial-event-constants}). Define
\bal\label{eq:vanilla-bootstrap-constants}
D_{\rm v}=100\kappa^2(p+u_*+1)+2B_{\rm init}+2,\qquad
H_n=\frac{2nq_{\max}Y_n}{\lambda_0},\qquad
L_n=H_n\left(1+\frac{2bnq_{\max}^2}{\lambda_0}\right).
\eal
Suppose
\bal\label{eq:vanilla-bootstrap-width}
m\ge\max\set{2,u_*,\,[4(H_n+1)(B_{\rm init}+2)]^4,\,
\left[\frac{4nq_{\max}^2D_{\rm v}}{\lambda_0}\right]^4}.
\eal
Let $\bH_m(t)$ denote the raw tangent Gram matrix of the network on the training inputs, evaluated along a trajectory using its selected measurable ReLU slopes. The proof below gives its entrywise expression. On $\cE_{m,\delta_*}$ every absolutely continuous negative Clarke-gradient trajectory for the empirical squared loss is global and satisfies, for every $t\ge0$,
\bal\label{eq:vanilla-all-time-parameter-bounds}
\ltwonorm{\ba(t)}\le\frac14m^{1/4},\qquad
\fnorm{\bW(t)-\bW(0)}\le\frac14m^{1/4},\qquad
\lambda_{\min}(\bH_m(t))\ge\frac34\lambda_0.
\eal
In particular, the raw tangent Gram matrix is bounded below by $\lambda_0\bI_n/2$. The residual $\bu(t)=(f_t^{\textup{(vanilla,bias)}}(\bbx_i)-y_i)_{i=1}^n$ and the network-to-kernel difference satisfy
\bal\label{eq:vanilla-all-time-coupling-bound}
\frac{\ltwonorm{\bu(t)}}{\sqrt n}\le Y_n\exp\pth{-\lambda_0t/(2n)},\qquad
\sup_{t\ge0,\,\bx\in\RR^d}
\frac{\abth{f_t^{\textup{(vanilla,bias)}}(\bx)-\hat f_t^K(\bx)}}{q(\bx)}
\le D_{\rm v}L_nm^{-1/4}.
\eal
These conclusions hold for every choice of a trajectory. A universally measurable trajectory selection can be fixed once and for all on the completed probability space.
\end{proposition}

\begin{proof}

\smallskip
\noindent Step 1. Limits of ordinary loss gradients and convex combinations represent every Clarke subgradient by slopes $\zeta_{ri}\in[0,1]$, equal to the usual indicator off zero preactivations. The limiting gradients impose joint constraints on the admissible slopes. The compact slope sections have measurable graph, so a selected subgradient has a measurable representation. Almost everywhere,
\bal\label{eq:vanilla-Clarke-flow-equations}
\dot a_r=-\frac1{n\sqrt m}\sum_{i=1}^nu_i\sigma(\bbw_r^\top\tilde\bbx_i),\qquad
\dot{\bbw}_r=-\frac{a_r}{n\sqrt m}\sum_{i=1}^nu_i\zeta_{ri}\tilde\bbx_i.
\eal
An absolutely continuous scalar has derivative zero almost everywhere on each level set. Hence zero preactivations have derivative zero there, and the ReLU chain rule holds for these same slopes. It gives
\bals
\dot\bu\stackrel{\circled{1}}{=}-\bH_m\bu/n,\qquad
[\bH_m]_{ij}=\frac1m\sum_{r=1}^m\set{\sigma(\bbw_r^\top\tilde\bbx_i)\sigma(\bbw_r^\top\tilde\bbx_j)+a_r^2\zeta_{ri}\zeta_{rj}\tilde\bbx_i^\top\tilde\bbx_j}.
\eals
This is a positive semidefinite Jacobian Gram matrix, for every admissible representation. Local existence follows from the upper semicontinuous inclusion theorem with nonempty compact convex values~\cite[pp.~93--138]{AubinCellina1984-differential-inclusions}: the locally Lipschitz piecewise-polynomial loss has precisely these properties. The flow equations also give $\fnorm{\dot\bW}^2+\ltwonorm{\dot\ba}^2=\bu^\top\bH_m\bu/n^2$. Thus the chain rule gives
\bals
\frac d{dt}\frac{\ltwonorm{\bu}^2}{2n}\stackrel{\circled{2}}{=}-\fnorm{\dot\bW}^2-\ltwonorm{\dot\ba}^2\le 0.
\eals
It follows that $\ltwonorm{\bu}/\sqrt n\le Y_n$. The flow equations, for $V=\ltwonorm{\ba}+\fnorm{\bW}$, give
$\dot V\le q_{\max}Y_nV/\sqrt m$, and $V(t)\le V(0)\exp\pth{q_{\max}Y_nt/\sqrt m}$.
Gronwall therefore prevents finite-time escape and gives global continuation.

To fix a measurable solution, write $\btheta=(\bW,\ba)$ and $\cD=((\bbx_i,y_i))_{i=1}^n$. Let $\cL_{\cD}$ denote the empirical squared loss and $\partial^C$ the Clarke subdifferential. Set $F(\cD,\btheta)=-\partial^C\cL_{\cD}(\btheta)$. Its graph is semialgebraic: finitely many activation regions have polynomial gradients. Closure membership with $\cD$ held fixed is a first-order real formula, preserved by quantifier elimination. Convex combinations of at most $mp+m+1$ limiting gradients describe the Clarke graph. In the Polish compact-open path space, absolute continuity on each compact interval is Borel by the criterion with rational endpoints and countably many tolerances. Its almost-everywhere derivative $D\btheta$ is a Borel limit of forward rational difference quotients, set to zero where no limit exists. The semialgebraic compact-valued $F$ has Borel distance, so for each positive integer $j$,
$\int_0^j\min\set{1,\operatorname{dist}(D\btheta(t),F(\cD,\btheta(t)))}\,dt=0$
is a Borel path condition. These conditions, absolute continuity and the initial value give a Borel solution graph with nonempty sections. The uniformization theorem~\cite[Section~18]{Kechris1995-classical-descriptive-set-theory} gives a universally measurable selection. Completion of the sampling space makes its evaluations measurable. We now establish the bounds uniformly over these solutions.

\smallskip
\noindent Step 2. Put $A(t)=\ltwonorm{\ba(t)}$, $U(t)=\fnorm{\bW(t)-\bW(0)}$, and work on $\cE_{m,\delta_*}$, using (\ref{eq:vanilla-initial-covariance-coarse}), up to the first exit from $A,U<m^{1/4}$. ReLU feature norms are at most $q(\bx)\fnorm{\bW(t)}$ and their movements at most $q(\bx)U$. For every test slope $\chi_r(\bx,t)\in[0,1]$, with its prescribed nonzero-preactivation value, define
\bals
\bH_m(t,\bx,\bbx_i)=\frac1m\sum_{r=1}^m\set{\sigma(\bbw_r(t)^\top\tilde \bx)\sigma(\bbw_r(t)^\top\tilde\bbx_i)+a_r(t)^2\chi_r(\bx,t)\zeta_{ri}(t)\tilde \bx^\top\tilde\bbx_i}.
\eals
The feature-product difference and $\abth{\chi_r\zeta_{ri}}\le1$ yield, also entrywise for training inputs,
\bals
\frac{\abth{\bH_m(t,\bx,\bbx_i)-K(\bx,\bbx_i)}}{q(\bx)q(\bbx_i)}
\stackrel{\circled{3}}{\le}\epsilon_m+2B_{\rm init}U/\sqrt m+(U^2+A^2)/m
\le D_{\rm v}m^{-1/4}.
\eals
For each fixed $\bx$ the ordinary test indicator, zero at zero preactivation, obeys the same almost-everywhere chain rule. The bound is uniform in input and admissible slopes. By the Gram width condition,
$\norm{\bH_m(t)-\bK}{\rm op}\le nq_{\max}^2D_{\rm v}m^{-1/4}\le \lambda_0/4.$
The tangent Gram matrix therefore has lower bound $3\lambda_0/4$ throughout this interval.

\smallskip
\noindent Step 3. Write $E(t)=\ltwonorm{\bu(t)}/\sqrt n$. The residual equation and weaker lower bound $\lambda_0/2$ give \newline
$E(t)\le Y_n\exp\pth{-\lambda_0t/(2n)}$, and $\int_0^tE(s)\,ds\le 2nY_n/\lambda_0$.
The factor $n$ comes from the loss normalization and raw Gram matrix. The flow equations (\ref{eq:vanilla-Clarke-flow-equations}) bound the speeds by $q_{\max}(B_{\rm init}+U/\sqrt m)E$ and $q_{\max}AE/\sqrt m$. Integrating using only the assumed $A,U<m^{1/4}$ gives
$A(t)\le H_n(B_{\rm init}+m^{-1/4})\le \tfrac14m^{1/4},$ and $U(t)\le H_nm^{-1/4}\le \tfrac14m^{1/4}.$
The width condition (\ref{eq:vanilla-bootstrap-width}) implies these parameter estimates and $m^{1/2}\ge4H_n$. Since $A(0)=U(0)=0$ and both functions are continuous, the strict improvements exclude a finite first exit. This proves (\ref{eq:vanilla-all-time-parameter-bounds}) and the residual bound for every time.

\smallskip
\noindent Step 4. Put $\bv=\bu-\bu_K$ and $\omega_m=D_{\rm v}m^{-1/4}$. Variation of constants for $\dot\bv=-\bK\bv/n-(\bH_m-\bK)\bu/n$, $\bv(0)=0$, gives
\bals
\frac{\ltwonorm{\bv(t)}}{\sqrt n}&\stackrel{\circled{4}}{\le}q_{\max}^2\omega_m\int_0^t\exp\pth{-\lambda_0(t-s)/n}Y_n\exp\pth{-\lambda_0s/(2n)}\,ds
\le \frac{2nq_{\max}^2\omega_mY_n}{\lambda_0}\exp\pth{-\lambda_0t/(2n)},\\
\int_0^\infty\frac{\ltwonorm{\bv(t)}}{\sqrt n}\,dt&\le \frac{4n^2q_{\max}^2\omega_mY_n}{\lambda_0^2}.
\eals
Evaluating the exponential convolution and then integrating establishes accumulated residual discrepancy before comparing functions. Integrate the output chain rule at fixed $\bx$, subtract (\ref{eq:vanilla-zero-kernel-flow}), and use $\abth{K(\bx,\bbx_i)}\le bq(\bx)q(\bbx_i)$:
\bals
\abth{f_t^{\textup{(vanilla,bias)}}(\bx)-\hat f_t^K(\bx)}
&\stackrel{\circled{5}}{\le}q(\bx)q_{\max}\set{\omega_m\int_0^tE(s)\,ds+b\int_0^t\frac{\ltwonorm{\bv(s)}}{\sqrt n}\,ds} \le q(\bx)\omega_mH_n\left(1+\frac{2bnq_{\max}^2}{\lambda_0}\right).
\eals
The two terms are cross-kernel drift and accumulated residual discrepancy. For each fixed input, the preactivations are absolutely continuous on bounded time intervals. Their derivatives vanish almost everywhere on their zero sets, so the test-output chain rule can be integrated from time zero despite activation changes. Take a countable dense set of test inputs and all integer time horizons. The integrated inequality holds for every time on each horizon, and its constants are independent of the input and horizon. Continuity of both outputs in the input then extends it to all of $\RR^d$. This proves the uniform estimate (\ref{eq:vanilla-all-time-coupling-bound}) on the initialization event.
\end{proof}

\begin{corollary}\label{corollary:vanilla-polynomial-width-transfer}
Assume $P(\ltwonorm{\bX}>R)\le2\exp\pth{-a R^2}$ for all $R\ge0$, with fixed $a>0$. Suppose the independent conditionally mean-zero noise is $\sigma_0$-sub-Gaussian and $f^*\in[\cH_K]^{s'}(\mu_0)$ with $s'\ge1$. Let $C$ be the population covariance on $\cH_K$, let $\lambda_1=\norm{C}{\rm op}$ and retain $D_0=b\lambda_1^{s'-1}$ and $M_2=\Expect{P}{q(\bX)^2}$. For $\delta\in(0,1)$ set $\delta_*=\delta/3$ in (\ref{eq:vanilla-initial-event-constants}) and (\ref{eq:vanilla-bootstrap-constants}). In addition to (\ref{eq:vanilla-bootstrap-width}), impose
\bal\label{eq:vanilla-transfer-width}
m\ge\pth{nD_{\rm v}L_n\sqrt{M_2}}^4.
\eal
For each deterministic width $m$, there is an event of probability at least $1-\delta$ over the sample, noise and initialization on which these width conditions imply
\bal\label{eq:vanilla-coarse-uniform-coupling}
\sup_{t\ge0}\norm{f_t^{\textup{(vanilla,bias)}}-\hat f_t^K}{L^2(P)}
\le D_{\rm v}L_n\sqrt{M_2}\,m^{-1/4}\le\frac1n.
\eal
The width conditions are implied by $m\ge C(1+n+\log(1/\delta)+\lambda_0^{-1})^{32}$ with a sufficiently large constant depending only on the fixed law and model parameters.
\end{corollary}
\begin{proof}
If the prescribed width is below $\max\set{2,u_*}$, its bootstrap antecedent is false and the whole probability space is an admissible event. Otherwise the initialization lemma applies, since $u_*>\log(4/\delta_*)$.

The source representation $f^*=C^{(s'-1)/2}\bv$, $\norm{\bv}{\cH_K}\le\mu_0$, gives by reproduction and spectral calculus
$\abth{f^*(\bx)}\le\sqrt{K(\bx,\bx)}\norm{C^{(s'-1)/2}\bv}{\cH_K}\le\mu_0\sqrt{D_0}\,q(\bx)$.
Applying the covariate and conditional noise tail bounds and a union bound over $n$ gives, with failure probability at most $2\delta/3$,
\bal\label{eq:vanilla-data-tail-event}
q_{\max}^2\le 1+a^{-1}\log(6n/\delta),\qquad
Y_n\le\mu_0\sqrt{D_0}\,q_{\max}+\sigma_0\sqrt{2\log(6n/\delta)}.
\eal
The conditional initialization failure is at most $\delta/3$, so the intersection has probability $1-\delta$. For $M=1+n+\log(1/\delta)+\lambda_0^{-1}$, (\ref{eq:vanilla-data-tail-event}) and the definitions give
$q_{\max}^2\le CM$, $Y_n^2\le CM$, $D_{\rm v}\le CM$, $H_n\le CM^3$ and $L_n\le CM^6$.
Indeed $\log(6n/\delta),u_*,q_{\max}Y_n\le CM$ and $n,\lambda_0^{-1}\le M$. The bootstrap thresholds in (\ref{eq:vanilla-bootstrap-width}) are therefore at most $C\max\set{M,M^{12},M^{16}}$. Since $M_2$ is fixed, the threshold in (\ref{eq:vanilla-transfer-width}) is at most $CM^{32}$. This proves the asserted polynomial sufficient width without a deterministic lower bound on $\lambda_0$.

To transfer the function approximation, retain the factor $q$ in Proposition~\ref{prop:vanilla-all-time-coupling} and integrate its uniform bound. The finite second moment makes this step valid on the entire input space. On the same event, for every trajectory and every $t\ge0$,
\bals
\norm{f_t^{\textup{(vanilla,bias)}}-\hat f_t^K}{L^2(P)}^2
&=\Expect{P}{q(\bX)^2\left(\frac{f_t^{\textup{(vanilla,bias)}}(\bX)-\hat f_t^K(\bX)}{q(\bX)}\right)^2}
\le M_2D_{\rm v}^2L_n^2m^{-1/2}.
\eals
Taking the supremum over $t$ and applying (\ref{eq:vanilla-transfer-width}) proves (\ref{eq:vanilla-coarse-uniform-coupling}) on the same event.
\end{proof}

\subsection{Completion of the regression risk comparison}
\begin{proof}[\textbf{\textup{Proof of Proposition~\ref{prop:vanilla-bias-regression}}}]
Apply Proposition~\ref{prop:vanilla-biased-kernel-statistics} and Corollary~\ref{corollary:vanilla-polynomial-width-transfer} with failure $\delta/2$ and initialization parameter $\delta_* =\delta/6$. For fixed $\delta$ or $\delta=1/n$, the conditions (\ref{eq:vanilla-covariance-HS}) and (\ref{eq:vanilla-leverage-trace-concentration}) hold for all sufficiently large $n$, independently of $\lambda_0$. The $M^{32}$ estimate in the corollary shows that (\ref{eq:vanilla-comparison-width}), with sufficiently large fixed $C_{\rm v}$, implies the bootstrap and transfer conditions. Thus the stated implication holds on an event of probability at least $1-\delta$ for each fixed $m,n$.

The transfer (\ref{eq:vanilla-coarse-uniform-coupling}) holds for every trajectory and every time. At deterministic $t_{\rm op}$, therefore,
\bals
\norm{f_{t_{\rm op}}^{\textup{(vanilla,bias)}}-f^*}{\mu}^2
&\le 2\norm{\hat f_{t_{\rm op}}^K-f^*}{\mu}^2+2\norm{f_{t_{\rm op}}^{\textup{(vanilla,bias)}}-\hat f_{t_{\rm op}}^K}{\mu}^2\\
&\le C\log^2(24/\delta)n^{-s'(d+3)/(s'(d+3)+d)}+2n^{-2}
\le C\log^2(12/\delta)n^{-s'(d+3)/(s'(d+3)+d)}.
\eals
The last step uses $\log(24/\delta)\le2\log(12/\delta)$ and the risk exponent below one. The completed-space estimator is measurable, with constants uniform over the source ball, proving (\ref{eq:vanilla-li-regression-risk}).
\end{proof}

\endgroup

\end{appendices}

\bibliographystyle{IEEEtran}
\bibliography{ref}

\end{document}